# A Unifying Framework in Vector-valued Reproducing Kernel Hilbert Spaces for Manifold Regularization and Co-Regularized Multi-view Learning

**Hà Quang Minh**                                        MINH.HAQUANG@IIT.IT

**Loris Bazzani**                                        LORIS.BAZZANI@IIT.IT

**Vittorio Murino**                                      VITTORIO.MURINO@IIT.IT
*Pattern Analysis and Computer Vision (PAVIS)*
*Istituto Italiano di Tecnologia*
*Via Morego 30, Genova 16163, ITALY*

**Editor:**

## Abstract

This paper presents a general vector-valued reproducing kernel Hilbert spaces (RKHS) framework for the problem of learning an unknown functional dependency between a structured input space and a structured output space. Our formulation encompasses both Vector-valued Manifold Regularization and Co-regularized Multi-view Learning, providing in particular a unifying framework linking these two important learning approaches. In the case of the least square loss function, we provide a closed form solution, which is obtained by solving a system of linear equations. In the case of Support Vector Machine (SVM) classification, our formulation generalizes in particular both the binary Laplacian SVM to the multi-class, multi-view settings and the multi-class Simplex Cone SVM to the semi-supervised, multi-view settings. The solution is obtained by solving a single quadratic optimization problem, as in standard SVM, via the Sequential Minimal Optimization (SMO) approach. Empirical results obtained on the task of object recognition, using several challenging datasets, demonstrate the competitiveness of our algorithms compared with other state-of-the-art methods.

## 1. Introduction

Reproducing kernel Hilbert spaces (RKHS) and kernel methods have been by now established as among the most powerful paradigms in modern machine learning and statistics (Schölkopf and Smola, 2002; Shawe-Taylor and Cristianini, 2004). While most of the literature on kernel methods has so far focused on scalar-valued functions, RKHS of vector-valued functions have received increasing research attention in machine learning recently, from both theoretical and practical perspectives (Micchelli and Pontil, 2005; Carmeli et al., 2006; Reisert and Burkhardt, 2007; Caponnetto et al., 2008; Brouard et al., 2011; Dinuzzo et al., 2011; Kadri et al., 2011; Minh and Sindhwani, 2011; Zhang et al., 2012; Sindhwani et al., 2013). In this paper, we present a general learning framework in the setting of vector-valued RKHS that encompasses learning across three different paradigms, namely vector-valued, multi-view, and semi-supervised learning, simultaneously.

The direction of Multi-view Learning we consider in this work is Co-Regularization, see e.g. (Brefeld et al., 2006; Sindhwani and Rosenberg, 2008; Rosenberg et al., 2009; Sun, 2011). In this approach, different hypothesis spaces are used to construct target functions based on different views of the input data, such as different features or modalities, and a data-dependent regularization term is used to enforce consistency of output values from different views of the same input example. The resulting target functions, each corresponding to one view, are then naturally combined together in a principled way to give the final solution.

The direction of Semi-supervised Learning we follow here is Manifold Regularization (Belkin et al., 2006; Brouard et al., 2011; Minh and Sindhwani, 2011), which attempts to learn the geometry of the input space by exploiting the given unlabeled data. The latter two papers are recent generalizations of the original scalar version of manifold regularization of (Belkin et al., 2006) to the vector-valued setting. In (Brouard et al., 2011), a vector-valued version of the graph Laplacian $L$ is used, and in (Minh and Sindhwani, 2011), $L$ is a general symmetric, positive operator, including the graph Laplacian. The vector-valued setting allows one to capture possible dependencies between output variables by the use of, for example, an output graph Laplacian.

The formulation we present in this paper gives a unified learning framework for the case the hypothesis spaces are vector-valued RKHS. Our formulation is general, encompassing many common algorithms as special cases, including both Vector-valued Manifold Regularization and Co-regularized Multi-view Learning. The current work is a significant extension of our conference paper (Minh et al., 2013). In the conference version, we stated the general learning framework and presented the solution for multi-view least square regression and classification. In the present paper, we also provide the solution for multi-view multi-class Support Vector Machine (SVM), which includes multi-view binary SVM as a special case. Furthermore, we present a principled optimization framework for computing the optimal weight vector for combining the different views, which correspond to different kernels defined on the different features in the input data. An important and novel feature of our formulation compared to traditional multiple kernel learning methods is that it does *not* constrain the combining weights to be non-negative, leading a considerably simpler optimization problem, with an almost closed form solution in the least square case.

Our numerical experiments were performed using a special case of our framework, namely Vector-valued Multi-view Learning. For the case of least square loss function, we give a closed form solution which can be implemented efficiently. For the multi-class SVM case, we implemented our formulation, under the simplex coding scheme, using a Sequential Minimal Optimization (SMO) algorithm, which we obtained by generalizing the SMO technique of (Platt, 1999) to our setting.

We tested our algorithms on the problem of multi-class image classification, using three challenging, publicly available datasets, namely Caltech-101 (Fei-Fei et al., 2006), Caltech-UCSD-Birds-200-2011 (Wah et al., 2011), and Oxford Flower 17 (Nilsback and Zisserman, 2006). The results obtained are promising and demonstrate the competitiveness of our learning framework compared with other state-of-the-art methods.

## 1.1 Related work

Recent papers in the literature that are closely related to our work include (Rosenberg et al., 2009; Sun, 2011; Luo et al., 2013a,b; Kadri et al., 2013). We analyze and compare each of these methods to our proposed framework in the following.

In the scalar setting, two papers that seek to generalize the manifold regularization framework of (Belkin et al., 2006) to the multi-view setting are (Rosenberg et al., 2009; Sun, 2011). In (Sun, 2011), the author proposed a version of the Multi-view Laplacian SVM, which, however, only deals with *two views* and is not generalizable to any number of views. In (Rosenberg et al., 2009), the authors formulated a version of the semi-supervised Multi-view learning problem for any number of views, but instead of solving it directly like we do, they proposed to compute the Multi-view kernel and reduce the problem to the supervised case. One problem with this approach is that the Multi-view kernel is complicated analytically, which makes it difficult to implement efficiently in practice. It is also unclear how this approach can be generalized to the multi-class setting.

In the vector-valued setting, papers dealing with multi-view learning include (Luo et al., 2013a,b), where each view is used to define a kernel and a graph Laplacian, and the resulting kernels and graph Laplacians are respectively linearly combined to give the final kernel and final graph Laplacian. Thus this approach does not take into account between-view consistency as in our approach. In (Luo et al., 2013a), which generalizes the vector-valued regularization formulation of (Minh and Sindhwani, 2011), the loss function is the least square loss. In (Luo et al., 2013b), the authors employed a multi-class SVM loss function, which is the average of the binary SVM hinge loss across all components of the output vector. To the best of our knowledge, we are not aware of any theoretical result on the statistical consistency of this loss function.

In the direction of multi-class learning, many versions of multi-class SVM have appeared in the literature, e.g. (Lee et al., 2004; Weston and Watkins, 1999; Crammer and Singer, 2001; Mroueh et al., 2012). In this paper, we employ a generalization of the multi-class Simplex Cone SVM (SC-SVM) loss function proposed in (Mroueh et al., 2012), where it was proved to be theoretically consistent.

Another work dealing with multi-view learning in the vector-valued approach is (Kadri et al., 2013), which considers multi-view learning from the multi-task learning perspective, see e.g. (Evgeniou et al., 2005), where different views of the same input example correspond to different tasks which share the same output label. Their formulation does not have an explicit view combination mechanism and is restricted to scalar-valued tasks and the supervised setting. The resulting optimization problem is vector-valued regularized least square regression in (Micchelli and Pontil, 2005), which is a special case of our general learning framework.

Our multi-view learning approach can also be viewed as a form of multiple kernel learning, but it is different from typical multiple kernel learning approaches, see e.g. (Bach et al., 2004; Bucak et al., 2014) in several aspects. First, it is formulated in both supervised and semi-supervised settings. Second, it incorporates between-view interactions. Third, it makes no mathematical constraints, such as non-negativity, on the combining weights. This last aspect of our framework contrasts sharply with typical multiple kernel learning methods, which need to constrain the combining weights to be non-negative in order to guarantee the positive definiteness of the combined kernel. As a consequence, our optimization procedure for the combining weights is considerably simpler and has an almost closed form solution in the least square case. We give a brief technical description on the connections between our framework and multiple kernel and multi-task learning in the final part of the paper. Empirically, experimental results reported in the current paper show that our framework performs very favorably compared with state of the art multiple kernel learning methods.

We compared the proposed framework from a methodological point of view with approaches that focus on combining different features in the input data. Our work is complementary to other approaches such as (Zeiler and Fergus, 2014; Razavian et al., 2014; He et al., 2015), which are focused on engineering or learning the best features for the task at hand. In fact, an interesting research direction would be the application of our framework on top of those methods, which will explored in a future work.

## 1.2 Our Contributions

Our learning framework provides a unified formulation for Manifold Regularization and Co-regularized Multi-view Learning in the vector-valued setting. In particular, it generalizes the Vector-valued Manifold Regularization framework of (Minh and Sindhwani, 2011), which was formulated in the single-view setting, with the least square loss, to the multi-view setting, with both least square and multi-class SVM loss functions. Consequently, it generalizes the Vector-valued Regularized Least Square formulation of (Micchelli and Pontil, 2005), which was formulated in the supervised, single-view settings, with the least square loss, to the semi-supervised, multi-view settings, with both the least square and multi-class SVM loss functions.

For the case of SVM classification, our framework is a generalization of the multi-class SC-SVM of (Mroueh et al., 2012), which is supervised and single-view, to the semi-supervised and multi-view learning settings. The loss function that we employ here is also a generalization of the SC-SVM loss functions proposed in (Mroueh et al., 2012). We also show that our formulation is a generalization of the semi-supervised Laplacian SVM of (Belkin et al., 2006), which is binary and single-view, to the multi-class and multi-view learning settings.

The generality and advantage of our vector-valued RKHS approach is illustrated by the fact that it can simultaneously (i) deal with any number of classes in multi-class classification, (ii) combine any number of views, (iii) combine the views using an arbitrary weight vector, and (iv) compute all the different output functions associated with the individual views, all by solving a single system linear equations (in the case of least square loss) or a single quadratic optimization problem (in the case of SVM loss). To the best of our knowledge, this work is the first attempt to present a unified general learning framework whose components have been only individually and partially covered in the literature.

Our optimization framework for computing the optimal weight vector for combining the different views is also novel compared to typical multiple kernel learning methods in that it does *not* constrain the combining weights to be non-negative, leading a considerably simpler optimization problem, with an almost closed form solution in the least square case.

### 1.3 Organization

We start by giving a review of vector-valued RKHS in Section 2. In Section 3, we state the general optimization problem for our learning formulation, together with the Representer Theorem, the explicit solution for the vector-valued least square case, and the quadratic optimization problem for the vector-valued SVM case. We describe Vector-valued Multi-view Learning in Section 4 and its implementations in Section 5, both for the least square and SVM loss functions. Section 6 provides the optimization of the operator that combines the different views for the least square case. Empirical experiments are described in detail in Section 7. Connections between our framework and multi-kernel learning and multi-task learning are briefly described in Section 8. **Proofs for all mathematical results in the paper are given in Appendix A.**

## 2. Vector-Valued RKHS

In this section, we give a brief review of RKHS of vector-valued functions[1], for more detail see e.g. (Carmeli et al., 2006; Micchelli and Pontil, 2005; Caponnetto et al., 2008; Minh and Sindhwani, 2011). In the following, denote by $\mathcal{X}$ a nonempty set, $\mathcal{W}$ a real, separable Hilbert space with inner product $\langle \cdot, \cdot \rangle_{\mathcal{W}}$, $\mathcal{L}(\mathcal{W})$ the Banach space of bounded linear operators on $\mathcal{W}$. Let $\mathcal{W}^{\mathcal{X}}$ denote the vector space of all functions $f : \mathcal{X} \to \mathcal{W}$. A function $K : \mathcal{X} \times \mathcal{X} \to \mathcal{L}(\mathcal{W})$ is said to be an **operator-valued positive definite kernel** if for each pair $(x, z) \in \mathcal{X} \times \mathcal{X}$, $K(x, z)^* = K(z, x)$, and

$$\sum_{i,j=1}^{N} \langle y_i, K(x_i, x_j) y_j \rangle_{\mathcal{W}} \geq 0 \tag{1}$$

for every finite set of points $\{x_i\}_{i=1}^{N}$ in $\mathcal{X}$ and $\{y_i\}_{i=1}^{N}$ in $\mathcal{W}$. Given such a $K$, there exists a unique $\mathcal{W}$-valued RKHS $\mathcal{H}_K$ with reproducing kernel $K$, which is constructed as follows. For each $x \in \mathcal{X}$ and $y \in \mathcal{W}$, form a function $K_x y = K(., x) y \in \mathcal{W}^{\mathcal{X}}$ defined by

$$(K_x y)(z) = K(z, x) y \quad \text{for all} \quad z \in \mathcal{X}.$$

Consider the set $\mathcal{H}_0 = \text{span}\{K_x y \mid x \in \mathcal{X}, \ y \in \mathcal{W}\} \subset \mathcal{W}^{\mathcal{X}}$. For $f = \sum_{i=1}^{N} K_{x_i} w_i$, $g = \sum_{i=1}^{N} K_{z_i} y_i \in \mathcal{H}_0$, we define the inner product

$$\langle \, f, g \, \rangle_{\mathcal{H}_K} = \sum_{i,j=1}^{N} \langle w_i, K(x_i, z_j) y_j \rangle_{\mathcal{W}},$$

which makes $\mathcal{H}_0$ a pre-Hilbert space. Completing $\mathcal{H}_0$ by adding the limits of all Cauchy sequences gives the Hilbert space $\mathcal{H}_K$. The **reproducing property** is

$$\langle f(x), y \rangle_{\mathcal{W}} = \langle f, K_x y \rangle_{\mathcal{H}_K} \quad \text{for all} \quad f \in \mathcal{H}_K. \tag{2}$$

**Sampling Operators.** For each $x \in \mathcal{X}$, let $K_x : \mathcal{W} \to \mathcal{H}_K$ be the operator with $K_x y$ defined as above, then

$$||K_x y||_{\mathcal{H}_K}^2 = \langle K(x, x) y, y \rangle_{\mathcal{W}} \leq ||K(x, x)|| \, ||y||_{\mathcal{W}}^2,$$

---

1. Some authors, e.g. (Kadri et al., 2011) employ the terminology *function-valued*, which is equivalent to *vector-valued*: a function is a vector in a vector space of functions (e.g. a Hilbert space of functions), and an $n$-dimensional vector is a discrete function defined on $n$ points.

which implies that

$$||K_x : \mathcal{W} \to \mathcal{H}_K|| \leq \sqrt{||K(x,x)||},$$

so that $K_x$ is a bounded operator. Let $K_x^* : \mathcal{H}_K \to \mathcal{W}$ be the adjoint operator of $K_x$, then from (2), we have

$$f(x) = K_x^* f \qquad \text{for all} \quad x \in \mathcal{X}, f \in \mathcal{H}_K. \tag{3}$$

From this we deduce that for all $x \in \mathcal{X}$ and all $f \in \mathcal{H}_K$,

$$||f(x)||_{\mathcal{W}} \leq ||K_x^*|| \, ||f||_{\mathcal{H}_K} \leq \sqrt{||K(x,x)||} \, ||f||_{\mathcal{H}_K},$$

that is the *sampling operator* $S_x : \mathcal{H}_K \to \mathcal{W}$ defined by

$$S_x f = K_x^* f = f(x)$$

is bounded. Let $\mathbf{x} = (x_i)_{i=1}^l \in \mathcal{X}^l$, $l \in \mathbb{N}$. For the sampling operator $S_{\mathbf{x}} : \mathcal{H}_K \to \mathcal{W}^l$ defined by $S_{\mathbf{x}}(f) = (f(x_i))_{i=1}^l$, for any $\mathbf{y} = (y_i)_{i=1}^l \in \mathcal{W}^l$,

$$\langle S_{\mathbf{x}} f, \mathbf{y} \rangle_{\mathcal{W}^l} = \sum_{i=1}^l \langle f(x_i), y_i \rangle_{\mathcal{W}} = \sum_{i=1}^l \langle K_{x_i}^* f, y_i \rangle_{\mathcal{H}_K}$$

$$= \sum_{i=1}^l \langle f, K_{x_i} y_i \rangle_{\mathcal{H}_K} = \langle f, \sum_{i=1}^l K_{x_i} y_i \rangle_{\mathcal{H}_K}.$$

Thus the adjoint operator $S_{\mathbf{x}}^* : \mathcal{W}^l \to \mathcal{H}_K$ is given by

$$S_{\mathbf{x}}^* \mathbf{y} = S_{\mathbf{x}}^*(y_1, \dots, y_l) = \sum_{i=1}^l K_{x_i} y_i, \quad \mathbf{y} \in \mathcal{W}^l, \tag{4}$$

and the operator $S_{\mathbf{x}}^* S_{\mathbf{x}} : \mathcal{H}_K \to \mathcal{H}_K$ is given by

$$S_{\mathbf{x}}^* S_{\mathbf{x}} f = \sum_{i=1}^l K_{x_i} f(x_i) = \sum_{i=1}^l K_{x_i} K_{x_i}^* f. \tag{5}$$

**Data-dependent Semi-norms.** Let $(x_1, \dots, x_{u+l}) \subset \mathcal{X}$. Let $M : \mathcal{W}^{u+l} \to \mathcal{W}^{u+l} \in \mathcal{L}(\mathcal{W}^{u+l})$ be a symmetric, positive operator, that is $\langle y, My \rangle_{\mathcal{W}^{u+l}} \geq 0$ for all $y \in \mathcal{W}^{u+l}$. For $f \in \mathcal{H}_K$, let $\mathbf{f} = (f(x_1), \dots, f(x_{u+l})) \in \mathcal{W}^{u+l}$. The operator $M : \mathcal{W}^{u+l} \to \mathcal{W}^{u+l}$ can be expressed as an operator-valued matrix $M = (M_{ij})_{i,j=1}^{u+l}$ of size $(u+l) \times (u+l)$, with each $M_{ij} : \mathcal{W} \to \mathcal{W}$ being a linear operator, so that

$$(M\mathbf{f})_i = \sum_{j=1}^{u+l} M_{ij} \mathbf{f}_j = \sum_{j=1}^{u+l} M_{ij} f(x_j). \tag{6}$$

We can then define the following semi-norm for $f$, which depends on the $x_i$'s:

$$\langle \mathbf{f}, M\mathbf{f} \rangle_{\mathcal{Y}^{u+l}} = \sum_{i,j=1}^{u+l} \langle f(x_i), M_{ij} f(x_j) \rangle_{\mathcal{W}}. \tag{7}$$

This form of semi-norm was utilized in vector-valued manifold regularization (Minh and Sindhwani, 2011).

## 3. General Learning Framework

In this section, we state the general minimization problem that we wish to solve, which includes Vector-valued Manifold Regularization and Multi-view Learning as special cases.

Let the input space be $\mathcal{X}$, an arbitrary non-empty set. Let $\mathcal{Y}$ be a separable Hilbert space, denoting the output space. Assume that there is an unknown probability measure $\rho$ on $\mathcal{X} \times \mathcal{Y}$ and that we have access to a random training sample $\mathbf{z} = \{(x_i, y_i)\}_{i=1}^l \cup \{x_i\}_{i=l+1}^{u+l}$ of $l$ labeled and $u$ unlabeled examples.

Let $\mathcal{W}$ be a separable Hilbert space. Let $K : \mathcal{X} \times \mathcal{X} \to \mathcal{L}(\mathcal{W})$ be an operator-valued positive definite kernel and $\mathcal{H}_K$ its induced Reproducing Kernel Hilbert Space of $\mathcal{W}$-valued functions.

Let $M : \mathcal{W}^{u+l} \to \mathcal{W}^{u+l}$ be a symmetric, positive operator. For each $f \in \mathcal{H}_K$, let

$$\mathbf{f} = (f(x_1), \ldots, f(x_{u+l})) \in \mathcal{W}^{u+l}. \tag{8}$$

Let $V : \mathcal{Y} \times \mathcal{Y} \to \mathbb{R}$ be a convex loss function. Let $C : \mathcal{W} \to \mathcal{Y}$ be a bounded linear operator, with $C^* : \mathcal{Y} \to \mathcal{W}$ its adjoint operator.

The following is the general minimization problem that we wish to solve:

$$f_{\mathbf{z},\gamma} = \operatorname{argmin}_{f \in \mathcal{H}_K} \frac{1}{l} \sum_{i=1}^l V(y_i, Cf(x_i)) + \gamma_A ||f||_{\mathcal{H}_K}^2 + \gamma_I \langle \mathbf{f}, M\mathbf{f} \rangle_{\mathcal{W}^{u+l}}, \tag{9}$$

with regularization parameters $\gamma_A > 0$, $\gamma_I \geq 0$.

Let us give a general multi-view learning interpretation of the different terms in our framework. If each input instance $x$ has many views, then $f(x) \in \mathcal{W}$ represents the output values from all the views, constructed by their corresponding hypothesis spaces. These values are combined by the operator $C$ to give the final output value in $\mathcal{Y}$, which is not necessarily the same as $\mathcal{W}$. In (9), the first term measures the error between the final output $Cf(x_i)$ for $x_i$ with the given output $y_i$, $1 \leq i \leq l$.

The second summand is the standard RKHS regularization term.

The third summand, Multi-view Manifold Regularization, is a generalization of vector-valued Manifold Regularization in (Minh and Sindhwani, 2011) and Multi-view Point Cloud regularization in (Rosenberg et al., 2009): if there is only one view, then it is simply manifold regularization; if there are many views, then it consists of manifold regularization along each view, as well as consistency regularization across different views. We describe one concrete realization of this term in Section 4.2.

**Remark 1** *The framework is readily generalizable to the case the point evaluation functional $f(x)$ is replaced by a general bounded linear operator - we describe this in Appendix B.*

### 3.1 Representer Theorem

The minimization problem (9) is guaranteed to always have a unique global solution, whose form is given by the following Representer Theorem.

**Theorem 2** *The minimization problem (9) has a unique solution, given by $f_{\mathbf{z},\gamma} = \sum_{i=1}^{u+l} K_{x_i} a_i$ for some vectors $a_i \in \mathcal{W}$, $1 \leq i \leq u + l$.*

In the next two sections, we derive the forms of the solution $f_{\mathbf{z},\gamma}$ for the cases where $V$ is the least square loss and the SVM loss, both in the binary and multi-class settings.

## 3.2 Least Square Case

For the case $V$ is the least square loss function, we solve the following problem:

$$f_{\mathbf{z},\gamma} = \operatorname{argmin}_{f \in \mathcal{H}_K} \frac{1}{l} \sum_{i=1}^{l} ||y_i - Cf(x_i)||^2_{\mathcal{Y}} + \gamma_A ||f||^2_{\mathcal{H}_K} + \gamma_I \langle \mathbf{f}, M\mathbf{f} \rangle_{\mathcal{W}_{u+l}}, \tag{10}$$

which has an explicit solution, given by the following.

**Theorem 3** *The minimization problem (10) has a unique solution* $f_{\mathbf{z},\gamma} = \sum_{i=1}^{u+l} K_{x_i} a_i$, *where the vectors* $a_i \in \mathcal{W}$ *are given by*

$$l\gamma_I \sum_{j,k=1}^{u+l} M_{ik} K(x_k, x_j) a_j + C^* C \left( \sum_{j=1}^{u+l} K(x_i, x_j) a_j \right) + l\gamma_A a_i = C^* y_i, \tag{11}$$

*for* $1 \leq i \leq l$, *and*

$$\gamma_I \sum_{j,k=1}^{u+l} M_{ik} K(x_k, x_j) a_j + \gamma_A a_i = 0, \tag{12}$$

*for* $l + 1 \leq i \leq u + l$.

### 3.2.1 OPERATOR-VALUED MATRIX FORMULATION

The system of equations (11) and (12) can be reformulated in matrix form, which is more readable and more convenient to implement efficiently. Let $K[\mathbf{x}]$ denote the $(u+l) \times (u+l)$ operator-valued matrix whose $(i,j)$ entry is $K(x_i, x_j)$. Let $J_l^{\mathcal{W}, u+l} : \mathcal{W}^{u+l} \to \mathcal{W}^{u+l}$ denote the diagonal matrix whose first $l$ entries on the main diagonal are the identity operator $I : \mathcal{W} \to \mathcal{W}$, with the rest being 0. Let $\mathbf{C}^*\mathbf{C} : \mathcal{W}^{u+l} \to \mathcal{W}^{u+l}$ be the $(u+l) \times (u+l)$ diagonal matrix, with each diagonal entry being $C^*C : \mathcal{W} \to \mathcal{W}$. Let $\mathbf{C}^* : \mathcal{Y}^l \to \mathcal{W}^{u+l}$ be the $(u+l) \times l$ block matrix defined by $\mathbf{C}^* = I_{(u+l) \times l} \otimes C^*$, where $I_{(u+l) \times l} = [I_l, 0_{l \times u}]^T$ and $C^* : \mathcal{Y} \to \mathcal{W}$.

**Theorem 4** *The system of equations (11) and (12) in Theorem 3 is equivalent to*

$$(\mathbf{C}^*\mathbf{C} J_l^{\mathcal{W}, u+l} K[\mathbf{x}] + l\gamma_I M K[\mathbf{x}] + l\gamma_A I) \mathbf{a} = \mathbf{C}^* \mathbf{y}, \tag{13}$$

*which has a unique solution* $\mathbf{a}$, *where* $\mathbf{a} = (a_1, \ldots, a_{u+l})$, $\mathbf{y} = (y_1, \ldots, y_l)$ *are considered as column vectors in* $\mathcal{W}^{u+l}$ *and* $\mathcal{Y}^l$, *respectively.*

## 3.3 Vector-valued Multi-view SVM

In this section, we give the solution of the optimization problem (9) when $V$ is a generalization of the binary SVM hinge loss function to the multi-class setting. We first point out one main difference between the least square and SVM cases. In the least square case, there is a natural generalization from the scalar setting to the vector-valued setting, which we treated in the previous section. In contrast, in the SVM case, many different versions of the multi-class SVM loss function have been proposed. In the following, we consider a generalization of the Simplex Cone SVM (SC-SVM) loss function proposed by (Mroueh et al., 2012), where it was shown to be theoretically consistent.

Let the input space $\mathcal{X}$ be an arbitrary non-empty set and the output label space be the discrete set $\mathrm{cl}(\mathcal{Y}) = \{1, \ldots, P\}$, with $P \in \mathbb{N}$, $P \geq 2$, representing the number of classes. In this setting, the random sample $\mathbf{z} = \{(x_i, y_i)\}_{i=1}^l \cup \{x_i\}_{i=l+1}^{u+l}$ is drawn from $\mathcal{X} \times \mathrm{cl}(\mathcal{Y})$.

Let $\mathcal{W}$ be a separable Hilbert space, $K : \mathcal{X} \times \mathcal{X} \to \mathcal{L}(\mathcal{W})$ be a positive definite kernel with value in the Banach space of bounded linear operators $\mathcal{L}(\mathcal{W})$ on $\mathcal{W}$, and $\mathcal{H}_K$ be the RKHS of $\mathcal{W}$-valued functions induced by $K$. Let $\mathcal{Y}$ be a separable Hilbert space. Let $S = [s_1, \ldots, s_P]$ as a matrix, which is potentially infinite, with the $i$th column being $s_i \in \mathcal{Y}$, then $S$ can be considered as a linear operator $S : \mathbb{R}^P \to \mathcal{Y}$, so that for $\mathbf{b} = (b_i)_{i=1}^P$, $S\mathbf{b} = \sum_{i=1}^P b_i s_i$.

Let $C : \mathcal{W} \to \mathcal{Y}$ be a bounded linear operator. Consider the following minimization problem

$$f_{\mathbf{z}, \gamma} = \mathrm{argmin}_{f \in \mathcal{H}_K} \frac{1}{l} \sum_{i=1}^l \sum_{k=1, k \neq y_i}^P \max\left(0, -\langle s_k, s_{y_i} \rangle_{\mathcal{Y}} + \langle s_k, Cf(x_i) \rangle_{\mathcal{Y}}\right)$$
$$+ \gamma_A ||f||_{\mathcal{H}_K}^2 + \gamma_I \langle \mathbf{f}, M\mathbf{f} \rangle_{\mathcal{W}^{u+l}}, \tag{14}$$

with regularization parameters $\gamma_A > 0$ and $\gamma_I \geq 0$.

The components of (14) and their multi-class and multi-view learning interpretations are as follows.

The vectors $s_k$'s in $S$ represent the $P$ different classes. One particular case for $S$, which we employ in our numerical experiments, is the *simplex coding* for multi-class encoding, see e.g. (Hill and Doucet, 2007; Wu and Lange, 2010; Saberian and Vasconcelos, 2011; Mroueh et al., 2012). Recall that a simplex coding is a map $s : \{1, \ldots, P\} \to \mathbb{R}^{P-1}$, such that: (i) $||s_k||^2 = 1$; (ii) $\langle s_j, s_k \rangle = -\frac{1}{P-1}$, $j \neq k$; and (iii) $\sum_{k=1}^P s_k = 0$. The simplex codes $s_k$'s form $P$ maximally and equally separated vectors on the sphere $\mathbb{S}^{P-2}$ in $\mathbb{R}^{P-1}$, each representing one category. For example, for $P = 3$, one set of three $\mathbb{R}^2$-valued code vectors is: $s_1 = (1, 0)$, $s_2 = (-1/2, \sqrt{3}/2)$, $s_3 = (-1/2, -\sqrt{3}/2)$. In general, the simplex codes can be computed by a recursive algorithm, see e.g. (Mroueh et al., 2012). The decoding process is straightforward: given a vector $\mathbf{b} \in \mathbb{R}^{P-1}$, the category we assign to $\mathbf{b}$ is

$$\mathrm{argmax}_{1 \leq k \leq P} \langle \mathbf{b}, s_k \rangle. \tag{15}$$

In the following, we assume that the map $s$ is fixed for each $P$ and also refer to the matrix $S = [s_1, \ldots, s_P]$, with the $i$th column being $s_i$, as the simplex coding, whenever this coding scheme is being used.

If the number of classes is $P$ and $S$ is the simplex coding, then $\mathcal{Y} = \mathbb{R}^{P-1}$ and $S$ is a $(P-1) \times P$ matrix. Let the number of views be $m \in \mathbb{N}$. Let $\mathcal{W} = \mathcal{Y}^m = \mathbb{R}^{(P-1)m}$. Then $K$ is a matrix-valued kernel: for each pair $(x, t) \in \mathcal{X} \times \mathcal{X}$, $K(x, t)$ is a $(P-1)m \times (P-1)m$ matrix. The Hilbert space $\mathcal{H}_K$ induced by $K$ consists of functions $f : \mathcal{X} \to \mathcal{W} = \mathbb{R}^{(P-1)m}$, that is for each $x \in \mathcal{X}$, $f(x) = (f^1(x), \ldots, f^m(x)) \in \mathbb{R}^{(P-1)m}$.

In the first component of (14), the loss function

$$\sum_{k=1, k \neq y_i}^{P} \max \left(0, -\langle s_k, s_{y_i} \rangle_{\mathcal{Y}} + \langle s_k, Cf(x_i) \rangle_{\mathcal{Y}}\right)$$

measures the error between the combined outputs from all the views for $x_i$ with every code vector $s_k$ such that $k \neq y_i$. It is a generalization of the SC-SVM loss function proposed in (Mroueh et al., 2012).

For any $x \in \mathcal{X}$,

$$f_{\mathbf{z}, \gamma}(x) \in \mathcal{W}, \quad Cf_{\mathbf{z}, \gamma}(x) \in \mathcal{Y}, \tag{16}$$

and the category assigned to $x$ is

$$\operatorname{argmax}_{1 \leq k \leq P} \langle s_k, Cf_{\mathbf{z}, \gamma}(x) \rangle_{\mathcal{Y}}. \tag{17}$$

**Remark 5** *We give the multi-class and multi-view learning interpretations and provide numerical experiments for $\mathcal{Y} = \mathbb{R}^{P-1}$, $\mathcal{W} = \mathcal{Y}^m = \mathbb{R}^{(P-1)m}$, with $S$ being the simplex coding. However, we wish to emphasize that optimization problems (14) and (18) and Theorems 6 and 7 are formulated for $\mathcal{W}$ and $\mathcal{Y}$ being arbitrary separable Hilbert spaces.*

### 3.3.1 SOLUTION OF THE SOFT-MARGIN MULTI-VIEW SVM

Introducing slack variables $\xi_{ki}$'s into the optimization problem (14), we obtain the minimization problem

$$f_{\mathbf{z}, \gamma} = \operatorname{argmin}_{f \in \mathcal{H}_K, \xi_{ki} \in \mathbb{R}} \frac{1}{l} \sum_{i=1}^{l} \sum_{k=1, k \neq y_i}^{P} \xi_{ki} + \gamma_A ||f||_{\mathcal{H}_K}^2 + \gamma_I \langle \mathbf{f}, M\mathbf{f} \rangle_{\mathcal{W}^{u+l}}, \tag{18}$$

subject to the constraints

$$\xi_{ki} \geq -\langle s_k, s_{y_i} \rangle_{\mathcal{Y}} + \langle s_k, Cf(x_i) \rangle_{\mathcal{Y}}, \quad 1 \leq i \leq l, k \neq y_i, \tag{19}$$

$$\xi_{ki} \geq 0, \quad 1 \leq i \leq l, k \neq y_i. \tag{20}$$

Let $\alpha_i = (\alpha_{1i}, \ldots, \alpha_{Pi})^T \in \mathbb{R}^P$ as a column vector, with $\alpha_{y_i, i} = 0$. Let $\alpha = (\alpha_1, \ldots, \alpha_l) \in \mathbb{R}^{P \times l}$ as a matrix of size $P \times l$.

**Theorem 6** *The minimization problem (18) has a unique solution given by*

$$f_{\mathbf{z}, \gamma}(x) = \sum_{i=1}^{u+l} K(x, x_i) a_i, \quad a_i \in \mathcal{W}, 1 \leq i \leq u+l, \tag{21}$$

*with $\mathbf{a} = (a_1, \ldots, a_{u+l}) \in \mathcal{W}^{u+l}$ given by*

$$\mathbf{a} = -\frac{1}{2} (\gamma_I M K[\mathbf{x}] + \gamma_A I)^{-1} (I_{(u+l) \times l} \otimes C^* S) \operatorname{vec}(\alpha^{\text{opt}}), \tag{22}$$

*where $\otimes$ denotes the Kronecker tensor product, $K[\mathbf{x}]$ is the $(u+l) \times (u+l)$ operator-valued matrix, with entry $K[\mathbf{x}]_{ij}$ being the operator $K(x_i, x_j) : \mathcal{W} \to \mathcal{W}$, $I_{(u+l) \times l}$ is the $(u+l) \times l$*

matrix of the form $I_{(u+l)\times l} = [I_l \ 0_{l\times u}]^T$, and $\alpha^{\mathrm{opt}} = (\alpha_1^{\mathrm{opt}}, \ldots, \alpha_l^{\mathrm{opt}}) \in \mathbb{R}^{P \times l}$ is a solution of the quadratic minimization problem

$$\alpha^{\mathrm{opt}} = \mathrm{argmin}_{\alpha \in \mathbb{R}^{P\times l}} \frac{1}{4}\mathrm{vec}(\alpha)^T Q[\mathbf{x}, C]\mathrm{vec}(\alpha) + \sum_{i=1}^{l}\sum_{k=1}^{P} \langle s_k, s_{y_i}\rangle_{\mathcal{Y}}\alpha_{ki}, \tag{23}$$

subject to the constraints

$$0 \leq \alpha_{ki} \leq \frac{1}{l}(1 - \delta_{k,y_i}), \quad 1 \leq i \leq l, 1 \leq k \leq P. \tag{24}$$

The symmetric, positive semidefinite, $Pl \times Pl$ matrix $Q[\mathbf{x}, C]$ is given by

$$Q[\mathbf{x}, C] = (I_{(u+l)\times l}^T \otimes S^*C)K[\mathbf{x}](\gamma_I M K[\mathbf{x}] + \gamma_A I)^{-1}(I_{(u+l)\times l} \otimes C^*S). \tag{25}$$

If $S$ is the simplex coding, then

$$\alpha^{\mathrm{opt}} = \mathrm{argmin}_{\alpha \in \mathbb{R}^{P\times l}} \frac{1}{4}\mathrm{vec}(\alpha)^T Q[\mathbf{x}, C]\mathrm{vec}(\alpha) - \frac{1}{P-1}\mathbf{1}_{Pl}^T\mathrm{vec}(\alpha), \tag{26}$$

with $\mathbf{1}_{Pl} = (1, \ldots, 1)^T \in \mathbb{R}^{Pl}$, under the same constraints.

**Special case: Simplex Cone Support Vector Machine (Mroueh et al., 2012)**. For $u = 0$, $\gamma_I = 0$, $\mathcal{W} = \mathcal{Y} = \mathbb{R}^{P-1}$, $C = I_{P-1}$ (single-view), we obtain

$$\mathbf{a} = -\frac{1}{2\gamma_A}(I_l \otimes S)\mathrm{vec}(\alpha^{\mathrm{opt}}), \tag{27}$$

$$Q[\mathbf{x}, C] = \frac{1}{\gamma_A}(I_l \otimes S^*)K[\mathbf{x}](I_l \otimes S), \tag{28}$$

If $S$ is the simplex coding, these together give us the quadratic optimization problem for the Simplex Cone Support Vector Machine (SC-SVM) of (Mroueh et al., 2012).

### 3.3.2 An Equivalent Formulation

For $P = 2$, the simplex coding is $S = [1, -1]$. With this choice of $S$ and $\mathcal{W} = \mathcal{Y} = \mathbb{R}$, $C = 1$, our formulation reduces to single-view binary SVM with manifold regularization, which is precisely the Laplacian SVM of (Belkin et al., 2006). In this section, we give an equivalent result to Theorem 6, namely Theorem 7 below, which includes the formulation of the Laplacian SVM as a special case.

Let $S_{\hat{y}_i}$ be the matrix obtained from $S$ by removing the $y_i$th column and $\beta_i \in \mathbb{R}^{P-1}$ be the vector obtained from $\alpha_i$ by deleting the $y_i$th entry, which is equal to zero by assumption. As a linear operator, $S_{\hat{y}_i} : \mathbb{R}^{P-1} \to \mathcal{Y}$ and

$$S\alpha_i = \sum_{k=1, k\neq y_i}^{P} \alpha_{ki}s_k = S_{\hat{y}_i}\beta_i. \tag{29}$$

Let $\mathrm{diag}(S_{\hat{\mathbf{y}}})$ be the $l \times l$ block diagonal matrix, with block $(i, i)$ being $S_{\hat{y}_i}$ and $\beta = (\beta_1, \ldots, \beta_l)$ be the $(P-1) \times l$ matrix with column $i$ being $\beta_i$. As a linear operator, $\mathrm{diag}(S_{\hat{\mathbf{y}}}) : \mathbb{R}^{(P-1)l} \to \mathcal{Y}^l$.

**Theorem 7** *The minimization problem (18) has a unique solution given by* $f_{\mathbf{z},\gamma}(x) = \sum_{i=1}^{u+l} K(x, x_i) a_i$, *with* $\mathbf{a} = (a_1, \ldots, a_{u+l}) \in \mathcal{W}^{u+l}$ *given by*

$$\mathbf{a} = -\frac{1}{2}(\gamma_I MK[\mathbf{x}] + \gamma_A I)^{-1}(I_{(u+l) \times l} \otimes C^*) \text{diag}(S_{\hat{\mathbf{y}}}) \text{vec}(\beta^{\text{opt}}), \tag{30}$$

*where* $\beta^{\text{opt}} = (\beta_1^{\text{opt}}, \ldots, \beta_l^{\text{opt}}) \in \mathbb{R}^{(P-1) \times l}$ *is a solution of the quadratic minimization problem*

$$\beta^{\text{opt}} = \text{argmin}_{\beta \in \mathbb{R}^{(P-1) \times l}} \frac{1}{4} \text{vec}(\beta)^T Q[\mathbf{x}, \mathbf{y}, C] \text{vec}(\beta) + \sum_{i=1}^{l} \langle s_{y_i}, S_{\hat{y}_i} \beta_i \rangle_{\mathcal{Y}}, \tag{31}$$

*subject to the constraints*

$$0 \le \beta_{ki} \le \frac{1}{l}, \quad 1 \le i \le l, 1 \le k \le P-1. \tag{32}$$

*The symmetric, positive semidefinite,* $(P-1)l \times (P-1)l$ *matrix* $Q[\mathbf{x}, \mathbf{y}, C]$ *is given by*

$$Q[\mathbf{x}, \mathbf{y}, C] = \text{diag}(S_{\hat{\mathbf{y}}}^*)(I_{(u+l) \times l}^T \otimes C) K[\mathbf{x}](\gamma_I MK[\mathbf{x}] + \gamma_A I)^{-1}(I_{(u+l) \times l} \otimes C^*) \text{diag}(S_{\hat{\mathbf{y}}}). \tag{33}$$

*If* $S$ *is the simplex coding, then, under the same constraints,*

$$\beta^{\text{opt}} = \text{argmin}_{\beta \in \mathbb{R}^{(P-1) \times l}} \frac{1}{4} \text{vec}(\beta)^T Q[\mathbf{x}, \mathbf{y}, C] \text{vec}(\beta) - \frac{1}{P-1} \mathbf{1}_{(P-1)l}^T \text{vec}(\beta). \tag{34}$$

It is straightforward to switch between $\alpha$ and $\beta$. Let $I_{P, \hat{y}_i}$ be the $P \times (P-1)$ matrix obtained by removing the $y_i$ column from the $P \times P$ identity matrix, then

$$\alpha_i = I_{P, \hat{y}_i} \beta_i \quad \text{and} \quad \beta_i = I_{P, \hat{y}_i}^T \alpha_i. \tag{35}$$

**Binary case with simplex coding**. For $P = 2$, we represent the discrete output label set $\text{cl}(\mathcal{Y})$ as $\text{cl}(\mathcal{Y}) = \{\pm 1\}$. In this case, $\beta$ is simply a vector in $\mathbb{R}^l$, and we solve the optimization problem

$$\beta^{\text{opt}} = \text{argmin}_{\beta \in \mathbb{R}^l} \frac{1}{4} \beta^T Q[\mathbf{x}, \mathbf{y}, C] \beta - \mathbf{1}_l^T \beta, \tag{36}$$

subject to the constraints $0 \le \beta_i \le \frac{1}{l}$, $1 \le i \le l$. The binary simplex code is $S = [1, -1]$, with $S_{\hat{1}} = -1$ and $S_{-\hat{1}} = 1$. Thus $S_{\hat{y}_i} = -y_i$. Furthermore, because $\mathcal{Y} = \mathbb{R}$, by the Riesz representation theorem, the bounded linear operator $C : \mathcal{W} \to \mathbb{R}$ and its adjoint $C^* : \mathbb{R} \to \mathcal{W}$ necessarily have the form

$$Cf(x) = \langle \mathbf{c}, f(x) \rangle_{\mathcal{W}} \quad \text{and} \quad C^* y = y\mathbf{c}, \tag{37}$$

respectively, for a unique vector $\mathbf{c} \in \mathcal{W}$. It follows immediately that

**Corollary 8 (Binary case)** *Let* $S$ *be the simplex coding and* $P = 2$. *Then in Theorem 7,*

$$\mathbf{a} = \frac{1}{2}(\gamma_I MK[\mathbf{x}] + \gamma_A I)^{-1}(I_{(u+l) \times l} \otimes \mathbf{c}) \text{diag}(\mathbf{y})(\beta^{\text{opt}}), \tag{38}$$

$$Q[\mathbf{x}, \mathbf{y}, C] = \text{diag}(\mathbf{y})(I_{(u+l) \times l}^T \otimes \mathbf{c}^T) K[\mathbf{x}](\gamma_I MK[\mathbf{x}] + \gamma_A I)^{-1}(I_{(u+l) \times l} \otimes \mathbf{c}) \text{diag}(\mathbf{y}). \tag{39}$$

**Special case: Laplacian SVM (Belkin et al., 2006)**. In (38) and (39), by setting $\mathbf{c} = 1$ ($\mathcal{W} = \mathcal{Y} = \mathbb{R}$) (single-view) and $M$ to be the graph Laplacian on the training data $\{x_i\}_{i=1}^{u+l}$, we obtain the Laplacian SVM of (Belkin et al., 2006).

### 3.4 Previous Work as Special Cases of the Current Framework

We have shown above that in the SVM case, our framework includes the multi-class, supervised Simplex Cone SVM of (Mroueh et al., 2012) and the binary, semi-supervised Laplacian SVM of (Belkin et al., 2006) as special cases. Before delving into concrete implementations, in this section we give a list of other common kernel-based learning algorithms which are special cases of our learning framework.

**Vector-valued Regularized Least Squares.** If $\mathbf{C}^*\mathbf{C} = I : \mathcal{W}^{u+l} \to \mathcal{W}^{u+l}$, then (13) reduces to

$$(J_l^{\mathcal{W},u+l}K[\mathbf{x}] + l\gamma_I MK[\mathbf{x}] + l\gamma_A I)\mathbf{a} = \mathbf{C}^*\mathbf{y}. \tag{40}$$

If $u = 0$, $\gamma_I = 0$, and $\gamma_A = \gamma$, then we have

$$(K[\mathbf{x}] + l\gamma I)\mathbf{a} = \mathbf{C}^*\mathbf{y}. \tag{41}$$

One particular case for this scenario is when $\mathcal{W} = \mathcal{Y}$ and $C : \mathcal{Y} \to \mathcal{Y}$ is a unitary operator, that is $C^*C = CC^* = I$. If $\mathcal{Y} = \mathbb{R}^n$ and $C : \mathbb{R}^n \to \mathbb{R}^n$ is real, then $C$ is an orthogonal matrix. If $C = I$, then we recover the vector-valued Regularized Least Squares algorithm (Micchelli and Pontil, 2005).

**Vector-valued Manifold Regularization.** Let $\mathcal{W} = \mathcal{Y}$ and $C = I$. Then we obtain the minimization problem for vector-valued Manifold Regularization (Minh and Sindhwani, 2011):

$$f_{\mathbf{z},\gamma} = \mathrm{argmin}_{f \in \mathcal{H}_K} \frac{1}{l}\sum_{i=1}^{l} V(y_i, f(x_i)) + \gamma_A \|f\|_{\mathcal{H}_K}^2 + \gamma_I \langle \mathbf{f}, M\mathbf{f}\rangle_{\mathcal{W}_{u+l}}. \tag{42}$$

**Scalar Multi-view Learning.** Let us show that the scalar multi-view learning formulation of (Sindhwani and Rosenberg, 2008; Rosenberg et al., 2009) can be cast as a special case of our framework. Let $\mathcal{Y} = \mathbb{R}$ and $k^1, \ldots, k^m$ be real-valued positive definite kernels on $\mathcal{X} \times \mathcal{X}$, with corresponding RKHS $\mathcal{H}_{k^i}$ of functions $f^i : \mathcal{X} \to \mathbb{R}$, with each $\mathcal{H}_{k^i}$ representing one view. Let $f = (f^1, \ldots, f^m)$, with $f^i \in \mathcal{H}_{k^i}$. Let $\mathbf{c} = (c_1, \ldots, c_m) \in \mathbb{R}^m$ be a fixed weight vector. In the notation of (Rosenberg et al., 2009), let

$$\mathbf{f} = (f^1(x_1), \ldots, f^1(x_{u+l}), \ldots, f^m(x_1), \ldots, f^m(x_{u+l}))$$

and $M \in \mathbb{R}^{m(u+l) \times m(u+l)}$ be positive semidefinite. The objective of Multi-view Point Cloud Regularization (formula (4) in (Rosenberg et al., 2009)) is

$$\mathrm{argmin}_{\varphi:\varphi(x)=\langle \mathbf{c}, f(x)\rangle} \frac{1}{l}\sum_{i=1}^{l} V(y_i, \varphi(x_i)) + \sum_{i=1}^{m} \gamma_i \|f^i\|_{k^i}^2 + \gamma\langle \mathbf{f}, M\mathbf{f}\rangle_{\mathbb{R}^{m(u+l)}}, \tag{43}$$

for some convex loss function $V$, with $\gamma_i > 0$, $i = 1, \ldots, m$, and $\gamma \geq 0$. Problem (43) admits a natural formulation in vector-valued RKHS. Let

$$K = \mathrm{diag}(\frac{1}{\gamma_1}, \ldots, \frac{1}{\gamma_m}) * \mathrm{diag}(k^1, \ldots, k^m) : \mathcal{X} \times \mathcal{X} \to \mathbb{R}^{m \times m}, \tag{44}$$

then $f = (f^1, \ldots, f^m) \in \mathcal{H}_K : \mathcal{X} \to \mathbb{R}^m$, with

$$\|f\|_{\mathcal{H}_K}^2 = \sum_{i=1}^{m} \gamma_i \|f^i\|_{k^i}^2. \tag{45}$$

By the reproducing property, we have

$$\langle \mathbf{c}, f(x) \rangle_{\mathbb{R}^m} = \langle f, K_x \mathbf{c} \rangle_{\mathcal{H}_K}. \tag{46}$$

We can now recast (43) into

$$f_{\mathbf{z}, \gamma} = \operatorname{argmin}_{f \in \mathcal{H}_K} \frac{1}{l} \sum_{i=1}^{l} V(y_i, \langle \mathbf{c}, f(x) \rangle_{\mathbb{R}^m}) + ||f||_{\mathcal{H}_K}^2 + \gamma \langle \mathbf{f}, M\mathbf{f} \rangle_{\mathbb{R}^{m(u+l)}}. \tag{47}$$

This is a special case of (9), with $\mathcal{W} = \mathbb{R}^m$, $\mathcal{Y} = \mathbb{R}$, and $C : \mathbb{R}^m \to \mathbb{R}$ given by

$$Cf(x) = \langle \mathbf{c}, f(x) \rangle_{\mathbb{R}^m} = c_1 f^1(x) + \cdots + c_m f^m(x). \tag{48}$$

The vector-valued formulation of scalar multi-view learning has the following advantages:

(i) The kernel $K$ is diagonal matrix-valued and is obviously positive definite. In contrast, it is nontrivial to prove that the multi-view kernel of (Rosenberg et al., 2009) is positive definite.

(ii) The kernel $K$ is independent of the $c_i$'s, unlike the multi-view kernel of (Rosenberg et al., 2009), which needs to be recomputed for each different set $c_i$'s.

(iii) One can recover all the component functions $f^i$'s using $K$. In contrast, in (Sindhwani and Rosenberg, 2008), it is shown how one can recover the $f^i$'s only when $m = 2$, but not in the general case.

## 4. Vector-valued Multi-view Learning

In this and subsequent sections, we focus on a special case of our formulation, namely vector-valued multi-view learning. For a general separable Hilbert space $\mathcal{Y}$, let $\mathcal{W} = \mathcal{Y}^m$ and $C_1, \ldots, C_m : \mathcal{Y} \to \mathcal{Y}$ be bounded linear operators. For $f(x) = (f^1(x), \ldots, f^m(x))$, with each $f^i(x) \in \mathcal{Y}$, we define the combination operator $C = [C_1, \ldots, C_m] : \mathcal{Y}^m \to \mathcal{Y}$ by

$$Cf(x) = C_1 f^1(x) + \cdots + C_m f^m(x) \in \mathcal{Y}. \tag{49}$$

This gives rise to a vector-valued version of multi-view learning, where outputs from $m$ views, each one being a vector in the Hilbert space $\mathcal{Y}$, are linearly combined. In the following, we give concrete definitions of both the combination operator $C$ and the multi-view manifold regularization term $M$ for our multi-view learning model.

### 4.1 The Combination Operator

In the present context, the bounded linear operator $C : \mathcal{W} \to \mathcal{Y}$ is a (potentially infinite) matrix of size $\dim(\mathcal{Y}) \times m \dim(\mathcal{Y})$. This operator transforms the output vectors obtained from the $m$ views $f^i$'s in $\mathcal{Y}^m$ into an output vector in $\mathcal{Y}$. The simplest form of $C$ is the average operator:

$$Cf(x) = \frac{1}{m}(f^1(x) + \cdots + f^m(x)) \in \mathcal{Y}. \tag{50}$$

Let $\otimes$ denote the Kronecker tensor product. For $m \in \mathbb{N}$, let $\mathbf{1}_m = (1, \ldots, 1)^T \in \mathbb{R}^m$. The matrix $C$ is then

$$C = \frac{1}{m} \mathbf{1}_m^T \otimes I_{\mathcal{Y}} = \frac{1}{m} [I_{\mathcal{Y}}, \ldots, I_{\mathcal{Y}}]. \tag{51}$$

More generally, we consider a weight vector $\mathbf{c} = (c_1, \ldots, c_m)^T \in \mathbb{R}^m$ and define $C$ as

$$C = \mathbf{c}^T \otimes I_{\mathcal{Y}}, \text{ with } Cf(x) = \sum_{i=1}^{m} c_i f^i(x) \in \mathcal{Y}. \tag{52}$$

### 4.2 Multi-view Manifold Regularization

Generalizing (Minh et al., 2013), we decompose the multi-view manifold regularization term $\gamma_I \langle \mathbf{f}, M\mathbf{f} \rangle_{\mathcal{W}^{u+l}}$ in (Eq. 9) into two components

$$\gamma_I \langle \mathbf{f}, M\mathbf{f} \rangle_{\mathcal{W}^{u+l}} = \gamma_B \langle \mathbf{f}, M_B \mathbf{f} \rangle_{\mathcal{W}^{u+l}} + \gamma_W \langle \mathbf{f}, M_W \mathbf{f} \rangle_{\mathcal{W}^{u+l}}, \tag{53}$$

where $M_B, M_W : \mathcal{W}^{u+l} \rightarrow \mathcal{W}^{u+l}$ are symmetric, positive operators, and $\gamma_B, \gamma_W \geq 0$. We call the first term *between-view regularization*, which measures the consistency of the component functions across different views, and the second term *within-view regularization*, which measures the smoothness of the component functions in their corresponding views. We describe next two concrete choices for $M_B$ and $M_W$.

**Between-view Regularization.** Let

$$M_m = mI_m - \mathbf{1}_m \mathbf{1}_m^T. \tag{54}$$

This is the $m \times m$ matrix with $(m-1)$ on the diagonal and $-1$ elsewhere. Then for $\mathbf{a} = (a_1, \ldots, a_m) \in \mathbb{R}^m$,

$$\mathbf{a}^T M_m \mathbf{a} = \sum_{j,k=1, j<k}^{m} (a_j - a_k)^2. \tag{55}$$

If each $a_i \in \mathcal{Y}$, then we have $\mathbf{a} \in \mathcal{Y}^m$ and

$$\mathbf{a}^T (M_m \otimes I_{\mathcal{Y}}) \mathbf{a} = \sum_{j,k=1, j<k}^{m} ||a_j - a_k||_{\mathcal{Y}}^2. \tag{56}$$

We define $M_B$ by

$$M_B = I_{u+l} \otimes (M_m \otimes I_{\mathcal{Y}}). \tag{57}$$

Then $M_B$ is a diagonal block matrix of size $m(u+l) \dim(\mathcal{Y}) \times m(u+l) \dim(\mathcal{Y})$, with each block $(i, i)$ being $M_m \otimes I_{\mathcal{Y}}$. For $\mathbf{f} = (f(x_1), \ldots, f(x_{u+l})) \in \mathcal{Y}^{m(u+l)}$, with $f(x_i) \in \mathcal{Y}^m$,

$$\langle \mathbf{f}, M_B \mathbf{f} \rangle_{\mathcal{Y}^{m(u+l)}} = \sum_{i=1}^{u+l} \langle f(x_i), (M_m \otimes I_{\mathcal{Y}}) f(x_i) \rangle_{\mathcal{Y}^m} = \sum_{i=1}^{u+l} \sum_{j,k=1, j<k}^{m} ||f^j(x_i) - f^k(x_i)||_{\mathcal{Y}}^2. \tag{58}$$

This term thus enforces the consistency between the different components $f^i$'s which represent the outputs on the different views. For $\mathcal{Y} = \mathbb{R}$, this is precisely the Point Cloud regularization term for scalar multi-view learning (Rosenberg et al., 2009; Brefeld et al., 2006). In particular, for $m = 2$, we have $M_2 = \begin{pmatrix} 1 & -1 \\ -1 & 1 \end{pmatrix}$, and

$$\langle \mathbf{f}, M_B \mathbf{f} \rangle_{\mathbb{R}^{2(u+l)}} = \sum_{i=1}^{u+l} (f^1(x_i) - f^2(x_i))^2, \tag{59}$$

which is the Point Cloud regularization term for co-regularization (Sindhwani and Rosenberg, 2008).

**Within-view Regularization.** One way to define $M_W$ is via the graph Laplacian. For view $i$, $1 \leq i \leq m$, let $G^i$ be a corresponding undirected graph, with symmetric, nonnegative weight matrix $W^i$, which induces the scalar graph Laplacian $L^i$, a matrix of size $(u+l) \times (u+l)$. For a vector $\mathbf{a} \in \mathbb{R}^{u+l}$, we have

$$\mathbf{a}^T L^i \mathbf{a} = \sum_{j,k=1,j<k}^{u+l} W_{jk}^i (a_j - a_k)^2.$$

Let $L$ be the block matrix of size $(u+l) \times (u+l)$, with block $(i,j)$ being the $m \times m$ diagonal matrix given by

$$L_{i,j} = \text{diag}(L_{ij}^1, \ldots L_{ij}^m). \tag{60}$$

Then for $\mathbf{a} = (a_1, \ldots, a_{u+l})$, with $a_j \in \mathbb{R}^m$, we have

$$\mathbf{a}^T L \mathbf{a} = \sum_{i=1}^m \sum_{j,k=1,j<k}^{u+l} W_{jk}^i (a_j^i - a_k^i)^2. \tag{61}$$

If $a_j \in \mathcal{Y}^m$, with $a_j^i \in \mathcal{Y}$, then

$$\mathbf{a}^T (L \otimes I_{\mathcal{Y}}) \mathbf{a} = \sum_{i=1}^m \sum_{j,k=1,j<k}^{u+l} W_{jk}^i ||a_j^i - a_k^i||_{\mathcal{Y}}^2. \tag{62}$$

Define

$$M_W = L \otimes I_{\mathcal{Y}}, \quad \text{then} \tag{63}$$

$$\langle \mathbf{f}, M_W \mathbf{f} \rangle_{\mathcal{Y}^{m(u+l)}} = \sum_{i=1}^m \sum_{j,k=1,j<k}^{u+l} W_{jk}^i ||f^i(x_j) - f^i(x_k)||_{\mathcal{Y}}^2. \tag{64}$$

The $i$th summand in the sum $\sum_{i=1}^m$ is precisely a manifold regularization term within view $i$. This term thus enforces the consistency of the output along each view $i$, $1 \leq i \leq m$.

**Single View Case.** When $m = 1$, we have $M_m = 0$ and therefore $M_B = 0$. In this case, we simply carry out manifold regularization within the given single view, using $M_W$.

## 5. Numerical Implementation

In this section, we give concrete forms of Theorems 4, for vector-valued multi-view least squares regression, and Theorem 6, for vector-valued multi-view SVM, that can be efficiently implemented. For our present purposes, let $m \in \mathbb{N}$ be the number of views and $\mathcal{W} = \mathcal{Y}^m$. Consider the case $\dim(\mathcal{Y}) < \infty$. Without loss of generality, we set $\mathcal{Y} = \mathbb{R}^{\dim(\mathcal{Y})}$.

**The Kernel**. For the current implementations, we define the kernel $K(x,t)$ by

$$K(x,t) = G(x,t) \otimes R, \tag{65}$$

where $G : \mathcal{X} \times \mathcal{X} \to \mathbb{R}^{m \times m}$ is a matrix-valued positive definite kernel, with $G(x,t)$ being an $m \times m$ matrix for each pair $(x,t) \in \mathcal{X} \times \mathcal{X}$. A concrete example of $G$, which we use in our experiments, is given in Section 7. The bounded linear operator $R : \mathcal{Y} \to \mathcal{Y}$ is symmetric and positive, and when $\dim(\mathcal{Y}) < \infty$, $R$ is a symmetric, positive semi-definite matrix of size $\dim(\mathcal{Y}) \times \dim(\mathcal{Y})$. The Gram matrices of $K$ and $G$ are block matrices $K[\mathbf{x}]$ and $G[\mathbf{x}]$, respectively, of size $(u+l) \times (u+l)$, with blocks $(i,j)$ given by $(K[\mathbf{x}])_{ij} = K(x_i, x_j)$ and $(G[\mathbf{x}])_{ij} = G(x_i, x_j)$. They are related by

$$K[\mathbf{x}] = G[\mathbf{x}] \otimes R. \tag{66}$$

**Lemma 9** *The matrix-valued kernel $K$ is positive definite.*

## 5.1 Numerical Implementation for Vector-valued Multi-view Least Squares

With the kernel $K$ as defined in (65) and $C$ and $M$ as defined in Section 4, the system of linear equations (13) in Theorem 4 becomes a Sylvester equation, which can be solved efficiently, as follows.

**Theorem 10** *For $C = \mathbf{c}^T \otimes I_{\mathcal{Y}}$, $\mathbf{c} \in \mathbb{R}^m$, $M_W = L \otimes I_{\mathcal{Y}}$, $M_B = I_{u+l} \otimes (M_m \otimes I_{\mathcal{Y}})$, and the kernel $K$ as defined in (65) the system of linear equations (13) in Theorem 4 is equivalent to the Sylvester equation*

$$BAR + l\gamma_A A = Y_C, \tag{67}$$

*where*

$$B = \left( (J_l^{u+l} \otimes \mathbf{c}\mathbf{c}^T) + l\gamma_B(I_{u+l} \otimes M_m) + l\gamma_W L \right) G[\mathbf{x}], \tag{68}$$

*which is of size $(u+l)m \times (u+l)m$, $A$ is the matrix of size $(u+l)m \times \dim(\mathcal{Y})$ such that $\mathbf{a} = \mathrm{vec}(A^T)$, and $Y_C$ is the matrix of size $(u+l)m \times \dim(\mathcal{Y})$ such that $\mathbf{C}^*\mathbf{y} = \mathrm{vec}(Y_C^T)$. $J_l^{u+l} : \mathbb{R}^{u+l} \to \mathbb{R}^{u+l}$ is a diagonal matrix of size $(u+l) \times (u+l)$, with the first $l$ entries on the main diagonal being $1$ and the rest being $0$.*

**Special cases**: For $m = 1$, $\mathbf{c} = 1$, Equation (67) reduces to Equation 17 of (Minh and Sindhwani, 2011). For $R = I_{\mathcal{Y}}$, with $\mathcal{Y} = \mathbb{R}^P$, Equation (67) reduces to Equation 43 in (Minh et al., 2013).

**Evaluation on a testing sample:** Having solved for the matrix $A$, and hence the vector $\mathbf{a}$ in Theorem 10, we next show how the resulting functions can be efficiently evaluated on a testing set. Let $\mathbf{v} = \{v_1, \ldots, v_t\} \in \mathcal{X}$ be an arbitrary set of testing input examples, with $t \in \mathbb{N}$. Let $\mathbf{f}_{\mathbf{z},\gamma}(\mathbf{v}) = (\{f_{\mathbf{z},\gamma}(v_1), \ldots, f_{\mathbf{z},\gamma}(v_t)\})^T \in \mathcal{Y}^{mt}$, with

$$f_{\mathbf{z},\gamma}(v_i) = \sum_{j=1}^{u+l} K(v_i, x_j) a_j.$$

Let $K[\mathbf{v}, \mathbf{x}]$ denote the $t \times (u+l)$ block matrix, where block $(i,j)$ is $K(v_i, x_j)$ and similarly, let $G[\mathbf{v}, \mathbf{x}]$ denote the $t \times (u+l)$ block matrix, where block $(i,j)$ is the $m \times m$ matrix $G(v_i, x_j)$. Then

$$\mathbf{f}_{\mathbf{z},\gamma}(\mathbf{v}) = K[\mathbf{v}, \mathbf{x}]\mathbf{a} = (G[\mathbf{v}, \mathbf{x}] \otimes R)\mathbf{a} = \mathrm{vec}(RA^T G[\mathbf{v}, \mathbf{x}]^T),$$

---

**Algorithm 1** $\mathcal{Y}$-valued, $m$-view, semi-supervised least square regression and classification

---

*This algorithm implements and evaluates the solution of Theorem 10.*

**Input**:
- Training data $\mathbf{z} = \{(x_i, y_i)\}_{i=1}^{l} \cup \{x_i\}_{i=l+1}^{u+l}$, with $l$ labeled and $u$ unlabeled examples.
- Number of views: $m$.
- Output values: vectors in $\mathcal{Y}$.
- Testing example: $v$.

**Parameters**:
- The regularization parameters $\gamma_A, \gamma_B, \gamma_W$.
- The weight vector $\mathbf{c}$.
- A matrix-valued kernel $G$, with $G(x, t)$ being an $m \times m$ matrix for each pair $(x, t)$.

**Procedure**:
- Compute kernel matrix $G[\mathbf{x}]$ on input set $\mathbf{x} = (x_i)_{i=1}^{u+l}$.
- Compute matrix $C$ according to (52).
- Compute graph Laplacian $L$ according to (60).
- Compute matrices $B, Y_C$ according to Theorem 10.
- Solve matrix equation $BAR + l\gamma_A A = Y_C$ for $A$.
- Compute kernel matrix $G[v, \mathbf{x}]$ between $v$ and $\mathbf{x}$.

**Output**: $f_{\mathbf{z}, \gamma}(v) = \text{vec}(RA^T G[v, \mathbf{x}]^T) \in \mathcal{Y}^m$.
$\mathcal{Y}$-valued regression: return $C f_{\mathbf{z}, \gamma}(v) \in \mathcal{Y}$.
Multi-class classification: return index of $\max(C f_{\mathbf{z}, \gamma}(v))$.

---

In particular, for $\mathbf{v} = \mathbf{x} = (x_i)_{i=1}^{u+l}$, the original training sample, we have $G[\mathbf{v}, \mathbf{x}] = G[\mathbf{x}]$.

**Algorithm**: All the necessary steps for implementing Theorem 10 and evaluating its solution are summarized in Algorithm 1. For $P$-class classification, $\mathcal{Y} = \mathbb{R}^P$, and $y_i = (-1, \ldots, 1, \ldots, -1)$, $1 \leq i \leq l$, with 1 at the $k$th location if $x_i$ is in the $k$th class.

## 5.2 Numerical Implementation for Vector-valued Multi-view SVM

This section gives a concrete form of Theorem 6 for vector-valued multi-view SVM which can be efficiently implemented. Let $\{\lambda_{i,R}\}_{i=1}^{\dim(\mathcal{Y})}$ be the eigenvalues of $R$, which are all nonnegative, with corresponding orthonormal eigenvectors $\{\mathbf{r}_i\}_{i=1}^{\dim(\mathcal{Y})}$. Then $R$ admits the orthogonal spectral decomposition

$$R = \sum_{i=1}^{\dim(\mathcal{Y})} \lambda_{i,R} \mathbf{r}_i \mathbf{r}_i^T. \tag{69}$$

Under this representation of $R$ and with the kernel $K$ as defined in (65), Theorem 6 takes the following concrete form.

**Theorem 11** *Let* $\gamma_I M = \gamma_B M_B + \gamma_W M_W$, $C = \mathbf{c}^T \otimes I_{\mathcal{Y}}$, *and* $K(x,t)$ *be defined as in (65). Then in Theorem 6,*

$$\mathbf{a} = -\frac{1}{2}[\sum_{i=1}^{\dim(\mathcal{Y})} M_{\text{reg}}^i (I_{(u+l)\times l} \otimes \mathbf{c}) \otimes \mathbf{r}_i \mathbf{r}_i^T S]\text{vec}(\alpha^{\text{opt}}), \tag{70}$$

$$Q[\mathbf{x}, C] = \sum_{i=1}^{\dim(\mathcal{Y})} (I_{(u+l)\times l}^T \otimes \mathbf{c}^T) G[\mathbf{x}] M_{\text{reg}}^i (I_{(u+l)\times l} \otimes \mathbf{c}) \otimes \lambda_{i,R} S^* \mathbf{r}_i \mathbf{r}_i^T S, \tag{71}$$

*where*

$$M_{\text{reg}}^i = [\lambda_{i,R}(\gamma_B I_{u+l} \otimes M_m + \gamma_W L) G[\mathbf{x}] + \gamma_A I_{m(u+l)}]^{-1}. \tag{72}$$

**Evaluation phase**: Having solved for $\alpha^{\text{opt}}$ and hence $\mathbf{a}$ in Theorem 11, we next show how the resulting functions can be efficiently evaluated on a testing set $\mathbf{v} = \{v_i\}_{i=1}^t \subset \mathcal{X}$.

**Proposition 12** *Let* $f_{\mathbf{z},\gamma}$ *be the solution obtained in Theorem 11. For any example* $v \in \mathcal{X}$,

$$f_{\mathbf{z},\gamma}(v) = -\frac{1}{2}\text{vec}[\sum_{i=1}^{\dim(\mathcal{Y})} \lambda_{i,R} \mathbf{r}_i \mathbf{r}_i^T S\alpha^{\text{opt}}(I_{(u+l)\times l}^T \otimes \mathbf{c}^T)(M_{\text{reg}}^i)^T G[v, \mathbf{x}]^T]. \tag{73}$$

*The combined function, using the combination operator* $C$, *is* $g_{\mathbf{z},\gamma}(v) = C f_{\mathbf{z},\gamma}(v)$ *and is given by*

$$g_{\mathbf{z},\gamma}(v) = -\frac{1}{2}\sum_{i=1}^{\dim(\mathcal{Y})} \lambda_{i,R} \mathbf{r}_i \mathbf{r}_i^T S\alpha^{\text{opt}}(I_{(u+l)\times l}^T \otimes \mathbf{c}^T)(M_{\text{reg}}^i)^T G[v, \mathbf{x}]^T \mathbf{c}. \tag{74}$$

*The final SVM decision function is* $h_{\mathbf{z},\gamma}(v) = S^T g_{\mathbf{z},\gamma}(v) \in \mathbb{R}^P$ *and is given by*

$$h_{\mathbf{z},\gamma}(v) = -\frac{1}{2}\sum_{i=1}^{\dim(\mathcal{Y})} \lambda_{i,R} S^T \mathbf{r}_i \mathbf{r}_i^T S\alpha^{\text{opt}}(I_{(u+l)\times l}^T \otimes \mathbf{c}^T)(M_{\text{reg}}^i)^T G[v, \mathbf{x}]^T \mathbf{c}. \tag{75}$$

*On a testing set* $\mathbf{v} = \{v_i\}_{i=1}^t \subset \mathcal{X}$,

$$h_{\mathbf{z},\gamma}(\mathbf{v}) = -\frac{1}{2}\sum_{i=1}^{\dim(\mathcal{Y})} \lambda_{i,R} S^T \mathbf{r}_i \mathbf{r}_i^T S\alpha^{\text{opt}}(I_{(u+l)\times l}^T \otimes \mathbf{c}^T)(M_{\text{reg}}^i)^T G[\mathbf{v}, \mathbf{x}]^T (I_t \otimes \mathbf{c}), \tag{76}$$

*as a matrix of size* $P \times t$, *with the* $i$th *column being* $h_{\mathbf{z},\gamma}(v_i)$.

**Algorithm**: All the necessary steps for implementing Theorem 11 and Proposition 12 are summarized in Algorithm 2.

**Algorithm 2** Multi-class Multi-view SVM

*This algorithm implements Theorem 11 and Proposition 12. In the case $R = I_{\mathcal{Y}}$, it implements Theorem 13 and Proposition 14, with $M^i_{\text{reg}} = M_{\text{reg}}$ in (79), and equations (71), (73), and (75) are replaced by (78), (80), and (82), respectively.*

**Input**:
- Training data $\mathbf{z} = \{(x_i, y_i)\}^l_{i=1} \cup \{x_i\}^{u+l}_{i=l+1}$, with $l$ labeled and $u$ unlabeled examples.
- Number of classes: $P$. Number of views: $m$.
- Testing example: $v$.
**Parameters**:
- The regularization parameters $\gamma_A, \gamma_B, \gamma_W$.
- The weight vector $\mathbf{c}$.
- A matrix-valued kernel $G$, with $G(x, t)$ being an $m \times m$ matrix for each pair $(x, t)$.
**Procedure**:
- Compute kernel matrices $G[\mathbf{x}]$ on $\mathbf{x} = (x_i)^{u+l}_{i=1}$ and $G[v, \mathbf{x}]$ between $v$ and $\mathbf{x}$.
- Compute graph Laplacian $L$ according to (60).
- Compute matrices $M^i_{\text{reg}}$ according to (72).
- Compute matrix $Q[\mathbf{x}, C]$ according to (71)
- Solve quadratic optimization problem (23) for $\alpha^{\text{opt}}$.
**Output**: $f_{\mathbf{z}, \gamma}(v)$, computed according to (73).
Classification: return $\text{argmax}(h_{\mathbf{z}, \gamma}(v))$, with $h_{\mathbf{z}, \gamma}(v) \in \mathbb{R}^P$ computed according to (75).

### 5.2.1 SPECIAL CASE

Consider the case $R = I_{\mathcal{Y}}$. Then Theorem 11 and Proposition 12 simplify to the following.

**Theorem 13** *Let $\gamma_I M = \gamma_B M_B + \gamma_W M_W$, $C = \mathbf{c}^T \otimes I_{\mathcal{Y}}$, and $K(x, t)$ be defined as in (65) with $R = I_{\mathcal{Y}}$. Then in Theorem 6,*

$$\mathbf{a} = -\frac{1}{2}[M_{\text{reg}}(I_{(u+l) \times l} \otimes \mathbf{c}) \otimes S] \text{vec}(\alpha^{\text{opt}}), \tag{77}$$

*and*

$$Q[\mathbf{x}, C] = (I^T_{(u+l) \times l} \otimes \mathbf{c}^T) G[\mathbf{x}] M_{\text{reg}}(I_{(u+l) \times l} \otimes \mathbf{c}) \otimes S^* S, \tag{78}$$

*where*

$$M_{\text{reg}} = [(\gamma_B I_{u+l} \otimes M_m + \gamma_W L) G[\mathbf{x}] + \gamma_A I_{m(u+l)}]^{-1}. \tag{79}$$

**Proposition 14** *Let $f_{\mathbf{z}, \gamma}$ be the solution obtained in Theorem 13. For any example $v \in \mathcal{X}$,*

$$f_{\mathbf{z}, \gamma}(v) = -\frac{1}{2} \text{vec}(S\alpha^{\text{opt}}(I^T_{(u+l) \times l} \otimes \mathbf{c}^T) M^T_{\text{reg}} G[v, \mathbf{x}]^T). \tag{80}$$

*The combined function, using the combination operator $C$, is $g_{\mathbf{z}, \gamma}(v) = C f_{\mathbf{z}, \gamma}(v) \in \mathbb{R}^{P-1}$ and is given by*

$$g_{\mathbf{z}, \gamma}(v) = -\frac{1}{2} S\alpha^{\text{opt}}(I^T_{(u+l) \times l} \otimes \mathbf{c}^T) M^T_{\text{reg}} G[v, \mathbf{x}]^T \mathbf{c}. \tag{81}$$

*The final SVM decision function is* $h_{\mathbf{z},\gamma}(v) = S^T g_{\mathbf{z},\gamma}(v) \in \mathbb{R}^P$ *and is given by*

$$h_{\mathbf{z},\gamma}(v) = -\frac{1}{2} S^T S \alpha^{\text{opt}} (I^T_{(u+l)\times l} \otimes \mathbf{c}^T) M^T_{\text{reg}} G[v, \mathbf{x}]^T \mathbf{c}. \tag{82}$$

*On a testing set* $\mathbf{v} = \{v_i\}_{i=1}^t$, *let* $h_{\mathbf{z},\gamma}(\mathbf{v}) \in \mathbb{R}^{P \times t}$ *be the matrix with the ith column being* $h_{\mathbf{z},\gamma}(v_i)$, *then*

$$h_{\mathbf{z},\gamma}(\mathbf{v}) = -\frac{1}{2} S^T S \alpha^{\text{opt}} (I^T_{(u+l)\times l} \otimes \mathbf{c}^T) M^T_{\text{reg}} G[\mathbf{v}, \mathbf{x}]^T (I_t \otimes \mathbf{c}). \tag{83}$$

### 5.2.2 Sequential Minimal Optimization (SMO)

We provide an SMO algorithm, which is described in detail in Appendix A.4, to solve the quadratic optimization problem (23) in Theorem 6, as part of Algorithm 2.

## 6. Optimizing the combination operator

In the learning formulation thus far, we have assumed that the combination operator $C$ is given and fixed. Our task then is to find the optimal function $f_{\mathbf{z},\gamma} \in \mathcal{H}_K$ that minimizes the general learning objective (9) in Section 3, given the training data $\mathbf{z}$ and $C$. In this section, we go one step further and show that both $f_{\mathbf{z},\gamma}$ and $C$ can be simultaneously optimized given the training data $\mathbf{z}$ alone.

For the time being, we consider the $m$-view least square learning setting, where $C$ is represented by a vector $\mathbf{c} \in \mathbb{R}^m$. Let $\mathcal{S}_\alpha^{m-1}$ denote the sphere centered at the origin in $\mathbb{R}^m$ with radius $\alpha > 0$, that is $\mathcal{S}_\alpha^{m-1} = \{x \in \mathbb{R}^m : ||x|| = \alpha\}$. Consider the problem of optimizing over both $f \in \mathcal{H}_K$ and $\mathbf{c} \in \mathcal{S}_\alpha^{m-1}$,

$$\begin{aligned}
f_{\mathbf{z},\gamma} = \operatorname{argmin}_{f \in \mathcal{H}_K, \mathbf{c} \in \mathcal{S}_\alpha^{m-1}} \frac{1}{l} \sum_{i=1}^l ||y_i - Cf(x_i)||^2_{\mathcal{Y}} \\
+ \gamma_A ||f||^2_{\mathcal{H}_K} + \gamma_I \langle \mathbf{f}, M\mathbf{f} \rangle_{\mathcal{W}_{u+l}}.
\end{aligned} \tag{84}$$

We first point out a *crucial difference* between our framework and a typical multi-kernel learning approach. Since our formulation does not place any constraint on $\mathbf{c}$, we do not require that $c_i \geq 0$, $i = 1, \ldots, m$. Thus $\mathbf{c}$ is allowed to range over the whole sphere $S_\alpha^{m-1}$, which considerably simplifies the optimization procedure.

The optimization problem (84) is not convex and one common approach to tackle it is via Alternating Minimization. First we fix $\mathbf{c} \in S_\alpha^{m-1}$ and solve for the optimal $f_{\mathbf{z},\gamma} \in \mathcal{H}_K$, which is what we have done so far. Then we fix $f$ and solve for $\mathbf{c}$. Consider $f$ of the form

$$f = \sum_{j=1}^{u+l} K_{x_j} a_j. \tag{85}$$

Then

$$f(x_i) = \sum_{j=1}^{u+l} K(x_i, x_j) a_j = K[x_i]\mathbf{a},$$

where $K[x_i] = (K(x_i, x_1), \ldots, K(x_i, x_{u+l}))$. Since $K[x_i] = G[x_i] \otimes R$, we have

$$f(x_i) = (G[x_i] \otimes R)\mathbf{a}, \quad G[x_i] \in \mathbb{R}^{m \times m(u+l)}.$$

Since $A$ is a matrix of size $m(u+l) \times \dim(\mathcal{Y})$, with $\mathbf{a} = \text{vec}(A^T)$, we have

$$Cf(x_i) = (\mathbf{c}^T \otimes I_{\mathcal{Y}})(G[x_i] \otimes R)\mathbf{a} = (\mathbf{c}^T G[x_i] \otimes R)\mathbf{a} = \text{vec}(RA^T G[x_i]^T \mathbf{c}) = RA^T G[x_i]^T \mathbf{c} \in \mathcal{Y}.$$

Let $F[\mathbf{x}]$ be an $l \times 1$ block matrix, with block $F[\mathbf{x}]_i = RA^T G[x_i]^T$, which is of size $\dim(\mathcal{Y}) \times m$, so that $F[\mathbf{x}]$ is of size $\dim(\mathcal{Y})l \times m$ and $F[\mathbf{x}]\mathbf{c} \in \mathcal{Y}^l$. Then

$$\frac{1}{l}\sum_{i=1}^{l} ||y_i - Cf(x_i)||_{\mathcal{Y}}^2 = \frac{1}{l}||\mathbf{y} - F[\mathbf{x}]\mathbf{c}||_{\mathcal{Y}^l}^2.$$

Thus for $f$ fixed, so that $F[\mathbf{x}]$ is fixed, the minimization problem (84) over $\mathbf{c}$ is equivalent to the following optimization problem

$$\min_{\mathbf{c} \in S_\alpha^{m-1}} \frac{1}{l}||\mathbf{y} - F[\mathbf{x}]\mathbf{c}||_{\mathcal{Y}^l}^2. \tag{86}$$

While the sphere $S_\alpha^{m-1}$ is not convex, it is a compact set and consequently, any continuous function on $S_\alpha^{m-1}$ attains a global minimum and a global maximum. We show in the next section how to obtain an almost closed form solution for the global minimum of (86) in the case $\dim(\mathcal{Y}) < \infty$.

## 6.1 Quadratic optimization on the sphere

Let $A$ be an $n \times m$ matrix, $\mathbf{b}$ be an $n \times 1$ vector, and $\alpha > 0$. Consider the optimization problem

$$\min_{\mathbf{x} \in \mathbb{R}^m} ||A\mathbf{x} - \mathbf{b}||_{\mathbb{R}^n} \text{ subject to } ||\mathbf{x}||_{\mathbb{R}^m} = \alpha. \tag{87}$$

The function $\psi(\mathbf{x}) = ||A\mathbf{x} - \mathbf{b}||_{\mathbb{R}^n} : \mathbb{R}^m \to \mathbb{R}$ is continuous. Thus over the sphere $||\mathbf{x}||_{\mathbb{R}^m} = \alpha$, which is a compact subset of $\mathbb{R}^m$, $\psi(\mathbf{x})$ has a global minimum and a global maximum.

The optimization problem (87) has been analyzed before in the literature under various assumptions, see e.g. (Forsythe and Golub, 1965; Gander, 1981; Golub and von Matt, 1991). In this work, we employ the singular value decomposition approach described in (Gander, 1981), but we *do not impose any constraint* on the matrix $A$ (in (Gander, 1981), it is assumed that $\text{rank}\begin{pmatrix} A \\ I \end{pmatrix} = m$ and $n \geq m$). We next describe the form of the global minimum of $\psi(\mathbf{x})$.

Consider the singular decomposition for $A$,

$$A = U\Sigma V^T, \tag{88}$$

where $U \in \mathbb{R}^{n \times n}$, $\Sigma \in \mathbb{R}^{n \times m}$, $V \in \mathbb{R}^{m \times m}$, with $UU^T = U^T U = I_n$, $VV^T = V^T V = I_m$. Let $r = \text{rank}(A)$, $1 \leq r \leq \min\{m, n\}$, then the main diagonal of $\Sigma$ has the form $(\sigma_1, \ldots, \sigma_r, 0, \ldots, 0)$, with $\sigma_1 \geq \cdots \sigma_r > 0$. Then

$$A^T A = V\Sigma^T \Sigma V^T = VDV^T, \tag{89}$$

where $D = \Sigma^T \Sigma = \text{diag}(\sigma_1^2, \ldots, \sigma_r^2, 0, \ldots, 0) = \text{diag}(\mu_1, \ldots, \mu_m) \in \mathbb{R}^{m \times m}$, with $\mu_i$, $1 \leq i \leq m$, being the eigenvalues of $A^T A \in \mathbb{R}^{m \times m}$.

**Theorem 15** *Assume that $A^T\mathbf{b} = 0$. A global solution of the minimization problem (87) is an eigenvector $\mathbf{x}^*$ of $A^T A$ corresponding to the smallest eigenvalue $\mu_m$, appropriately normalized so that $||\mathbf{x}^*||_{\mathbb{R}^m} = \alpha$. This solution is unique if and only if $\mu_m$ is single. Otherwise, there are infinitely many solutions, each one being a normalized eigenvector in the eigenspace of $\mu_m$.*

**Theorem 16** *Assume that $A^T\mathbf{b} \neq 0$. Let $\mathbf{c} = U^T\mathbf{b}$. Let $\gamma^*$ be the unique real number in the interval $(-\sigma_r^2, \infty)$ such that*

$$s(\gamma^*) = \sum_{i=1}^{r} \frac{\sigma_i^2 c_i^2}{(\sigma_i^2 + \gamma^*)^2} = \alpha^2. \tag{90}$$

*(I) The vector*

$$\mathbf{x}(\gamma^*) = (A^T A + \gamma^* I_m)^{-1} A^T \mathbf{b}, \tag{91}$$

*is the unique global solution of the minimization problem (87) in one of the following cases:*

1. *$\operatorname{rank}(A) = m$.*

2. *$\operatorname{rank}(A) = r < m$ and $\gamma^* > 0$.*

3. *$\operatorname{rank}(A) = r < m$, $\gamma^* < 0$, and $\sum_{i=1}^{r} \frac{c_i^2}{\sigma_i^2} > \alpha^2$.*

*(II) In the remaining case, namely $\operatorname{rank}(A) = r < m$, $\gamma^* \leq 0$, and $\sum_{i=1}^{r} \frac{c_i^2}{\sigma_i^2} \leq \alpha^2$, then the global solution of the minimization problem (87) is given by*

$$\mathbf{x}(0) = V\mathbf{y}, \tag{92}$$

*where $y_i = \frac{c_i}{\sigma_i}$, $1 \leq i \leq r$, with $y_i$, $r+1 \leq i \leq m$, taking arbitrary values such that*

$$\sum_{i=r+1}^{m} y_i^2 = \alpha^2 - \sum_{i=1}^{r} \frac{c_i^2}{\sigma_i^2}. \tag{93}$$

*This solution is unique if and only if $\sum_{i=1}^{r} \frac{c_i^2}{\sigma_i^2} = \alpha^2$. If $\sum_{i=1}^{r} \frac{c_i^2}{\sigma_i^2} < \alpha^2$, then there are infinitely many solutions.*

**Remark 17** *To solve Equation (90), the so-called secular equation, we note that the function $s(\gamma) = \sum_{i=1}^{r} \frac{\sigma_i^2 c_i^2}{(\sigma_i^2 + \gamma)^2}$ is monotonically decreasing on $(-\sigma_r^2, \infty)$ and thus (90) can be solved via a bisection procedure.*

**Remark 18** *We have presented here the solution to the problem of optimizing $C$ in the least square case. The optimization of $C$ in the SVM case is substantially different and will be treated in a future work.*

## 7. Experiments

In this section, we present an extensive empirical analysis of the proposed methods on the challenging tasks of multiclass image classification and species recognition with attributes. We show that the proposed framework[2] is able to combine different types of views and modalities and that it is competitive with other state-of-the-art approaches that have been developed in the literature to solve these problems.

The following methods, which are instances of the presented theoretical framework, were implemented and tested: multi-view learning with least square loss function (MVL-LS), MVL-LS with the optimization of the combination operator (MVL-LS-optC), multi-view learning with binary SVM loss function in the one-vs-all setup (MVL-binSVM), and multi-view learning with multi-class SVM loss function (MVL-SVM).

Our experiments demonstrate that: 1) multi-view learning achieves significantly better performance compared to single-view learning (Sec. 7.4); 2) unlabeled data can be particularly helpful in improving performance when the number of labeled data is small (Sec. 7.4 and Sec. 7.5); 3) the choice and therefore the optimization of the combination operator $C$ is important (Sec. 7.6); and 4) the proposed framework outperforms other state-of-the-art approaches even in the case when we use fewer views (Sec. 7.7).

In the following sections, we first describe the designs for the experiments: the construction of the kernels is described in Sec. 7.1, the used datasets and evaluation protocols in Sec. 7.2 and the selection/validation of the regularization parameters in Sec. 7.3. Afterwards, Sections 7.4, 7.5, 7.6, and 7.7 report the analysis of the obtained results with comparisons to the literature.

### 7.1 Kernels

Assume that each input $x$ has the form $x = (x^1, \ldots, x^m)$, where $x^i$ represents the $i$th view. We set $G(x, t)$ to be the diagonal matrix of size $m \times m$, with

$$(G(x,t))_{i,i} = k^i(x^i, t^i), \quad \text{that is} \quad G(x,t) = \sum_{i=1}^{m} k^i(x^i, t^i) \mathbf{e}_i \mathbf{e}_i^T, \tag{94}$$

where $k^i$ is a scalar-valued kernel defined on view $i$ and $\mathbf{e}_i = (0, \ldots, 1, \ldots, 0) \in \mathbb{R}^m$ is the $i$th coordinate vector. The corresponding Gram matrices are related by

$$G[\mathbf{x}] = \sum_{i=1}^{m} k^i[\mathbf{x}] \otimes \mathbf{e}_i \mathbf{e}_i^T. \tag{95}$$

Note that for each pair $(x, t)$, $G(x, t)$ is a diagonal matrix, but it is *not* separable, that is it cannot be expressed in the form $k(x, t)D$ for a scalar kernel $k$ and a positive semi-definite matrix $D$, because the kernels $k^i$'s are in general different.

To carry out multi-class classification with $P$ classes, $P \geq 2$, using vector-valued least squares regression (Algorithm 1), we set $\mathcal{Y} = \mathbb{R}^P$, and $K(x, t) = G(x, t) \otimes R$, with $R = I_P$. For each $y_i$, $1 \leq i \leq l$, in the labeled training sample, we set $y_i = (-1, \ldots, 1, \ldots, -1)$, with 1 at the $k$th location if $x_i$ is in the $k$th class. When using vector-valued multi-view SVM (Algorithm 2), we set $S$ to be the simplex coding, $\mathcal{Y} = \mathbb{R}^{P-1}$, and $K(x,t) = G(x,t) \otimes R$, with $R = I_{P-1}$.

We remark that since the views are coupled by both the loss functions and the multiview manifold regularization term $M$, even in the simplest scenario, that is fully supervised multi-view binary classification, Algorithm 1 with a diagonal $G(x,t)$ is not equivalent to solving $m$ independent scalar-valued least square regression problems, and Algorithm 2 is not equivalent to solving $m$ independent binary SVMs.

We used $R = I_{\mathcal{Y}}$ for the current experiments. For multi-label learning applications, one can set $R$ to be the output graph Laplacian as done in (Minh and Sindhwani, 2011).

We empirically analyzed the optimization framework of the combination operator $\mathbf{c}$ in the least square setting, as theoretically presented in Section 6. For the experiments with the SVM loss, we set the weight vector $\mathbf{c}$ to be the uniform combination $\mathbf{c} = \frac{1}{m}(1, \ldots, 1)^T \in \mathbb{R}^m$, leaving its optimization, which is substantially different from the least square case, to a future work.

In all experiments, the kernel matrices are used as the weight matrices for the graph Laplacians, unless stated otherwise. This is not necessarily the best choice in practice but we did not use additional information to compute more informative Laplacians at this stage to have a fair comparison with other state of the art techniques.

## 7.2 Datasets and Evaluation Protocols

Three datasets were used in our experiments to test the proposed methods, namely, the Oxford flower species (Nilsback and Zisserman, 2006), Caltech-101 (Fei-Fei et al., 2006), and Caltech-UCSD Birds-200-2011 (Wah et al., 2011). For these datasets, the views are the different features extracted from the input examples as detailed below.

The Flower species dataset (Nilsback and Zisserman, 2006) consists of 1360 images of 17 flower species segmented out from the background. We used the following 7 extracted features in order to fairly compare with (Gehler and Nowozin, 2009): HOG, HSV histogram, boundary SIFT, foreground SIFT, and three features derived from color, shape and texture vocabularies. The features, the respective $\chi^2$ kernel matrices and the training/testing splits[3] are taken from (Nilsback and Zisserman, 2006) and (Nilsback and Zisserman, 2008). The total training set provided by (Nilsback and Zisserman, 2006) consists of 680 labeled images (40 images per class). In our experiments, we varied the number of labeled data $l_c = \{1, 5, 10, 20, 40\}$ images per category and used 85 unlabeled images ($u_c = 5$ per class) taken from the validation set in (Nilsback and Zisserman, 2006) when explicitly stated. The testing set consists of 20 images per class as in (Nilsback and Zisserman, 2006).

The Caltech-101 dataset (Fei-Fei et al., 2006) is a well-known dataset for object recognition that contains 102 classes of objects and about 40 to 800 images per category. We used the features and $\chi^2$ kernel matrices[4] provided in (Vedaldi et al., 2009), consisting of 4 descriptors extracted using a spatial pyramid of three levels, namely PHOW gray and color, geometric blur, and self-similarity. In our experiments, we selected only the lower level of the pyramid, resulting in 4 kernel matrices as in (Minh et al., 2013). We report results using all 102 classes (background class included) averaged over three splits as provided in (Vedaldi et al., 2009). In our tests, we varied the number of labeled data ($l_c = \{5, 10, 15\}$ images per category) in the supervised setup. The test set contained 15 images per class for all of the experiments.

---

3. Available at `http://www.robots.ox.ac.uk/~vgg/data/flowers/17/index.html`.
4. Available at `http://www.robots.ox.ac.uk/~vgg/software/MKL/`.

The Caltech-UCSD Birds-200-2011 dataset (Wah et al., 2011) is used for bird categorization and contains both images and manually-annotated attributes (two modalities)[5]. This dataset is particularly challenging because it contains 200 very similar bird species (classes) for a total of $11,788$ annotated images split between training and test sets. We used the same evaluation protocol and kernel matrices of (Minh et al., 2013). Different training sets were created by randomly selecting 5 times a set of $l_c = \{1, 5, 10, 15\}$ images for each class. All testing samples were used to evaluate the method. We used 5 unlabeled images per class in the semi-supervised setup. The descriptors consist of two views: PHOW gray (Vedaldi et al., 2009) from images and the 312-dimensional binary vector representing attributes provided in (Wah et al., 2011). The $\chi^2$ and Gaussian kernels were used for the appearance and attribute features, respectively.

### 7.3 Regularization Parameters

Let us specify the parameters we used in the experiments. Each method has three regularization parameters, namely, $\gamma_A$ for the standard RKHS regularization, and $\gamma_B$ and $\gamma_W$ for the multi-view manifold regularization. The only dataset for which it was possible to perform independent cross-validation is the Flower species dataset which has a separate validation set from the training set. For the other datasets, cross-validation was omitted in order to have the same number of training examples and therefore to have a fair comparison with the other state-of-the-art methods.

Cross-validation on the flower species dataset was performed using the following set of parameters: $\gamma_A = \{10^{-5}, 10^{-6}, 10^{-7}\}$, $\gamma_B = \{10^{-6}, 10^{-8}, 10^{-9}\}$ and $\gamma_W = \{10^{-6}, 10^{-8}, 10^{-9}\}$. Cross-validation was run on the experiment with $l_c = 10$ labeled data per category. The parameters found during validation were left the same for all the other experiments $l_c = \{1, 5, 20, 40\}$ to have a fair comparison.

The parameters that performed the best on the validation set for the Flower dataset are reported in Table 1a. We also report the parameters chosen for Caltech-101 and the Caltech-UCSD Birds-200-2011 dataset in Table 1b and 1c respectively. Notice that the parameters vary across the different implementations of the proposed framework and especially across the different datasets, as might be expected.

### 7.4 Single-view Vs. Multi-view

The purpose of the experimental analysis in this section is to demonstrate that multi-view learning significantly outperforms single-view learning.

First, we analyzed the contributions of each of the between-view (Eq. 58) and within-view (Eq. 64) regularization terms in (Eq. 9). To this end, we tested multi-view learning with the least squares loss function on Caltech-101. A subset of 10 images for each class were randomly selected, with half used as labeled data $l_c = 5$ and the other half as unlabeled data $u_c = 5$ (see Table 2, last column). We also tested the proposed method in the one-shot learning setup, where the number of labeled images is one per class $l_c = 1$ (see Table 2, third column). The testing set consisted of 15 images per category. For this test, we selected the features at the bottom of each pyramid, because they give the best performance in practice. We can see from Table 2 that both the between-view and within-view regularization terms contribute to increase the recognition rate, *e.g.* with $l_c = 1$ the improvement is 2.35%. As one would expect, the improvement resulting from the use of unlabeled data is bigger when there are more unlabeled data than labeled data, which can be seen by comparing the third and forth columns.

---

5. The dataset is available at `http://www.vision.caltech.edu/visipedia/CUB-200-2011.html`.

Table 1: Parameters for Flower species, Caltech-101 and Caltech-UCSD Birds-200-2011 datasets.

| Method | $\gamma_A$ | $\gamma_B$ | $\gamma_W$ | $\gamma_A$ | $\gamma_B$ | $\gamma_W$ | $\gamma_A$ | $\gamma_B$ | $\gamma_W$ |
|---|---|---|---|---|---|---|---|---|---|
| MVL-LS | $10^{-7}$ | $10^{-9}$ | $10^{-8}$ | $10^{-5}$ | $10^{-6}$ | $10^{-6}$ | $10^{-5}$ | $10^{-6}$ | $10^{-6}$ |
| MVL-binSVM | $10^{-7}$ | $10^{-8}$ | $10^{-9}$ | $10^{-5}$ | $10^{-6}$ | $10^{-6}$ | $10^{-5}$ | $10^{-6}$ | $0$ |
| MVL-SVM | $10^{-6}$ | $10^{-8}$ | $10^{-8}$ | $10^{-6}$ | $10^{-8}$ | $10^{-8}$ | $10^{-5}$ | $10^{-6}$ | $0$ |

| (a) Flower | (b) Caltech-101 | (c) Caltech Birds |
|---|---|---|

Table 2: Results of MVL-LS on Caltech-101 using PHOW color and gray L2, SSIM L2 and GB. The training set is made of 1 or 5 labeled data $l_c$ and 5 unlabeled data per class $u_c$, and 15 images per class are left for testing.

| $\gamma_B$ | $\gamma_W$ | Accuracy $l_c = 1, u_c = 5$ | Accuracy $l_c = u_c = 5$ |
|---|---|---|---|
| $0$ | $0$ | 30.59% | 63.68% |
| $0$ | $10^{-6}$ | 31.81% | 63.97% |
| $10^{-6}$ | $0$ | 32.44% | 64.18% |
| $10^{-6}$ | $10^{-6}$ | **32.94%** | **64.2%** |

To demonstrate that multi-view learning is able to combine features properly, we report in Table 3 the performance in terms of average accuracy of each feature independently and of the proposed methods with all 10 views combined (last three rows). The improvement with respect to the view that gives the best results (PHOW gray L2) is 4.77% for the case with $l_c = 1$ (second column) and 5.62% for the case with $l_c = 5$ (last column). It is also worth noticing that all the proposed methods outperform the best single view (PHOW gray L2). Moreover, it is important to point out that the best views for each feature correspond to the L2 level. We show in Sec. 7.6 that the optimization of the combination operator leads to very similar findings.

To further demonstrate the performance of multi-view learning, we run a similar experiment on the Caltech-UCSD Birds-200-2011 dataset, with the results shown in Table 4. We compare the results obtained by the single views (PHOW and attributes) with the proposed multi-view learning methods (last three rows) when increasing the number of labeled data per class $l_c = \{1, 5, 10, 15\}$. In all the cases shown in the table, we obtain better results using the proposed multi-view learning framework compared with single-view learning.

Table 3: Results on Caltech-101 using each feature in the single-view learning framework and all 10 features in the multi-view learning framework (last three rows).

| Feature | Accuracy $l_c = 1, u_c = 5$ | Accuracy $l_c = u_c = 5$ |
|---------|------------|------------|
| PHOW color L0 | 13.66% | 33.14% |
| PHOW color L1 | 17.1% | 42.03% |
| PHOW color L2 | 18.71% | 45.86% |
| PHOW gray L0 | 20.31% | 45.38% |
| PHOW gray L1 | 24.53% | 54.86% |
| PHOW gray L2 | 25.64% | 56.75% |
| SSIM L0 | 15.27% | 35.27% |
| SSIM L1 | 20.83% | 45.12% |
| SSIM L2 | 22.64% | 48.47% |
| GB | 25.01% | 44.49% |
| MVL-LS | **30.41%** | 61.46% |
| MVL-binSVM | 30.20% | **62.37%** |
| MVL-SVM | 27.23% | 60.04% |

Table 4: Results on the Caltech-UCSD Birds-200-2011 dataset in the semi-supervised setup.

| | $l_c = 1$ | $l_c = 5$ | $l_c = 10$ | $l_c = 15$ |
|---|---|---|---|---|
| PHOW | 2.75% | 5.51% | 8.08% | 9.92% |
| Attributes | 13.53% | 30.99% | 38.96% | 43.79% |
| MVL-LS | 14.31% | 33.25% | 41.98% | 46.74% |
| MVL-binSVM | **14.57%** | **33.50%** | **42.24%** | **46.88%** |
| MVL-SVM | 14.15% | 31.54% | 39.30% | 43.86% |

## 7.5 Increasing the Label Set Size

In this section, we analyze the behavior of the proposed methods when increasing the size of the set of labeled data, in both supervised and semi-supervised settings.

In Table 5, we reported the results in terms of accuracy and its standard deviation (between brackets) on the Caltech-101 dataset comparing with other state of the art methods. The first three rows report the results of the methods tested by (Gehler and Nowozin, 2009). The forth, fifth and sixth rows show the statistics of the proposed methods in the supervised setup. We also reported the results of the best methods among the proposed ones in the semisupervised setup (with 5 unlabeled data for each class).

First, the results demonstrate that the proposed methods improve significantly when increasing the size of the labeled set. This fact can be observed also for the Caltech-UCSD Birds-200-2011 experiment in Table 4. More interestingly, when the number of labeled data is 5 per class (third column), our methods strongly improve the best result of (Gehler and Nowozin, 2009) by at least 9.4 percentage points. Similar observations can be made by examining the results obtained by Bucak et al. (2014) for $l_c = 10$ (Table 4 in their paper): our best result in Table 5 (71.4%) outperforms their best result (60.3%) by 11.1 percentage points. Moreover, one can see that the improvement when using unlabeled data (last row) is bigger when there are many more of them compared with labeled data, as expected (see the columns with 1 and 5 labeled images per class). When the number of labeled data increases, the proposed methods in the supervised setup can give comparable or better results (see the column with 10 labeled images per class). A similar behavior is shown in Table 6, when dealing the problem of species recognition with the Flower dataset. The best improvement we obtained in the semi-supervised setup is with 1 labeled data per category. This finding suggests that the unlabeled data provide additional information about the distribution in the input space when there are few labeled examples. On the other hand, when there are sufficient labeled data to represent well the distribution in the input space, the unlabeled data will not provide an improvement of the results.

Table 5: Results on Caltech-101 when increasing the number of labeled data and comparisons with other state of the art methods reported by (Gehler and Nowozin, 2009). Best score in bold, second best score in italic.

| | $l_c = 1$ | $l_c = 5$ | $l_c = 10$ | $l_c = 15$ |
|---|---|---|---|---|
| MKL | N/A | 42.1% (1.2%) | 55.1% (0.7%) | 62.3% (0.8%) |
| LP-B | N/A | 46.5% (0.9%) | 59.7% (0.7%) | 66.7% (0.6%) |
| LP-$\beta$ | N/A | 54.2% (0.6%) | 65.0% (0.9%) | 70.4% (0.7%) |
| MVL-LS | *31.2%* (1.1%) | 64.0% (1.0%) | 71.0% (0.3%) | 73.3% (1.3%) |
| MVL-binSVM | 31.0% (1.3%) | *64.1%* (0.7%) | **71.4%** (0.3%) | **74.1%** (0.9%) |
| MVL-SVM | 30.6% (1.0%) | 63.6% (0.4%) | 70.6% (0.2%) | *73.5%* (1.0%) |
| MVL-binSVM (semi-sup. $u_c = 5$) | **32.4%** (1.2%) | **64.4%** (0.4%) | *71.4%* (0.2%) | N/A |

Table 6: Results on the Flower dataset (17 classes) when increasing the number of training images per class. Best score in bold, second best score in italic.

| | $l_c = 1$ | $l_c = 5$ | $l_c = 10$ | $l_c = 20$ | $l_c = 40$ |
|---|---|---|---|---|---|
| MVL-LS | 39.41% (1.06%) | *65.78%* (3.68%) | 74.41% (1.28%) | 81.76% (3.28%) | **86.37%** (1.80%) |
| MVL-binSVM | 39.71% (1.06%) | 64.80% (4.42%) | *74.41%* (0.29%) | 81.08% (3.09%) | 86.08% (2.21%) |
| MVL-SVM | 39.31% (1.62%) | 65.29% (4.04%) | 74.41% (1.28%) | 81.67% (2.78%) | 86.08% (1.80%) |
| MVL-LS (semi-sup.) | **41.86%** (2.50%) | **66.08%** (3.45%) | **75.00%** (1.06%) | **82.35%** (2.70%) | 85.78% (2.78%) |
| MVL-binSVM (semi-sup.) | *40.59%* (2.35%) | 65.49% (4.58%) | 74.22% (0.68%) | 81.57% (2.67%) | 85.49% (0.74%) |
| MVL-SVM (semi-sup.) | 34.80% (1.11%) | 65.49% (4.17%) | 74.41% (0.49%) | *81.78%* (2.61%) | *86.08%* (1.51%) |

### 7.6 Optimizing the Combination Operator

In the previous experiments, the combination weight vector $\mathbf{c}$ was uniform, meaning that each view (*i.e.* kernel) has the same importance during classification. However, in practice it often happens that some views are more useful and informative than others. We observed this in our experiments, where different choices of the weights $\mathbf{c}$ gave rise to different classification accuracies. In particular, we empirically found for the Flower dataset using MVL-LS that $\mathbf{c} = (0.1431, 0.1078, 0.1452, 0.1976, 0.0991, 0.1816, 0.1255)^T$ yields an accuracy of 87.75%, the state-of-the-art result for that dataset. This suggests that there exists at least one better choice for $\mathbf{c}$.

In this section, we carry out an empirical analysis of the strategy presented in Section 6 which performs optimization to obtain the optimal weight vector $\mathbf{c}$. We call this method MVL-LS-optC. The analysis was performed on the Caltech-101 dataset and the Flower dataset. For the experiment using the Caltech-101 dataset, we created a validation set by selecting 5 examples for each class from the training set. For the experiment using the Flower dataset, the validation set was already provided (see Sec. 7.2 for detail). The validation set is used to determine the best value of $\mathbf{c}$ found over all the iterations using different initializations. We carried out the iterative optimization procedure 20 times, each time with a different random unit vector as the initialization vector for $\mathbf{c}$, and reported the run with the best performance over the validation set.

The results of MVL-LS-optC for the Caltech-101 dataset and the Flower dataset are reported in Tables 7, 8 and 9. We empirically set $\alpha = 2$ and $\alpha = 1$ in Equation 87 for the Caltech-101 dataset and the Flower dataset, respectively. MVL-LS-optC is compared with MVL-LS which uses uniform weights. We analyze in the next section how MVL-LS-optC compares with MVL-binSVM, MVL-SVM, and the state of the art.

We first discuss the results on the Caltech-101 dataset using all 10 kernels. Table 7 shows that there is a significant improvement from 0.4% to 2.5% with respect to the results with uniform weights for the Caltech-101 dataset. The best $\mathbf{c}$ found during training in the case of $l_c = 10$ was $\mathbf{c}^* = (0.1898, 0.6475, -0.7975, 0.3044, 0.1125, -0.4617, -0.1531, 0.1210, 1.2634, 0.9778)^T$. Note that the $c_i$'s can assume negative values (as is the case here) and as we show in Section 8.1, the contribution of the $i$th view is determined by the square weight $c_i^2$. This experiment confirms our findings in Sec. 7.4: the best 4 views are PHOW color L2, PHOW gray L2, SSIM L2 and GB, which are the $c_3$, $c_6$, $c_9$ and $c_{10}$ components of $\mathbf{c}$, respectively.

We now focus on the top 4 views and apply again the optimization method to see if there is still a margin of improvement. We expect to obtain better results with respect to 10 views because the 4-dimentional optimization should in practice be easier than the 10-dimensional one, given that the size of the search space is smaller. Table 8 shows the results with the top 4 kernels. We observe that there is an improvement with respect to MVL-LS that varies from 0.3% to 1.1%. We can also notice that there is not a significant improvement of the results when using more iteration (25 vs. 50 iterations). We again inspected the learned combination weights and discovered that in average they are very close to the uniform distribution, *i.e.* $\mathbf{c}^* = (-0.4965, -0.5019, -0.4935, -0.5073)^T$. This is mainly because we pre-selected the best set of 4 kernels accordingly to the previous 10-kernel experiment.

Table 7: Results using the procedure to optimize the combination operator on Caltech-101 considering all 10 kernels. Best score in bold, second best score in italic.

|  | $l_c = 1$ | $l_c = 5$ | $l_c = 10$ |
|---|---|---|---|
| MVL-LS (uniform) | 28.4% (1.8%) | 61.4% (1.1%) | 68.1% (0.3%) |
| MVL-LS-optC (25 it.) | **28.8%** (1.7%) | **63.1%** (0.1%) | **70.6%** (0.5%) |

Table 8: Results using the procedure to optimize the combination operator on Caltech-101 using the top 4 kernels. Best score in bold, second best score in italic.

|  | $l_c = 1$ | $l_c = 5$ | $l_c = 10$ |
|---|---|---|---|
| MVL-LS (uniform) | 31.2% (1.1%) | 64.0% (1.0%) | 71.0% (0.3%) |
| MVL-LS-optC (25 it.) | **32.1%** (1.5%) | *64.5%* (0.9%) | **71.3%** (0.4%) |
| MVL-LS-optC (50 it.) | *32.1%* (2.3%) | **64.7%** (1.1%) | *71.3%* (0.5%) |

We finally used the best **c** learned in the case of $l_c = 10$ to do an experiment[6] with $l_c = 15$ on the Caltech-101. MVL-LS-optC obtains an accuracy of 73.85%, outperforming MVL-LS (uniform), which has an accuracy of 73.33% (see Table 10).

For the Flower dataset, Table 9 shows consistent results with the previous experiment. MVL-LS-optC outperforms MVL-LS (uniform weights) in terms of accuracy with an improvement ranging from 0.98% to 4.22%. To have a deeper understanding about which views are more important, we analyzed the combination weights of the best result in Table 9 (last row, last column). The result of the optimization procedure is $\mathbf{c}^* = (-0.3648, -0.2366, 0.3721, 0.5486, -0.4108, 0.3468, 0.2627)^T$ which suggests that the best accuracy is obtained by exploiting the complementarity between shape-based features ($c_3$ and $c_4$) and color-based features ($c_5$) relevant for flower recognition[7].

Table 9: Results using the procedure to optimize the combination operator on the Flower dataset. Best score in bold, second best score in italic.

|  | $l_c = 1$ | $l_c = 5$ | $l_c = 10$ | $l_c = 20$ | $l_c = 40$ |
|---|---|---|---|---|---|
| MVL-LS (uniform) | 39.41% (1.06%) | 65.78% (3.68%) | 74.41% (1.28%) | *81.76%* (3.28%) | 86.37% (1.80%) |
| MVL-LS-optC (25 it.) | *43.14%* (3.38%) | *68.53%* (2.90%) | *75.00%* (0.29%) | 81.47% (2.06%) | *87.25%* (1.51%) |
| MVL-LS-optC (50 it.) | **43.63%** (3.25%) | **68.63%** (2.86%) | **75.39%** (0.90%) | **82.45%** (3.51%) | **87.35%** (1.35%) |

The proposed optimization procedure is powerful, with clear improvements in classification accuracies over the uniform weight approach. However, it comes with a price during the training phase. Firstly, it is an iterative procedure, and therefore it is more computationally expensive with respect to the original MVL-LS formulation. In particular, it is $N_C$

---

Table 10: Comparison with state-of-the-art methods on the Caltech-101 dataset using PHOW color and gray L2, SSIM L2 and GB in the supervised setup (15 labeled images per class). Best score in bold, second best score in italic.

| Method | # of Kernels | Accuracy |
|---|---|---|
| (Yang et al., 2009) | $\geq 10$ | 73.2% |
| (Christoudias et al., 2009) | 4 | 73.00% |
| LP-$\beta$ (Gehler and Nowozin, 2009) | 39 | 70.40% |
| MKL (Vedaldi et al., 2009) | 10 | 71.10% |
| MVL-LS | 4 | 73.33% |
| MVL-LS-optC | 4 | *73.85%* |
| MVL-binSVM | 4 | **74.05%** |
| MVL-SVM | 4 | 73.55% |

times more expensive than MVL-LS, where $N_C$ is the number of iterations. Secondly, since the joint optimization of $(\mathbf{c}, f_{\mathbf{z},\gamma})$ is non-convex, even though we are guaranteed to obtain the global minimum for $\mathbf{c}$ *during each single iteration*, the final $\mathbf{c}$ is *not* guaranteed to be the global minimum of the joint optimization problem itself.

### 7.7 Comparing with the State of the Art

In this section, we show how the proposed methods compare with other state-of-the-art approaches for each recognition problem.

In Table 10, we reported the best results we obtained for the task of object recognition using Caltech-101 in the supervised setup. Observe that all the proposed methods outperform the other techniques, even though they use much less information: 4 kernels versus e.g. 39 kernels in (Gehler and Nowozin, 2009).

In particular, we obtained the best result with the binary version of MVL in the one-vs-all setup. This is not surprising since the one-vs-all approach has often been shown to be very competitive in many computer vision tasks compared to proper multi-class formulations. The second best result is obtained by MVL-LS-optC since it uses an additional optimization step (of $\mathbf{c}$) with respect to the other methods. The optimization of $\mathbf{c}$ for MVL-binSVM and MVL-SVM is substantially different from the least square case and will be treated in a future work.

In Table 11, we reported the best results obtained for the task of species recognition using the Flower dataset in the supervised setup. The proposed methods are compared with MKL, LP-B and LP-$\beta$ by (Gehler and Nowozin, 2009) as well as the more recent results of MK-SVM Shogun, MK-SVM OBSCURE and MK-FDA from (Yan et al., 2012). For this dataset, our best result is obtained by the MVL-LS-optC method outperforming also the recent method MK-FDA from (Yan et al., 2012). We note also, that even with the uniform weight vector (MVL-LS), our methods outperform MK-FDA on Caltech-101, which uses 10 kernels, see Figures 6 and 9 in (Yan et al., 2012).

Table 11: Results on the Flower dataset comparing the proposed method with other state-of-the-art techniques in the supervised setup. The first four rows are from (Gehler and Nowozin, 2009) while rows 5-7 are methods presented by (Yan et al., 2012).

| Method | Accuracy | |
|---|---|---|
| MKL (SILP) | 85.2% | (1.5%) |
| MKL (Simple) | 85.2% | (1.5%) |
| LP-B | 85.4% | (2.4%) |
| LP-$\beta$ | 85.5% | (3.0%) |
| MK-SVM Shogun | 86.0% | (2.4%) |
| MK-SVM OBSCURE | 85.6% | (0.0%) |
| MK-FDA | *87.2%* | (1.6%) |
| MVL-LS | 86.4% | (1.8%) |
| MVL-LS-optC | **87.35%** | (1.3%) |
| MVL-binSVM | 86.1% | (2.2%) |
| MVL-SVM | 86.1% | (1.8%) |

## 8. Further theoretical analysis

### 8.1 Connection with Multiple Kernel Learning

In this section, we briefly explore the connection between our multi-view learning framework and multiple kernel learning, see e.g. (Bach et al., 2004). We show that in the purely supervised setting, when $\gamma_I = 0$, $u = 0$, that is without unlabeled data and without between-view regularization, for $C = \mathbf{c}^T \otimes I_{\mathcal{Y}}$, $K(x,t) = G(x,t) \otimes I_{\mathcal{Y}}$, $G(x,t) = \sum_{i=1}^m k^i(x,t)\mathbf{e}_i\mathbf{e}_i^T$, we obtain supervised learning (vector-valued least square regression and SVM) with the combined kernel $\sum_{i=1}^m c_i^2 k^i(x,t)I_{\mathcal{Y}}$, where $k^i$ is a scalar-valued kernel corresponding to view $i$. In particular, for $\mathcal{Y} = \mathbb{R}$, we obtain scalar-values least square regression and binary SVM with the combined kernel $\sum_{i=1}^m c_i^2 k^i(x,t)$. Specifically, we have the following results.

**Corollary 19** *Consider the special case* $\gamma_I = 0$, $u = 0$. *The system of linear equations (13) in Theorem 4 has solution*

$$\mathbf{a} = (I_l \otimes C^*) \left[ (I_l \otimes C)K[\mathbf{x}](I_l \otimes C^*) + l\gamma_A I_{\mathcal{Y}^l} \right]^{-1} \mathbf{y}. \qquad (96)$$

*For* $C = \mathbf{c}^T \otimes I_{\mathcal{Y}}$, $K(x,t) = G(x,t) \otimes I_{\mathcal{Y}}$, *and* $G(x,t) = \sum_{i=1}^m k^i(x,t)\mathbf{e}_i\mathbf{e}_i^T$, *for any* $v \in \mathcal{X}$,

$$C f_{\mathbf{z},\gamma}(v) = \left\{ \sum_{i=1}^m c_i^2 k^i[v,\mathbf{x}] \left( \sum_{i=1}^m c_i^2 k^i[\mathbf{x}] + l\gamma_A I_l \right)^{-1} \otimes I_{\mathcal{Y}} \right\} \mathbf{y}. \qquad (97)$$

*In particular, if* $\mathcal{Y} = \mathbb{R}$, *then*

$$C f_{\mathbf{z},\gamma}(v) = \left\{ \sum_{i=1}^m c_i^2 k^i[v,\mathbf{x}] \left( \sum_{i=1}^m c_i^2 k^i[\mathbf{x}] + l\gamma_A I_l \right)^{-1} \right\} \mathbf{y}. \qquad (98)$$

*This is precisely the solution of scalar-valued regularized least square regression with the combined kernel* $\sum_{i=1}^m c_i^2 k^i(x,t)$.

**Corollary 20** *Consider the special case $\gamma_I = 0$, $u = 0$. Then in Theorem 7,*

$$Q[\mathbf{x}, \mathbf{y}, C] = \frac{1}{\gamma_A} \mathrm{diag}(S_{\hat{\mathbf{y}}}^*)(I_l \otimes C) K[\mathbf{x}](I_l \otimes C^*) \mathrm{diag}(S_{\hat{\mathbf{y}}}).$$

*For $C = \mathbf{c}^T \otimes I_{\mathcal{Y}}$, $K(x,t) = G(x,t) \otimes I_{\mathcal{Y}}$, $G(x,t) = \sum_{i=1}^m k^i(x,t)\mathbf{e}_i\mathbf{e}_i^T$,*

$$Q[\mathbf{x}, \mathbf{y}, C] = \frac{1}{\gamma_A} \mathrm{diag}(S_{\hat{\mathbf{y}}}^*) \left( \sum_{i=1}^m c_i^2 k^i[\mathbf{x}] \otimes I_{\mathcal{Y}} \right) \mathrm{diag}(S_{\hat{\mathbf{y}}}),$$

*and for any $v \in \mathcal{X}$,*

$$C f_{\mathbf{z}, \gamma}(v) = -\frac{1}{2\gamma_A} \left( \sum_{i=1}^m c_i^2 k^i[v, \mathbf{x}] \otimes I_{\mathcal{Y}} \right) \mathrm{diag}(S_{\hat{\mathbf{y}}}) \mathrm{vec}(\beta^{\mathrm{opt}}).$$

*In the binary case, $\mathcal{Y} = \mathbb{R}$, so that*

$$Q[\mathbf{x}, \mathbf{y}, C] = \frac{1}{\gamma_A} \mathrm{diag}(\mathbf{y}) \left( \sum_{i=1}^m c_i^2 k^i[\mathbf{x}] \right) \mathrm{diag}(\mathbf{y}),$$

$$C f_{\mathbf{z}, \gamma}(v) = -\frac{1}{2\gamma_A} \left( \sum_{i=1}^m c_i^2 k^i[v, \mathbf{x}] \right) \mathrm{diag}(\mathbf{y}) \beta^{\mathrm{opt}}.$$

*This is precisely the solution of binary SVM with the combined kernel $\sum_{i=1}^m c_i^2 k^i(x,t)$.*

**Remark 21** *In the sum $\sum_{i=1}^m c_i^2 k^i(x,t)$, the coefficients $c_i$'s are automatically non-negative. This is in accordance with the fact that our formulation makes no mathematical constraint on the coefficients $c_i$'s in the sum $\sum_{i=1}^m c_i f^i(x)$. This is one difference between our approach and the typical multiple kernel learning setting (Bach et al., 2004), where one considers a sum of the form $\sum_{i=1}^m d_i k^i(x,t)$, where the $d_i$'s must be non-negative to guarantee the positive definiteness of the combined kernel.*

## 8.2 Connection with Multi-task Learning

In this section, we briefly explore the connection between our learning formulation and multi-task learning, see e.g. (Evgeniou et al., 2005) . Let $n$ be the number of tasks, $n \in \mathbb{N}$.

Consider the case where the tasks have the same input space. Let $\mathcal{T}$ be a separable Hilbert space. Let $G : \mathcal{X} \times \mathcal{X} \to \mathcal{L}(\mathcal{T})$ be an operator-valued positive definite kernel, which induces an RKHS of functions with values in the Hilbert space $\mathcal{T}$. Consider the kernel $K(x,t)$ of the form

$$K(x,t) = R \otimes G(x,t), \tag{99}$$

where $R$ is a symmetric, positive semidefinite matrix of size $n \times n$. The kernel $K(x,t)$ induces an RKHS of functions with values in the Hilbert space $\mathcal{T}^n$. Each function $f \in \mathcal{H}_K$ has the form $f(x) = (f^1(x), \ldots, f^n(x))$, with $f^k \in \mathcal{H}_G$, where $f^k(x)$ represents the output corresponding to the $k$th task.

In the simplest scenario, $\mathcal{W} = \mathcal{Y} = \mathcal{T}^n$, $C = I$, and the minimization problem (9) thus gives us a vector-valued semi-supervised multi-task learning formulation.

The tasks $f^k$'s are related by the following, which is a generalization of (Evgeniou et al., 2005) (see their formulas (19), (20), (23)) to the nonlinear setting.

**Lemma 22** *Let $K$ be defined by (99), where $R$ is strictly positive definite. For $f = (f^1, \ldots, f^n) \in \mathcal{H}_K$, with $f^k \in \mathcal{H}_G$, we have*

$$||f||^2_{\mathcal{H}_K} = \sum_{k,l=1}^{n} R^{-1}_{kl} \langle f^k, f^l \rangle_{\mathcal{H}_G}. \tag{100}$$

*In particular, for*

$$R = I_n + \frac{1-\lambda}{n\lambda} \mathbf{1}_n \mathbf{1}_n^T, \quad 0 < \lambda \le 1, \tag{101}$$

*we have*

$$||f||^2_{\mathcal{H}_K} = \lambda \sum_{k=1}^{n} ||f^k||^2_{\mathcal{H}_G} + (1-\lambda) \sum_{k=1}^{n} ||f^k - \frac{1}{n} \sum_{l=1}^{n} f^l||^2_{\mathcal{H}_G}. \tag{102}$$

Consider the case when the tasks have different input spaces, such as in the approach to multi-view learning (Kadri et al., 2013), in which each view corresponds to one task and the tasks all share the same output label. Then we have $m$ tasks for $m$ views and we define

$$K(x, t) = G(x, t) \otimes R,$$

as in Section 5, where $G : \mathcal{X} \times \mathcal{X} \to \mathbb{R}^{m \times m}$ is a matrix-valued positive definite kernel, $R \in \mathcal{L}(\mathcal{T})$ is a symmetric, positive operator, so that each task has output in the Hilbert space $\mathcal{T}$. We obtain the formulation of (Kadri et al., 2013) if we set $\mathcal{T} = \mathbb{R}$, so that $R = 1$, duplicate each label $y_i \in \mathbb{R}$ into a vector $(y_i, \ldots, y_i) \in \mathbb{R}^m$, and set $G(x, t)$ to be their covariance-based kernel, with $\gamma_I = 0$, $u = 0$.

We have thus shown how two different scenarios in multi-task learning fall within the scope of our learning formulation. A more in-depth study of our framework in connection with multi-task learning is left to a future work.

## 9. Discussion, Conclusion, and Future Work

We have presented a general learning framework in vector-valued RKHS which encompasses and generalizes many kernel-based learning algorithms in the literature. In particular, we generalize

- the Vector-valued Manifold Regularization framework of (Minh and Sindhwani, 2011), and thus also the vector-valued Regularized Least Square regression formulation of (Micchelli and Pontil, 2005), which are single-view and formulated with the least square loss, to the multi-view setting, formulated with both the least square and multi-class SVM loss functions;

- the Simplex Cone SVM of (Mroueh et al., 2012) , which is supervised, to the multi-view and semi-supervised settings, together with a more general loss function;

- the Laplacian SVM of (Belkin et al., 2006), which is binary and single-view, to the multi-class and multi-view settings.

The generality of the framework and the competitive numerical results we have obtained so far demonstrate that this is a promising venue for further research exploration. Some potential directions for our future work include

- a principled optimization framework for the weight vector $\mathbf{c}$ in the SVM setting, as well as the study of more general forms of the combination operator $C$;

- numerical experiments with different forms of the matrix-valued kernel $K$;

- theoretical and empirical analysis for the SVM under different coding schemes other than the simplex coding;

- theoretical analysis of our formulation, in particular when the numbers of labeled and unlabeled data points go to infinity;

- further connections between our framework and Multi-task learning;

- exploration of our framework in combination with feature learning methods, particularly those coming from deep learning;

- further analysis to optimize the framework for large-scale classification problems.

Apart from the numerical experiments on object recognition reported in this paper, practical applications for our learning framework so far include person re-identification in computer vision (Figueira et al., 2013) and user recognition and verification in Skype chats (Roffo et al., 2013). As we further develop and refine the current formulation, we expect to apply it to other applications in computer vision, image processing, and bioinformatics.

## Appendices.

The Appendices contain three sections. First, in Appendix A, we give the proofs for all the main mathematical results in the paper. Second, in Appendix B, we provide a natural generalization of our framework to the case the point evaluation operator $f(x)$ is replaced by a general bounded linear operator. Last, in Appendix C, we provide an exact description of Algorithm 1 with the Gaussian or similar kernels in the degenerate case, when the kernel width $\sigma \to \infty$.

## Appendix A. Proofs of Main Results

**Notation**: The definition of $\mathbf{f}$ as given by

$$\mathbf{f} = (f(x_1), \ldots, f(x_{u+l})) \in \mathcal{W}^{u+l}, \tag{103}$$

is adopted because it is also applicable when $\mathcal{W}$ is an infinite-dimensional Hilbert space. For $\mathcal{W} = \mathbb{R}^m$,

$$\mathbf{f} = (f^1(x_1), \ldots, f^m(x_1), \ldots, f^1(x_{u+l}), \ldots, f^m(x_{u+l})).$$

This is different from (Rosenberg et al., 2009), where

$$\mathbf{f} = (f^1(x_1), \ldots, f^1(x_{u+l}), \ldots, f^m(x_1), \ldots, f^m(x_{u+l})).$$

This means that our matrix $M$ is necessarily a permutation of the matrix $M$ in (Rosenberg et al., 2009) when they give rise to the same semi-norm.

## A.1 Proof of the Representer Theorem

Since $f(x) = K_x^* f$, the minimization problem (9) is

$$f_{\mathbf{z},\gamma} = \operatorname{argmin}_{f \in \mathcal{H}_K} \frac{1}{l} \sum_{i=1}^{l} V(y_i, CK_{x_i}^* f) + \gamma_A ||f||_{\mathcal{H}_K}^2 + \gamma_I \langle \mathbf{f}, M\mathbf{f} \rangle_{\mathcal{W}_{u+l}}. \tag{104}$$

Consider the operator $E_{C,\mathbf{x}} : \mathcal{H}_K \to \mathcal{Y}^l$, defined by

$$E_{C,\mathbf{x}} f = (CK_{x_1}^* f, \ldots, CK_{x_l}^* f), \tag{105}$$

with $CK_{x_i}^* : \mathcal{H}_K \to \mathcal{Y}$ and $K_{x_i} C^* : \mathcal{Y} \to \mathcal{H}_K$. For $\mathbf{b} = (b_1, \ldots, b_l) \in \mathcal{Y}^l$, we have

$$\langle \mathbf{b}, E_{C,\mathbf{x}} f \rangle_{\mathcal{Y}^l} = \sum_{i=1}^{l} \langle b_i, CK_{x_i}^* f \rangle_{\mathcal{Y}} = \sum_{i=1}^{l} \langle K_{x_i} C^* b_i, f \rangle_{\mathcal{H}_K}. \tag{106}$$

The adjoint operator $E_{C,\mathbf{x}}^* : \mathcal{Y}^l \to \mathcal{H}_K$ is thus

$$E_{C,\mathbf{x}}^* : (b_1, \ldots, b_l) \to \sum_{i=1}^{l} K_{x_i} C^* b_i. \tag{107}$$

The operator $E_{C,\mathbf{x}}^* E_{C,\mathbf{x}} : \mathcal{H}_K \to \mathcal{H}_K$ is then

$$E_{C,\mathbf{x}}^* E_{C,\mathbf{x}} f \to \sum_{i=1}^{l} K_{x_i} C^* CK_{x_i}^* f, \tag{108}$$

with $C^* C : \mathcal{W} \to \mathcal{W}$.

**Proof of Theorem 2**. Denote the right handside of (9) by $I_l(f)$. Then $I_l(f)$ is coercive and strictly convex in $f$, and thus has a unique minimizer. Let $\mathcal{H}_{K,\mathbf{x}} = \{\sum_{i=1}^{u+l} K_{x_i} w_i : \mathbf{w} \in \mathcal{W}^{u+l}\}$. For $f \in \mathcal{H}_{K,\mathbf{x}}^\perp$, the operator $E_{C,\mathbf{x}}$ satisfies

$$\langle \mathbf{b}, E_{C,\mathbf{x}} f \rangle_{\mathcal{Y}^l} = \langle f, \sum_{i=1}^{l} K_{x_i} C^* b_i \rangle_{\mathcal{H}_K} = 0,$$

for all $\mathbf{b} \in \mathcal{Y}^l$, since $C^* b_i \in \mathcal{W}$. Thus

$$E_{C,\mathbf{x}} f = (CK_{x_1}^* f, \ldots, CK_{x_l}^* f) = 0.$$

Similarly, by the reproducing property, the sampling operator $S_{\mathbf{x}}$ satisfies

$$\langle S_{\mathbf{x}} f, \mathbf{w} \rangle_{\mathcal{W}^{u+l}} = \langle f, \sum_{i=1}^{u+l} K_{x_i} w_i \rangle_{\mathcal{H}_K} = 0,$$

for all $\mathbf{w} \in \mathcal{W}^{u+l}$. Thus

$$\mathbf{f} = S_{\mathbf{x}} f = (f(x_1), \ldots, f(x_{u+l})) = 0.$$

For an arbitrary $f \in \mathcal{H}_K$, consider the orthogonal decomposition $f = f_0 + f_1$, with $f_0 \in \mathcal{H}_{K,\mathbf{x}}$, $f_1 \in \mathcal{H}_{K,\mathbf{x}}^\perp$. Then, because $||f_0 + f_1||_{\mathcal{H}_K}^2 = ||f_0||_{\mathcal{H}_K}^2 + ||f_1||_{\mathcal{H}_K}^2$, the result just obtained shows that

$$I_l(f) = I_l(f_0 + f_1) \geq I_l(f_0)$$

with equality if and only if $||f_1||_{\mathcal{H}_K} = 0$, that is $f_1 = 0$. Thus the minimizer of (9) must lie in $\mathcal{H}_{K,\mathbf{x}}$. $\blacksquare$

## A.2 Proofs for the Least Square Case

We have for the least square case:

$$f_{\mathbf{z},\gamma} = \operatorname{argmin}_{f \in \mathcal{H}_K} \frac{1}{l} \sum_{i=1}^l ||y_i - CK_{x_i}^* f||_{\mathcal{Y}}^2 + \gamma_A ||f||_K^2 + \gamma_I \langle \mathbf{f}, M\mathbf{f} \rangle_{\mathcal{W}(u+l)}. \tag{109}$$

With the operator $E_{C,\mathbf{x}}$, (109) is transformed into the minimization problem

$$f_{\mathbf{z},\gamma} = \operatorname{argmin}_{f \in \mathcal{H}_K} \frac{1}{l} ||E_{C,\mathbf{x}} f - \mathbf{y}||_{\mathcal{Y}^l}^2 + \gamma_A ||f||_K^2 + \gamma_I \langle \mathbf{f}, M\mathbf{f} \rangle_{\mathcal{W}^{u+l}}. \tag{110}$$

**Proof of Theorem 3**. By the Representer Theorem, (10) has a unique solution. Differentiating (110) and setting the derivative to zero gives

$$(E_{C,\mathbf{x}}^* E_{C,\mathbf{x}} + l\gamma_A I + l\gamma_I S_{\mathbf{x},u+l}^* M S_{\mathbf{x},u+l}) f_{\mathbf{z},\gamma} = E_{C,\mathbf{x}}^* \mathbf{y}.$$

By definition of the operators $E_{C,\mathbf{x}}$ and $S_{\mathbf{x}}$, this is

$$\sum_{i=1}^l K_{x_i} C^* C K_{x_i}^* f_{\mathbf{z},\gamma} + l\gamma_A f_{\mathbf{z},\gamma} + l\gamma_I \sum_{i=1}^{u+l} K_{x_i} (M\mathbf{f}_{\mathbf{z},\gamma})_i = \sum_{i=1}^l K_{x_i} C^* y_i,$$

which we rewrite as

$$f_{\mathbf{z},\gamma} = -\frac{\gamma_I}{\gamma_A} \sum_{i=1}^{u+l} K_{x_i} (M\mathbf{f}_{\mathbf{z},\gamma})_i + \sum_{i=1}^l K_{x_i} \frac{C^* y_i - C^* C K_{x_i}^* f_{\mathbf{z},\gamma}}{l\gamma_A}.$$

This shows that there are vectors $a_i$'s in $\mathcal{W}$ such that

$$f_{\mathbf{z},\gamma} = \sum_{i=1}^{u+l} K_{x_i} a_i.$$

We have $f_{\mathbf{z},\gamma}(x_i) = \sum_{j=1}^{u+l} K(x_i, x_j) a_j$, and

$$(M\mathbf{f}_{\mathbf{z},\gamma})_i = \sum_{k=1}^{u+l} M_{ik} \sum_{j=1}^{u+l} K(x_k, x_j) a_j = \sum_{j,k=1}^{u+l} M_{ik} K(x_k, x_j) a_j.$$

Also $K_{x_i}^* f_{\mathbf{z},\gamma} = f_{\mathbf{z},\gamma}(x_i) = \sum_{j=1}^{u+l} K(x_i, x_j)a_j$. Thus for $1 \leq i \leq l$:

$$a_i = -\frac{\gamma_I}{\gamma_A} \sum_{j,k=1}^{u+l} M_{ik}K(x_k, x_j)a_j + \frac{C^* y_i - C^* C(\sum_{j=1}^{u+l} K(x_i, x_j)a_j)}{l\gamma_A},$$

which gives the formula

$$l\gamma_I \sum_{j,k=1}^{u+l} M_{ik}K(x_k, x_j)a_j + C^* C(\sum_{j=1}^{u+l} K(x_i, x_j)a_j) + l\gamma_A a_i = C^* y_i.$$

Similarly, for $l+1 \leq i \leq u+l$,

$$a_i = -\frac{\gamma_I}{\gamma_A} \sum_{j,k=1}^{u+l} M_{ik}K(x_k, x_j)a_j,$$

which is equivalent to

$$\gamma_I \sum_{j,k=1}^{u+l} M_{ik}K(x_k, x_j)a_j + \gamma_A a_i = 0.$$

This completes the proof. ∎

**Proof (first proof) of Theorem 4**. This is straightforward to obtain from Theorem 3 using the operator-valued matrix formulation described in the main paper. ∎

In the following, we give a second proof of Theorem 4, which is based entirely on operator-theoretic notations. The proof technique should be of interest in its own right.
**Proof (second proof) of Theorem 4**. By the Representer Theorem, (10) has a unique solution. Differentiating (110) and setting the derivative to zero gives

$$(E_{C,\mathbf{x}}^* E_{C,\mathbf{x}} + l\gamma_A I + l\gamma_I S_{\mathbf{x},u+l}^* M S_{\mathbf{x},u+l})f_{\mathbf{z},\gamma} = E_{C,\mathbf{x}}^* \mathbf{y}. \tag{111}$$

For $\gamma_A > 0, \gamma_I \geq 0$, the operator

$$E_{C,\mathbf{x}}^* E_{C,\mathbf{x}} + l\gamma_A I + l\gamma_I S_{\mathbf{x},u+l}^* M S_{\mathbf{x},u+l} \tag{112}$$

is clearly symmetric and strictly positive, so that the unique solution $f_{\mathbf{z},\gamma}$ is given by

$$f_{\mathbf{z},\gamma} = (E_{C,\mathbf{x}}^* E_{C,\mathbf{x}} + l\gamma_A I + l\gamma_I S_{\mathbf{x},u+l}^* M S_{\mathbf{x},u+l})^{-1} E_{C,\mathbf{x}}^* \mathbf{y}.$$

Recall the definitions of the operators $S_{\mathbf{x},u+l} : \mathcal{H}_K \to \mathcal{W}^{u+l}$ and $S_{\mathbf{x},u+l}^* : \mathcal{W}^{u+l} \to \mathcal{H}_K$:

$$S_{\mathbf{x},u+l}f = (K_{x_i}^* f)_{i=1}^{u+l}, \quad S_{\mathbf{x},u+l}^* \mathbf{b} = \sum_{i=1}^{u+l} K_{x_i} b_i$$

with the operator $S_{\mathbf{x},u+l}S^*_{\mathbf{x},u+l} : \mathcal{W}^{u+l} \to \mathcal{W}^{u+l}$ given by

$$S_{\mathbf{x},u+l}S^*_{\mathbf{x},u+l}\mathbf{b} = \left( K^*_{x_i} \left( \sum_{j=1}^{u+l} K_{x_j}b_j \right) \right)_{i=1}^{u+l} = \left( \sum_{j=1}^{u+l} K(x_i,x_j)b_j \right)_{i=1}^{u+l} = K[\mathbf{x}]\mathbf{b},$$

so that

$$S_{\mathbf{x},u+l}S^*_{\mathbf{x},u+l} = K[\mathbf{x}].$$

The operator $E_{C,\mathbf{x}} : \mathcal{H}_K \to \mathcal{Y}^l$ is

$$E_{C,\mathbf{x}}f = (CK^*_{x_i}f)_{i=1}^l = (I^T_{(u+l)\times l} \otimes C)S_{\mathbf{x},u+l}f,$$

so that

$$E_{C,\mathbf{x}} = (I^T_{(u+l)\times l} \otimes C)S_{\mathbf{x},u+l}, \tag{113}$$

and the operator $E^*_{C,\mathbf{x}} : \mathcal{Y}^l \to \mathcal{H}_K$ is

$$E^*_{C,\mathbf{x}} = S^*_{\mathbf{x},u+l}(I_{(u+l)\times l} \otimes C^*). \tag{114}$$

As operators, $I_{(u+l)\times l} \otimes C^* : \mathcal{Y}^l \to \mathcal{W}^{u+l}$ and $I^T_{(u+l)\times l} \otimes C : \mathcal{W}^{u+l} \to \mathcal{Y}^l$. The operator $E^*_{C,\mathbf{x}}E_{C,\mathbf{x}} : \mathcal{H}_K \to \mathcal{H}_K$ is then given by

$$E^*_{C,\mathbf{x}}E_{C,\mathbf{x}} = S^*_{\mathbf{x},u+l}(J^{u+l}_l \otimes C^*C)S_{\mathbf{x},u+l} : \mathcal{H}_K \to \mathcal{H}_K, \tag{115}$$

where $J^{u+l}_l = I_{(u+l)\times l}I^T_{(u+l)\times l}$ is the $(u+l)\times(u+l)$ diagonal matrix, with the first $l$ entries on the main diagonal being 1, and the rest 0. As an operator, $J^{u+l}_l \otimes C^*C : \mathcal{W}^{u+l} \to \mathcal{W}^{u+l}$. The operator $E_{C,\mathbf{x}}E^*_{C,\mathbf{x}} : \mathcal{W}^{u+l} \to \mathcal{W}^{u+l}$ is given by

$$E_{C,\mathbf{x}}E^*_{C,\mathbf{x}} = (I^T_{(u+l)\times l} \otimes C)S_{\mathbf{x},u+l}S^*_{\mathbf{x},u+l}(I_{(u+l)\times l} \otimes C^*) = (I^T_{(u+l)\times l} \otimes C)K[\mathbf{x}](I_{(u+l)\times l} \otimes C^*). \tag{116}$$

Equation (111) becomes

$$\left[ S^*_{\mathbf{x},u+l}(J^{u+l}_l \otimes C^*C + l\gamma_I M)S_{\mathbf{x},u+l} + l\gamma_A I \right] f_{\mathbf{z},\gamma} = S^*_{\mathbf{x},u+l}(I_{(u+l)\times l} \otimes C^*)\mathbf{y}, \tag{117}$$

which gives

$$f_{\mathbf{z},\gamma} = S^*_{\mathbf{x},u+l} \left[ \frac{-(J^{u+l}_l \otimes C^*C + l\gamma_I M)S_{\mathbf{x},u+l}f_{\mathbf{z},\gamma} + (I_{(u+l)\times l} \otimes C^*)\mathbf{y}}{l\gamma_A} \right] \tag{118}$$

$$= S^*_{\mathbf{x},u+l}\mathbf{a}, \tag{119}$$

where $\mathbf{a} = (a_i)_{i=1}^{u+l} \in \mathcal{W}^{u+l}$ is

$$\mathbf{a} = \frac{-(J^{u+l}_l \otimes C^*C + l\gamma_I M)S_{\mathbf{x},u+l}f_{\mathbf{z},\gamma} + (I_{(u+l)\times l} \otimes C^*)\mathbf{y}}{l\gamma_A}. \tag{120}$$

By definition of $S_{\mathbf{x},u+l}$ and $S_{\mathbf{x},u+l}^*$,

$$S_{\mathbf{x},u+l}f_{\mathbf{z},\gamma} = S_{\mathbf{x},u+l}S_{\mathbf{x},u+l}^*\mathbf{a} = K[\mathbf{x}]\mathbf{a}.$$

Substituting this into Equation (120), we obtain

$$\mathbf{a} = \frac{-(J_l^{u+l} \otimes C^*C + l\gamma_I M)K[\mathbf{x}]\mathbf{a} + (I_{(u+l)\times l} \otimes C^*)\mathbf{y}}{l\gamma_A},$$

or equivalently

$$[(J_l^{u+l} \otimes C^*C + l\gamma_I M)K[\mathbf{x}] + l\gamma_A I_{\mathcal{W}^{u+l}}]\mathbf{a} = (I_{(u+l)\times l} \otimes C^*)\mathbf{y}. \tag{121}$$

The operator-valued matrix on the left hand side,

$$(J_l^{u+l} \otimes C^*C + l\gamma_I M)K[\mathbf{x}] + l\gamma_A I_{\mathcal{W}^{u+l}} : \mathcal{W}^{u+l} \to \mathcal{W}^{u+l},$$

is invertible by Lemma 25, with a bounded inverse. Thus the above system of linear equations always has a unique solution

$$\mathbf{a} = [(J_l^{u+l} \otimes C^*C + l\gamma_I M)K[\mathbf{x}] + l\gamma_A I_{\mathcal{W}^{u+l}}]^{-1}(I_{(u+l)\times l} \otimes C^*)\mathbf{y}.$$

This completes the proof of the theorem. ∎

**Remark 23 (Uniqueness of a)** . *While the solution $f_{\mathbf{z},\gamma} = \sum_{i=1}^{u+l} K_{x_i}a_i$ in Theorem 2 is always unique, the expansion coefficient vectors $a_i$'s for $f_{\mathbf{z},\gamma}$ need not be unique. In fact, we have*

$$||f_{\mathbf{z},\gamma}||_{\mathcal{H}_K}^2 = \langle S_{\mathbf{x},u+l}^*\mathbf{a}, S_{\mathbf{x},u+l}^*\mathbf{a}\rangle_{\mathcal{H}_K} = \langle \mathbf{a}, S_{\mathbf{x},u+l}S_{\mathbf{x},u+l}^*\mathbf{a}\rangle_{\mathcal{W}^{u+l}} = \langle \mathbf{a}, K[\mathbf{x}]\mathbf{a}\rangle_{\mathcal{W}^{u+l}}.$$

*By the reproducing property,*

$$f_{\mathbf{z},\gamma} = 0 \Longleftrightarrow ||f_{\mathbf{z},\gamma}||_{\mathcal{H}_K} = 0 \Longleftrightarrow \mathbf{a} = 0 \text{ or } \mathbf{a} \in \text{null}(K[\mathbf{x}]).$$

*Thus $\mathbf{a}$ is unique if and only if $K[\mathbf{x}]$ is invertible, or equivalently, $K[\mathbf{x}]$ is of full rank. For us, our choice for $\mathbf{a}$ is always the unique solution of the system of linear equations (13) in Theorem 4 (see also Remark 24 below).*

**Remark 24** *The coefficient matrix of the system of linear equations (13) in Theorem 4 has the form $(\gamma I + AB)$, where $A, B$ are two symmetric, positive operators on a Hilbert space $\mathcal{H}$. We show in Lemma 25 that the operator $(\gamma I + AB)$ is always invertible for $\gamma > 0$ and that the inverse operator $(\gamma I + AB)^{-1}$ is bounded, so that the system (13) is always guaranteed a unique solution, as we claim in Theorem 4. Furthermore, the eigenvalues of $AB$, when they exist, are always non-negative, as we show in Lemma 26. This gives another proof of the invertibility of $(\gamma I + AB)$ when $\mathcal{H}$ is finite-dimensional, in Corollary 27. This invertibility is also necessary for the proofs of Theorems 6 and 7 in the SVM case.*

Минh, Bazzani, and Murino

**Lemma 25** *Let $\mathcal{H}$ be a Hilbert space and $A, B : \mathcal{H} \to \mathcal{H}$ be two bounded, symmetric, positive operators. Then the operator $(\gamma I + AB)$ is invertible for any $\gamma > 0$ and the inverse $(\gamma I + AB)^{-1}$ is bounded.*

**Proof** Let $T = \gamma I + AB$. We need to show that $T$ is 1-to-1 and onto. First, to show that $T$ is 1-to-1, suppose that

$$Tx = \gamma x + ABx = 0.$$

This implies that

$$BTx = \gamma Bx + BABx = 0 \implies \langle x, BTx \rangle = \gamma \langle x, Bx \rangle + \langle x, BABx \rangle = 0.$$

By the symmetry and positivity of $A$ and $B$, this is equivalent to

$$\gamma ||B^{1/2}x||^2 + ||A^{1/2}Bx||^2 = 0.$$

This is possible if and only if $x = 0$ or $B^{1/2}x = 0$. If $B^{1/2}x = 0$, $x \neq 0$, then $Tx = \gamma x \neq 0$. Thus

$$Tx = 0 \Longleftrightarrow x = 0.$$

This shows that $T$ is 1-to-1. Similar arguments show that its adjoint $T^* = \gamma I + BA$ is 1-to-1, so that

$$\overline{\text{range}(T)} = (\ker(T^*))^\perp = \{0\}^\perp = \mathcal{H}.$$

It thus remains for us to show that range($T$) is closed. Let $\{y_n\}_{n \in \mathbb{N}}$ be a Cauchy sequence in range($T$), with $y_n = Tx_n$ for $x_n \in \mathcal{H}$. Then we have

$$By_n = \gamma Bx_n + BABx_n \implies \langle x_n, By_n \rangle = \gamma \langle x_n, Bx_n \rangle + \langle x_n, BABx_n \rangle.$$

By the symmetry and positivity of $A$ and $B$, this is

$$\langle x_n, By_n \rangle = \gamma ||B^{1/2}x_n||^2 + ||A^{1/2}Bx_n||^2.$$

It follows that

$$\gamma ||B^{1/2}x_n||^2 \leq \langle x_n, By_n \rangle \leq ||B^{1/2}x_n|| \, ||B^{1/2}y_n||,$$

so that

$$\gamma ||B^{1/2}x_n|| \leq ||B^{1/2}y_n|| \leq ||B^{1/2}|| \, ||y_n||.$$

From the assumption $y_n = Tx_n = \gamma x_n + ABx_n$, we have

$$\gamma x_n = y_n - ABx_n.$$

This implies that

$$\gamma ||x_n|| \leq ||y_n|| + ||AB^{1/2}|| \, ||B^{1/2}x_n|| \leq ||y_n|| + \frac{||AB^{1/2}|| \, ||B^{1/2}||}{\gamma} \, ||y_n||,$$

which simplifies to

$$||x_n|| \leq \frac{1}{\gamma} \left( 1 + \frac{||AB^{1/2}|| \, ||B^{1/2}||}{\gamma} \right) ||y_n||.$$

Since $T$ is linear, $y_{n+1} - y_n = T(x_{n+1} - x_n)$ and thus

$$||x_{n+1} - x_n|| \leq \frac{1}{\gamma} \left( 1 + \frac{||AB^{1/2}|| \; ||B^{1/2}||}{\gamma} \right) ||y_{n+1} - y_n||.$$

Thus if $\{y_n\}_{n \in \mathbb{N}}$ is a Cauchy sequence in $\mathcal{H}$, then $\{x_n\}_{n \in \mathbb{N}}$ is also a Cauchy sequence in $\mathcal{H}$. Let $x_0 = \lim_{n \to \infty} x_n$ and $y_0 = Tx_0$, then clearly $\lim_{n \to \infty} y_n = y_0$. This shows that range$(T)$ is closed, as we claimed, so that range$(T) = \overline{\text{range}(T)} = \mathcal{H}$, showing that $T$ is onto. This completes the proof. ∎

**Lemma 26** *Let $\mathcal{H}$ be a Hilbert space. Let $A$ and $B$ be two symmetric, positive, bounded operators in $\mathcal{L}(\mathcal{H})$. Then all eigenvalues of the product operator $AB$ are real and non-negative.*

**Proof** Let $\lambda$ be an eigenvalue of $AB$, corresponding to eigenvector $x$. Then

$$ABx = \lambda x \implies BABx = \lambda Bx \implies \langle x, BABx \rangle = \lambda \langle x, Bx \rangle.$$

Since both $A$ and $B$ are symmetric, positive, the operator $BAB$ is symmetric, positive, and therefore $\langle x, BABx \rangle \geq 0$. Since $B$ is symmetric, positive, we have $\langle x, Bx \rangle \geq 0$, with $\langle x, Bx \rangle = ||B^{1/2}x||^2 = 0$ if and only if $x \in \text{null}(B^{1/2})$.

If $x \in \text{null}(B^{1/2})$, then $ABx = 0$, so that $\lambda = 0$.

If $x \notin \text{null}(B^{1/2})$, then $\langle x, Bx \rangle > 0$, and

$$\lambda = \frac{\langle x, BABx \rangle}{\langle x, Bx \rangle} \geq 0.$$

Consequently, we always have $\lambda \geq 0$. ∎

**Corollary 27** *Let $A$ and $B$ be two symmetric positive semi-definite matrices. Then the matrix $(\gamma I + AB)$ is invertible for any $\gamma > 0$.*

**Proof** The eigenvalues of $(\gamma I + AB)$ have the form $\gamma + \lambda$, where $\lambda$ is an eigenvalue of $AB$ and satisfies $\lambda \geq 0$ by Lemma 26. Thus all eigenvalues of $(\gamma I + AB)$ are strictly positive, with magnitude at least $\gamma$. It follows that $\det(\gamma I + AB) > 0$ and therefore $(\gamma I + AB)$ is invertible. ∎

**Proof of Theorem 10**. Recall some properties of the Kronecker tensor product:

$$(A \otimes B)(C \otimes D) = AC \otimes BD, \tag{122}$$

$$(A \otimes B)^T = A^T \otimes B^T, \tag{123}$$

and

$$\text{vec}(ABC) = (C^T \otimes A)\text{vec}(B). \tag{124}$$

Thus the equation

$$AXB = C \tag{125}$$

is equivalent to

$$(B^T \otimes A)\text{vec}(X) = \text{vec}(C). \tag{126}$$

In our context, $\gamma_I M = \gamma_B M_B + \gamma_W M_W$, which is

$$\gamma_I M = \gamma_B I_{u+l} \otimes M_m \otimes I_{\mathcal{Y}} + \gamma_W L \otimes I_{\mathcal{Y}}.$$

Using the property stated in Equation (123), we have for $C = \mathbf{c}^T \otimes I_{\mathcal{Y}}$,

$$C^* C = (\mathbf{c} \otimes I_{\mathcal{Y}})(\mathbf{c}^T \otimes I_{\mathcal{Y}}) = (\mathbf{c}\mathbf{c}^T \otimes I_{\mathcal{Y}}). \tag{127}$$

So then

$$\mathbf{C}^* \mathbf{C} = (I_{u+l} \otimes \mathbf{c}\mathbf{c}^T \otimes I_{\mathcal{Y}}). \tag{128}$$

$$J_l^{\mathcal{W}, u+l} = J_l^{u+l} \otimes I_m \otimes I_{\mathcal{Y}}. \tag{129}$$

It follows that

$$\mathbf{C}^* \mathbf{C} J_l^{\mathcal{W}, u+l} = (J_l^{u+l} \otimes \mathbf{c}\mathbf{c}^T \otimes I_{\mathcal{Y}}). \tag{130}$$

Then with

$$K[\mathbf{x}] = G[\mathbf{x}] \otimes R,$$

we have

$$\mathbf{C}^* \mathbf{C} J_l^{\mathcal{W}, u+l} K[\mathbf{x}] = (J_l^{u+l} \otimes \mathbf{c}\mathbf{c}^T) G[\mathbf{x}] \otimes R.$$

$$\gamma_I M K[\mathbf{x}] = (\gamma_B I_{u+l} \otimes M_m + \gamma_W L) G[\mathbf{x}] \otimes R.$$

Consider again now the system

$$(\mathbf{C}^* \mathbf{C} J_l^{\mathcal{W}, u+l} K[\mathbf{x}] + l\gamma_I M K[\mathbf{x}] + l\gamma_A I)\mathbf{a} = \mathbf{C}^* \mathbf{y}.$$

The left hand side is

$$(B \otimes R + l\gamma_A I_{(u+l)m} \otimes I_{\mathcal{Y}})\text{vec}(A^T),$$

where $\mathbf{a} = \text{vec}(A^T)$, $A$ is of size $(u+l)m \times \dim(\mathcal{Y})$, and

$$B = \left( (J_l^{u+l} \otimes \mathbf{c}\mathbf{c}^T) + l\gamma_B (I_{u+l} \otimes M_m) + l\gamma_W L \right) G[\mathbf{x}].$$

Then we have the linear system

$$(B \otimes R + l\gamma_A I_{(u+l)m} \otimes I_{\mathcal{Y}})\text{vec}(A^T) = \text{vec}(Y_C^T),$$

which, by properties (125) and (126), is equivalent to

$$RA^T B^T + l\gamma_A A^T = Y_C^T \iff BAR + l\gamma_A A = Y_C.$$

This completes the proof. ∎

**Remark 28** *The* vec *operator is implemented by the flattening operation* (:) *in MATLAB. To compute the matrix* $Y_C^T$, *note that by definition*

$$\text{vec}(Y_C^T) = \mathbf{C}^* \mathbf{y} = (I_{(u+l) \times l} \otimes C^*) \mathbf{y} = \text{vec}(C^* Y_l I_{(u+l) \times l}^T) = \text{vec}(C^* Y_{u+l}),$$

*where* $Y_l$ *is the* $\dim(\mathcal{Y}) \times l$ *matrix, whose ith column is* $y_i$, $1 \leq i \leq l$, *that is*

$$Y_l = [y_1, \ldots, y_l], \quad \text{with} \quad \mathbf{y} = \text{vec}(Y_l),$$

*and* $Y_{u+l}$ *is the* $\dim(\mathcal{Y}) \times (u+l)$ *matrix with the ith column being* $y_i$, $1 \leq i \leq l$, *with the remaining u columns being zero, that is*

$$Y_{u+l} = [y_1, \ldots, y_l, 0, \ldots, 0] = [Y_l, 0, \ldots, 0] = Y_l I_{(u+l) \times l}^T.$$

*Note that* $Y_C^T$ *and* $C^* Y_{u+l}$ *in general are not the same:* $Y_C^T$ *is of size* $\dim(\mathcal{Y}) \times (u+l)m$, *whereas* $C^* Y_{u+l}$ *is of size* $\dim(\mathcal{Y})m \times (u+l)$.

**Proof of Corollary 19** For $\gamma_I = 0$, $u = 0$, Equation (111) becomes

$$(E_{C,\mathbf{x}}^* E_{C,\mathbf{x}} + l\gamma_A I) f_{\mathbf{z},\gamma} = E_{C,\mathbf{x}}^* \mathbf{y},$$

which is equivalent to

$$f_{\mathbf{z},\gamma} = (E_{C,\mathbf{x}}^* E_{C,\mathbf{x}} + l\gamma_A I_{\mathcal{H}_K})^{-1} E_{C,\mathbf{x}}^* \mathbf{y} = E_{C,\mathbf{x}}^* (E_{C,\mathbf{x}} E_{C,\mathbf{x}}^* + l\gamma_A I_{\mathcal{Y}^l})^{-1} \mathbf{y},$$

that is

$$f_{\mathbf{z},\gamma} = S_{\mathbf{x},l}^* (I_l \otimes C^*) \left[ (I_l \otimes C)K[\mathbf{x}](I_l \otimes C^*) + l\gamma_A I_{\mathcal{Y}^l} \right]^{-1} \mathbf{y}.$$

Thus in this case $f_{\mathbf{z},\gamma} = S_{\mathbf{x},l}^* \mathbf{a}$, where $\mathbf{a} = (a_i)_{i=1}^l$ is given by

$$\mathbf{a} = (I_l \otimes C^*) \left[ (I_l \otimes C)K[\mathbf{x}](I_l \otimes C^*) + l\gamma_A I_{\mathcal{Y}^l} \right]^{-1} \mathbf{y}.$$

In this expression, the operator $[(I_l \otimes C)K[\mathbf{x}](I_l \otimes C^*) + l\gamma_A] : \mathcal{Y}^l \to \mathcal{Y}^l$ is clearly symmetric and strictly positive, hence is invertible. For $C = \mathbf{c}^T \otimes I_{\mathcal{Y}}$ and $K[\mathbf{x}] = G[\mathbf{x}] \otimes R$, we have

$$\mathbf{a} = (I_l \otimes \mathbf{c} \otimes I_{\mathcal{Y}}) \left[ (I_l \otimes \mathbf{c}^T)G[\mathbf{x}](I_l \otimes \mathbf{c}) \otimes R + l\gamma_A I_{\mathcal{Y}^l} \right]^{-1} \mathbf{y}.$$

With $R = I_{\mathcal{Y}}$, this becomes

$$\mathbf{a} = \{(I_l \otimes \mathbf{c}) \left[ (I_l \otimes \mathbf{c}^T)G[\mathbf{x}](I_l \otimes \mathbf{c}) + l\gamma_A I_l \right]^{-1} \otimes I_{\mathcal{Y}}\} \mathbf{y}.$$

For any $v \in \mathcal{X}$,

$$f_{\mathbf{z},\gamma}(v) = K[v, \mathbf{x}]\mathbf{a} = \{G[v, \mathbf{x}](I_l \otimes \mathbf{c}) \left[ (I_l \otimes \mathbf{c}^T)G[\mathbf{x}](I_l \otimes \mathbf{c}) + l\gamma_A I_l \right]^{-1} \otimes I_{\mathcal{Y}}\} \mathbf{y}.$$

$$C f_{\mathbf{z},\gamma}(v) = \{\mathbf{c}^T G[v, \mathbf{x}](I_l \otimes \mathbf{c}) \left[ (I_l \otimes \mathbf{c}^T)G[\mathbf{x}](I_l \otimes \mathbf{c}) + l\gamma_A I_l \right]^{-1} \otimes I_{\mathcal{Y}}\} \mathbf{y}.$$

With $G[\mathbf{x}] = \sum_{i=1}^m k^i[\mathbf{x}] \otimes \mathbf{e}_i \mathbf{e}_i^T$, we have

$$(I_l \otimes \mathbf{c}^T)G[\mathbf{x}](I_l \otimes \mathbf{c}) = (I_l \otimes \mathbf{c}^T)(\sum_{i=1}^m k^i[\mathbf{x}] \otimes \mathbf{e}_i \mathbf{e}_i^T)(I_l \otimes \mathbf{c}) = \sum_{i=1}^m c_i^2 k^i[\mathbf{x}],$$

$$\mathbf{c}^T G[v, \mathbf{x}](I_l \otimes \mathbf{c}) = \mathbf{c}^T (\sum_{i=1}^m k^i[v, \mathbf{x}] \otimes \mathbf{e}_i \mathbf{e}_i^T)(I_l \otimes \mathbf{c}) = \sum_{i=1}^m c_i^2 k^i[v, \mathbf{x}].$$

With these, we obtain

$$C f_{\mathbf{z}, \gamma}(v) = \left\{ \sum_{i=1}^m c_i^2 k^i[v, \mathbf{x}] \left( \sum_{i=1}^m c_i^2 k^i[\mathbf{x}] + l \gamma_A I_l \right)^{-1} \otimes I_{\mathcal{Y}} \right\} \mathbf{y}.$$

In particular, for $\mathcal{Y} = \mathbb{R}$, we obtain

$$C f_{\mathbf{z}, \gamma}(v) = \left\{ \sum_{i=1}^m c_i^2 k^i[v, \mathbf{x}] \left( \sum_{i=1}^m c_i^2 k^i[\mathbf{x}] + l \gamma_A I_l \right)^{-1} \right\} \mathbf{y}.$$

This completes the proof. ∎

**Proof of Lemma 22** Consider the function $f \in \mathcal{H}_K$ of the form

$$f(x) = \sum_{i=1}^m K(x, x_i) a_i = \sum_{i=1}^m [R \otimes G(x, x_i)] a_i \in \mathcal{Y}^n,$$

where $a_i \in \mathcal{Y}^n$. Let $A_i$ be the (potentially infinite) matrix of size $\dim(\mathcal{Y}) \times n$ such that $a_i = \text{vec}(A_i)$. Then

$$f(x) = \sum_{i=1}^m [R \otimes G(x, x_i)] \text{vec}(A_i) = \sum_{i=1}^m \text{vec}(G(x, x_i) A_i R),$$

with norm

$$||f||^2_{\mathcal{H}_K} = \sum_{i,j=1}^m \langle a_i, K(x_i, x_j) a_j \rangle_{\mathcal{Y}^n} = \sum_{i,j=1}^m \langle a_i, (R \otimes G(x_i, x_j)) a_j \rangle_{\mathcal{Y}^n}$$

$$= \sum_{i,j=1}^m \langle \text{vec}(A_i), \text{vec}(G(x_i, x_j) A_j R) \rangle_{\mathcal{Y}^n} = \sum_{i,j=1}^m \text{tr}(A_i^T G(x_i, x_j) A_j R).$$

Each component $f^k$, $1 \leq k \leq n$, has the form

$$f^k(x) = \sum_{i=1}^m G(x, x_i) A_i R_{:,k} \in \mathcal{H}_G,$$

where $R_{:,k}$ is the $k$th column of $R$, with norm

$$||f^k||^2_{\mathcal{H}_G} = \sum_{i,j=1}^m \langle A_i R_{:,k}, G(x_i, x_j) A_j R_{:,k} \rangle_{\mathcal{Y}} = \sum_{i,j=1}^m R_{:,k}^T A_i^T G(x_i, x_j) A_j R_{:,k}.$$

For

$$f^l(x) = \sum_{i=1}^m G(x, x_i) A_i R_{:,l},$$

we have

$$\langle f^k, f^l \rangle_{\mathcal{H}_G} = \sum_{i,j=1}^m R_{:,k}^T A_i^T G(x_i, x_j) A_j R_{:,l}.$$

Let $B$ be a symmetric, positive definite matrix of size $n \times n$. Consider the form

$$\sum_{k,l=1}^n B_{kl} \langle f^k, f^l \rangle_{\mathcal{H}_G} = \sum_{k,l=1}^n \sum_{i,j=1}^m B_{kl} R_{:,k}^T A_i^T G(x_i, x_j) A_j R_{:,l}$$

$$= \sum_{i,j=1}^m \sum_{k,l=1}^n B_{kl} R_{:,k}^T A_i^T G(x_i, x_j) A_j R_{:,l} = \sum_{i,j=1}^m \text{tr}(B R^T A_i^T G(x_i, x_j) A_j R)$$

$$= \sum_{i,j=1}^m \text{tr}(B R A_i^T G(x_i, x_j) A_j R), \quad \text{since } R \text{ is symmetric.}$$

It follows that for $R$ strictly positive definite and $B = R^{-1}$, we have

$$||f||_{\mathcal{H}_K}^2 = \sum_{k,l=1}^n B_{kl} \langle f^k, f^l \rangle_{\mathcal{H}_G}.$$

In particular, for $0 < \lambda \le 1$ and

$$R = I_n + \frac{1-\lambda}{n\lambda} \mathbf{1}_n \mathbf{1}_n^T,$$

we have

$$B = R^{-1} = I_n - \frac{1-\lambda}{n} \mathbf{1}_n \mathbf{1}_n^T.$$

Then

$$||f||_{\mathcal{H}_K}^2 = \sum_{k,l=1}^n B_{kl} \langle f^k, f^l \rangle_{\mathcal{H}_G} = \sum_{k=1}^n ||f^k||_{\mathcal{H}_G}^2 - \frac{1-\lambda}{n} \sum_{k,l=1}^n \langle f^k, f^l \rangle_{\mathcal{H}_G}$$

$$= \lambda \sum_{k=1}^n ||f^k||_{\mathcal{H}_G}^2 + (1-\lambda) \sum_{k=1}^n ||f^k - \frac{1}{n} \sum_{l=1}^n f^l||_{\mathcal{H}_G}^2.$$

This result then extends to all $f \in \mathcal{H}_K$ by a limiting argument. This completes the proof. ■

### A.3 Proofs for the SVM case

Recall the optimization problem that we aim to solve

$$f_{\mathbf{z},\gamma} = \text{argmin}_{f \in \mathcal{H}_K, \xi_{ki} \in \mathbb{R}} \frac{1}{l} \sum_{i=1}^l \sum_{k=1, k \neq y_i}^P \xi_{ki} + \gamma_A ||f||_{\mathcal{H}_K}^2 + \gamma_I \langle \mathbf{f}, M\mathbf{f} \rangle_{\mathcal{W}^{u+l}}, \quad (131)$$

subject to the constraints

$$\xi_{ki} \ge -\langle s_k, s_{y_i} \rangle_{\mathcal{Y}} + \langle s_k, Cf(x_i) \rangle_{\mathcal{Y}}, \quad 1 \le i \le l, k \neq y_i,$$
$$\xi_{ki} \ge 0, \quad 1 \le i \le l, k \neq y_i. \quad (132)$$

**Proof of Theorem 6** The Lagrangian is

$$L(f, \xi, \alpha, \beta) = \frac{1}{l} \sum_{i=1}^{l} \sum_{k \neq y_i} \xi_{ki} + \gamma_A ||f||_{\mathcal{H}_K}^2 + \gamma_I \langle \mathbf{f}, M\mathbf{f} \rangle_{\mathcal{W}^{u+l}}$$

$$- \sum_{i=1}^{l} \sum_{k \neq y_i} \alpha_{ki} \left( \xi_{ki} - [-\langle s_k, s_{y_i} \rangle_{\mathcal{Y}} + \langle s_k, Cf(x_i) \rangle_{\mathcal{Y}}] \right) - \sum_{i=1}^{l} \sum_{k \neq y_i} \beta_{ki} \xi_{ki}, \tag{133}$$

where

$$\alpha_{ki} \geq 0, \beta_{ki} \geq 0, \quad 1 \leq i \leq l, k \neq y_i. \tag{134}$$

By the reproducing property

$$\langle s_k, Cf(x_i) \rangle_{\mathcal{Y}} = \langle C^* s_k, f(x_i) \rangle_{\mathcal{W}} = \langle f, K_{x_i}(C^* s_k) \rangle_{\mathcal{H}_K}. \tag{135}$$

Thus the Lagrangian is

$$L(f, \xi, \alpha, \beta) = \frac{1}{l} \sum_{i=1}^{l} \sum_{k \neq y_i} \xi_{ki} + \gamma_A ||f||_{\mathcal{H}_K}^2 + \gamma_I \langle \mathbf{f}, M\mathbf{f} \rangle_{\mathcal{W}^{u+l}}$$

$$- \sum_{i=1}^{l} \sum_{k \neq y_i} \alpha_{ki} \left( \xi_{ki} - [-\langle s_k, s_{y_i} \rangle_{\mathcal{Y}} + \langle f, K_{x_i}(C^* s_k) \rangle_{\mathcal{H}_K}] \right) - \sum_{i=1}^{l} \sum_{k \neq y_i} \beta_{ki} \xi_{ki}, \tag{136}$$

Since

$$\langle \mathbf{f}, M\mathbf{f} \rangle_{\mathcal{W}^{u+l}} = \langle S_{\mathbf{x}, u+l} f, M S_{\mathbf{x}, u+l} f \rangle_{\mathcal{W}^{u+l}} = \langle f, S_{\mathbf{x}, u+l}^* M S_{\mathbf{x}, u+l} f \rangle_{\mathcal{H}_K}, \tag{137}$$

we have

$$\frac{\langle \mathbf{f}, M\mathbf{f} \rangle_{\mathcal{W}^{u+l}}}{\partial f} = 2 S_{\mathbf{x}, u+l}^* M S_{\mathbf{x}, u+l} f = 2 \sum_{i=1}^{u+l} K_{x_i}(M\mathbf{f})_i. \tag{138}$$

Differentiating the Lagrangian with respect to $\xi_{ki}$ and setting to zero, we obtain

$$\frac{\partial L}{\partial \xi_{ki}} = \frac{1}{l} - \alpha_{ki} - \beta_{ki} = 0 \iff \alpha_{ki} + \beta_{ki} = \frac{1}{l}. \tag{139}$$

Differentiating the Lagrangian with respect to $f$, we obtain

$$\frac{\partial L}{\partial f} = 2\gamma_A f + 2\gamma_I S_{\mathbf{x}, u+l}^* M S_{\mathbf{x}, u+l} f + \sum_{i=1}^{l} \sum_{k \neq y_i} \alpha_{ki} K_{x_i}(C^* s_k). \tag{140}$$

Setting this derivative to zero, we obtain

$$f = -\frac{\gamma_I}{\gamma_A} \sum_{i=1}^{u+l} K_{x_i}(M\mathbf{f})_i - \frac{1}{2\gamma_A} \sum_{i=1}^{l} \sum_{k \neq y_i} \alpha_{ki} K_{x_i}(C^* s_k). \tag{141}$$

This means there are vectors $a_i \in \mathcal{W}$, $1 \leq i \leq u+l$, such that

$$f = \sum_{i=1}^{u+l} K_{x_i} a_i. \tag{142}$$

This gives

$$\mathbf{f}_k = f(x_k) = \sum_{j=1}^{u+l} K(x_k, x_j) a_j, \tag{143}$$

so that

$$(M\mathbf{f})_i = \sum_{k=1}^{u+l} M_{ik} \mathbf{f}_k = \sum_{k=1}^{u+l} M_{ik} \sum_{j=1}^{u+l} K(x_k, x_j) a_j = \sum_{j,k=1}^{u+l} M_{ik} K(x_k, x_j) a_j. \tag{144}$$

For $1 \le i \le l$,

$$a_i = -\frac{\gamma_I}{\gamma_A} \sum_{j,k=1}^{u+l} M_{ik} K(x_k, x_j) a_j - \frac{1}{2\gamma_A} \sum_{k \neq y_i} \alpha_{ki}(C^* s_k), \tag{145}$$

or equivalently,

$$\gamma_I \sum_{j,k=1}^{u+l} M_{ik} K(x_k, x_j) a_j + \gamma_A a_i = -\frac{1}{2} \sum_{k \neq y_i} \alpha_{ki}(C^* s_k) = -\frac{1}{2} C^* S \alpha_i, \tag{146}$$

since $\alpha_{y_i,i} = 0$. For $l + 1 \le i \le u + l$,

$$a_i = -\frac{\gamma_I}{\gamma_A} \sum_{j,k=1}^{u+l} M_{ik} K(x_k, x_j) a_j, \tag{147}$$

or equivalently,

$$\gamma_I \sum_{j,k=1}^{u+l} M_{ik} K(x_k, x_j) a_j + \gamma_A a_i = 0. \tag{148}$$

In operator-valued matrix notation, (146) and (148) together can be expressed as

$$(\gamma_I M K[\mathbf{x}] + \gamma_A I)\mathbf{a} = -\frac{1}{2}(I_{(u+l) \times l} \otimes C^* S)\mathrm{vec}(\alpha). \tag{149}$$

By Lemma 25, the operator $(\gamma_I M K[\mathbf{x}] + \gamma_A I) : \mathcal{W}^{u+l} \to \mathcal{W}^{u+l}$ is invertible, with a bounded inverse, so that

$$\mathbf{a} = -\frac{1}{2}(\gamma_I M K[\mathbf{x}] + \gamma_A I)^{-1}(I_{(u+l) \times l} \otimes C^* S)\mathrm{vec}(\alpha). \tag{150}$$

With condition (139), the Lagrangian (136) simplifies to

$$L(f, \xi, \alpha, \beta) = \gamma_A ||f||_K^2 + \gamma_I \langle \mathbf{f}, M\mathbf{f} \rangle_{\mathcal{W}^{u+l}}$$
$$+ \sum_{i=1}^{l} \sum_{k \neq y_i} \alpha_{ki} \left( [-\langle s_k, s_{y_i} \rangle_{\mathcal{Y}} + \langle s_k, Cf(x_i) \rangle_{\mathcal{Y}} \right). \tag{151}$$

From expression (140), we have

$$\frac{\partial L}{\partial f} = 0 \iff \gamma_A f + \gamma_I S_{\mathbf{x},u+l}^* M S_{\mathbf{x},u+l} f = -\frac{1}{2} \sum_{i=1}^{l} \sum_{k \neq y_i} \alpha_{ki} K_{x_i}(C^* s_k). \tag{152}$$

Taking inner product with $f$ on both sides, we get

$$\gamma_A||f||^2_{\mathcal{H}_K} + \gamma_I \langle \mathbf{f}, M\mathbf{f} \rangle_{\mathcal{W}^{u+l}} = -\frac{1}{2} \sum_{i=1}^{l} \sum_{k \neq y_i} \alpha_{ki} \langle f, K_{x_i}(C^* s_k) \rangle_{\mathcal{H}_K}. \tag{153}$$

With $f = \sum_{j=1}^{u+l} K_{x_j} a_j$, we have

$$\langle f, K_{x_i}(C^* s_k) \rangle_{\mathcal{H}_K} = \sum_{j=1}^{u+l} \langle K(x_i, x_j) a_j, C^* s_k \rangle_{\mathcal{W}}, \tag{154}$$

so that

$$\sum_{k \neq y_i} \alpha_{ki} \langle f, K_{x_i}(C^* s_k) \rangle_{\mathcal{H}_K} = \sum_{j=1}^{u+l} \langle K(x_i, x_j) a_j, \sum_{k \neq y_i} \alpha_{ki} C^* s_k \rangle_{\mathcal{W}}$$
$$= \sum_{j=1}^{u+l} \langle K(x_i, x_j) a_j, C^* S \alpha_i \rangle_{\mathcal{W}}. \tag{155}$$

Combining this with (153), we obtain

$$\gamma_A||f||^2_{\mathcal{H}_K} + \gamma_I \langle \mathbf{f}, M\mathbf{f} \rangle_{\mathcal{W}^{u+l}} = -\frac{1}{2} \sum_{i=1}^{l} \langle \sum_{j=1}^{u+l} K(x_i, x_j) a_j, C^* S \alpha_i \rangle_{\mathcal{W}}$$
$$= -\frac{1}{2} \sum_{i=1}^{l} \langle S^* C \sum_{j=1}^{u+l} K(x_i, x_j) a_j, \alpha_i \rangle_{\mathbb{R}^P}. \tag{156}$$

In operator-valued matrix notation, this is

$$\gamma_A||f||^2_{\mathcal{H}_K} + \gamma_I \langle \mathbf{f}, M\mathbf{f} \rangle_{\mathcal{W}^{u+l}} = -\frac{1}{2} \text{vec}(\alpha)^T (I^T_{(u+l) \times l} \otimes S^* C) K[\mathbf{x}] \mathbf{a}. \tag{157}$$

Substituting the expression for $\mathbf{a}$ in (150) into (157), we obtain

$$\gamma_A||f||^2_{\mathcal{H}_K} + \gamma_I \langle \mathbf{f}, M\mathbf{f} \rangle_{\mathcal{W}^{u+l}} = \frac{1}{4} \text{vec}(\alpha)^T (I^T_{(u+l) \times l} \otimes S^* C) K[\mathbf{x}]$$
$$\times (\gamma_I M K[\mathbf{x}] + \gamma_A I)^{-1} (I_{(u+l) \times l} \otimes C^* S) \text{vec}(\alpha). \tag{158}$$

Combining (151), (153), and (158), we obtain the final form of the Lagrangian

$$L(\alpha) = -\sum_{i=1}^{l} \sum_{k \neq y_i} \langle s_k, s_{y_i} \rangle_{\mathcal{Y}} \alpha_{ki} - \frac{1}{4} \text{vec}(\alpha)^T Q[\mathbf{x}, C] \text{vec}(\alpha), \tag{159}$$

where the matrix $Q[\mathbf{x}, C]$ is given by

$$Q[\mathbf{x}, C] = (I^T_{(u+l) \times l} \otimes S^* C) K[\mathbf{x}] \tag{160}$$
$$\times (\gamma_I M K[\mathbf{x}] + \gamma_A I)^{-1} (I_{(u+l) \times l} \otimes C^* S). \tag{161}$$

We need to maximize the Lagrangian subject to the constraints

$$0 \leq \alpha_{ki} \leq \frac{1}{l}, \quad 1 \leq i \leq l, k \neq y_i. \tag{162}$$

Since $\alpha_{y_i,i} = 0$, these constraints can be written as

$$0 \leq \alpha_{ki} \leq \frac{1}{l}(1 - \delta_{k,y_i}), \quad 1 \leq i \leq l, 1 \leq k \leq P. \tag{163}$$

Equivalently, under the same constraints, we minimize

$$D(\alpha) = \frac{1}{4}\text{vec}(\alpha)^T Q[\mathbf{x}, C]\text{vec}(\alpha) + \sum_{i=1}^{l}\sum_{k=1}^{P}\langle s_k, s_{y_i}\rangle_{\mathcal{Y}}\alpha_{ki}. \tag{164}$$

When $S$ is the simplex coding, we have $\langle s_k, s_{y_i}\rangle_{\mathcal{Y}} = -\frac{1}{P-1}$ for $k \neq y_i$, and $\alpha_{y_i,i} = 0$, so that

$$\sum_{i=1}^{l}\sum_{k=1}^{P}\langle s_k, s_{y_i}\rangle_{\mathcal{Y}}\alpha_{ki} = -\frac{1}{P-1}\sum_{i=1}^{l}\sum_{k=1}^{P}\alpha_{ki} = -\frac{1}{P-1}\mathbf{1}_{Pl}^T\text{vec}(\alpha).$$

This gives the last expression of the theorem.

Let us show that $Q[\mathbf{x}, C]$ is symmetric and positive semidefinite. To show that $Q[\mathbf{x}, C]$ is symmetric, it suffices to show that $K[\mathbf{x}](\gamma_I M K[\mathbf{x}] + \gamma_A I)^{-1}$ is symmetric. We have

$$(\gamma_I K[\mathbf{x}]M + \gamma_A I)K[\mathbf{x}] = K[\mathbf{x}](\gamma_I M K[\mathbf{x}] + \gamma_A I),$$

which is equivalent to

$$K[\mathbf{x}](\gamma_I M K[\mathbf{x}] + \gamma_A I)^{-1} = (\gamma_I K[\mathbf{x}]M + \gamma_A I)^{-1}K[\mathbf{x}]$$
$$= (K[\mathbf{x}](\gamma_I M K[\mathbf{x}] + \gamma_A I)^{-1})^T$$

by the symmetry of $K[\mathbf{x}]$ and $M$, showing that $K[\mathbf{x}](\gamma_I M K[\mathbf{x}] + \gamma_A I)^{-1}$ is symmetric. The positive semidefiniteness of $Q[\mathbf{x}, C]$ simply follows from (158). This completes the proof of the theorem. ∎

**Proof of Theorem 7** Let $S_{\hat{y}_i}$ be the matrix obtained from $S$ by removing the $y_i$th column and $\beta_i \in \mathbb{R}^{P-1}$ be the vector obtained from $\alpha_i$ by deleting the $y_i$th entry, which is equal to zero by assumption. As in the proof of Theorem 6, for $1 \leq i \leq l$,

$$\gamma_I \sum_{j,k=1}^{u+l} M_{ik}K(x_k, x_j)a_j + \gamma_A a_i = -\frac{1}{2}C^* S\alpha_i = -\frac{1}{2}C^* S_{\hat{y}_i}\beta_i. \tag{165}$$

For $l + 1 \leq i \leq u + l$,

$$\gamma_I \sum_{j,k=1}^{u+l} M_{ik}K(x_k, x_j)a_j + \gamma_A a_i = 0. \tag{166}$$

Let $\mathrm{diag}(S_{\hat{\mathbf{y}}})$ be the $l \times l$ block diagonal matrix, with block $(i, i)$ being $S_{\hat{y}_i}$. Let $\beta = (\beta_1, \ldots, \beta_l)$ be the $(P-1) \times l$ matrix with column $i$ being $\beta_i$. In operator-valued matrix notation, (165) and (166) together can be expressed as

$$(\gamma_I M K[\mathbf{x}] + \gamma_A I)\mathbf{a} = -\frac{1}{2}(I_{(u+l) \times l} \otimes C^*)\mathrm{diag}(S_{\hat{\mathbf{y}}})\mathrm{vec}(\beta). \tag{167}$$

By Lemma 25, the operator $(\gamma_I M K[\mathbf{x}] + \gamma_A I) : \mathcal{W}^{u+l} \to \mathcal{W}^{u+l}$ is invertible, with a bounded inverse, so that

$$\mathbf{a} = -\frac{1}{2}(\gamma_I M K[\mathbf{x}] + \gamma_A I)^{-1}(I_{(u+l) \times l} \otimes C^*)\mathrm{diag}(S_{\hat{\mathbf{y}}})\mathrm{vec}(\beta). \tag{168}$$

As in the proof of Theorem 6,

$$\gamma_A ||f||^2_{\mathcal{H}_K} + \gamma_I \langle \mathbf{f}, M\mathbf{f} \rangle_{\mathcal{W}^{u+l}} = -\frac{1}{2} \sum_{i=1}^{l} \langle \sum_{j=1}^{u+l} K(x_i, x_j)a_j, C^* S \alpha_i \rangle_{\mathcal{W}}$$

$$= -\frac{1}{2} \sum_{i=1}^{l} \langle \sum_{j=1}^{u+l} K(x_i, x_j)a_j, C^* S_{\hat{y}_i} \beta_i \rangle_{\mathcal{W}} \tag{169}$$

$$= -\frac{1}{2} \sum_{i=1}^{l} \langle S_{\hat{y}_i}^* C \sum_{j=1}^{u+l} K(x_i, x_j)a_j, \beta_i \rangle_{\mathbb{R}^{P-1}}. \tag{170}$$

In operator-valued matrix notation, this is

$$\gamma_A ||f||^2_{\mathcal{H}_K} + \gamma_I \langle \mathbf{f}, M\mathbf{f} \rangle_{\mathcal{W}^{u+l}} = -\frac{1}{2}\mathrm{vec}(\beta)^T \mathrm{diag}(S_{\hat{\mathbf{y}}}^*)(I_{(u+l) \times l}^T \otimes C)K[\mathbf{x}]\mathbf{a}. \tag{171}$$

Substituting the expression for $\mathbf{a}$ in (168) into (171), we obtain

$$\gamma_A ||f||^2_{\mathcal{H}_K} + \gamma_I \langle \mathbf{f}, M\mathbf{f} \rangle_{\mathcal{W}^{u+l}} = \frac{1}{4}\mathrm{vec}(\beta)^T \mathrm{diag}(S_{\hat{\mathbf{y}}}^*)(I_{(u+l) \times l}^T \otimes C)K[\mathbf{x}]$$

$$\times (\gamma_I M K[\mathbf{x}] + \gamma_A I)^{-1}(I_{(u+l) \times l} \otimes C^*)\mathrm{diag}(S_{\hat{\mathbf{y}}})\mathrm{vec}(\beta). \tag{172}$$

We now note that

$$\sum_{i=1}^{l} \sum_{k \neq y_i} \alpha_{ki} \langle s_k, s_{y_i} \rangle_{\mathcal{Y}} = \sum_{i=1}^{l} \langle s_{y_i}, S_{\hat{y}_i} \beta_i \rangle_{\mathcal{Y}}. \tag{173}$$

Combining (151), (173), (153), and (172), we obtain the final form of the Lagrangian

$$L(\beta) = -\sum_{i=1}^{l} \langle s_{y_i}, S_{\hat{y}_i} \beta_i \rangle_{\mathcal{Y}} - \frac{1}{4}\mathrm{vec}(\beta)^T Q[\mathbf{x}, \mathbf{y}, C]\mathrm{vec}(\beta), \tag{174}$$

where the matrix $Q[\mathbf{x}, \mathbf{y}, C]$ is given by

$$Q[\mathbf{x}, \mathbf{y}, C] = \mathrm{diag}(S_{\hat{\mathbf{y}}}^*)(I_{(u+l) \times l}^T \otimes C)K[\mathbf{x}] \tag{175}$$

$$\times (\gamma_I M K[\mathbf{x}] + \gamma_A I)^{-1}(I_{(u+l) \times l} \otimes C^*)\mathrm{diag}(S_{\hat{\mathbf{y}}}). \tag{176}$$

We need to maximize the Lagrangian subject to the constraints

$$0 \leq \beta_{ki} \leq \frac{1}{l}, \quad 1 \leq i \leq l, 1 \leq k \leq P-1. \tag{177}$$

Equivalently, under the same constraints, we minimize

$$D(\beta) = \frac{1}{4}\text{vec}(\beta)^T Q[\mathbf{x}, \mathbf{y}, C]\text{vec}(\beta) + \sum_{i=1}^{l} \langle s_{y_i}, S_{\hat{y}_i}\beta_i \rangle_{\mathcal{Y}}. \tag{178}$$

If $S$ is the simplex coding, then

$$\langle s_{y_i}, S_{\hat{y}_i}\beta_i \rangle_{\mathcal{Y}} = \langle S_{\hat{y}_i}^T s_{y_i}, \beta_i \rangle_{\mathcal{Y}} = -\frac{1}{P-1}\mathbf{1}_{P-1}^T \beta_i.$$

It follows then that

$$\sum_{i=1}^{l} \langle s_{y_i}, S_{\hat{y}_i}\beta_i \rangle_{\mathcal{Y}} = -\frac{1}{P-1}\mathbf{1}_{(P-1)l}^T \text{vec}(\beta),$$

giving the last expression of the theorem. This completes the proof. ∎

**Lemma 29** *The matrix-valued kernel $K$ is positive definite.*

**Proof** Let $d = \dim(\mathcal{Y})$. Consider an arbitrary set of points $\mathbf{x} = \{x_i\}_{i=1}^N$ in $\mathcal{X}$ and an arbitrary set of vectors $\{y_i\}_{i=1}^N$ in $\mathbb{R}^{md}$. We need to show that

$$\sum_{i,j=1}^{N} \langle y_i, K(x_i, x_j)y_j \rangle_{\mathbb{R}^{md}} = \mathbf{y}^T K[\mathbf{x}]\mathbf{y} \geq 0,$$

where $\mathbf{y} = (y_1, \ldots, y_N) \in \mathbb{R}^{mdN}$ as a column vector. This is equivalent to showing that the Gram matrix $K[\mathbf{x}]$ of size $mdN \times mdN$ is positive semi-definite for any set $\mathbf{x}$.

By assumption, $G$ is positive definite, so that the Gram matrix $G[\mathbf{x}]$ of size $mN \times mN$ is positive semi-definite for any set $\mathbf{x}$. Since the Kronecker tensor product of two positive semi-definite matrices is positive semi-definite, the matrix

$$K[\mathbf{x}] = G[\mathbf{x}] \otimes R$$

is positive semi-definite for any set $\mathbf{x}$. This completes the proof. ∎

To prove Theorem 11, we need the following result.

**Lemma 30** *Let $N, n \in \mathbb{N}$ and $\gamma > 0$. Let $U$ be an orthogonal matrix of size $n \times n$, with columns $\mathbf{u}_1, \ldots, \mathbf{u}_n$. Let $A_i$ be $N \times N$ matrices such that $(A_i + \gamma I_N)$ is invertible for all $i$, $1 \leq i \leq n$. Then*

$$\left( \sum_{i=1}^{n} A_i \otimes \mathbf{u}_i\mathbf{u}_i^T + \gamma I_{Nn} \right)^{-1} = \sum_{i=1}^{n} (A_i + \gamma I_N)^{-1} \otimes \mathbf{u}_i\mathbf{u}_i^T. \tag{179}$$

**Proof** By definition of orthogonal matrices, we have $UU^T = I_n$, which is equivalent to $\sum_{i=1}^{n} \mathbf{u}_i \mathbf{u}_i^T = I_n$, so that

$$\sum_{i=1}^{n} A_i \otimes \mathbf{u}_i \mathbf{u}_i^T + \gamma I_{Nn} = \sum_{i=1}^{n} A_i \otimes \mathbf{u}_i \mathbf{u}_i^T + \gamma I_N \otimes \sum_{i=1}^{n} \mathbf{u}_i \mathbf{u}_i^T = \sum_{i=1}^{n} (A_i + \gamma I_N) \otimes \mathbf{u}_i \mathbf{u}_i^T.$$

Noting that $\langle \mathbf{u}_i, \mathbf{u}_j \rangle = \delta_{ij}$, the expression for the inverse matrix then follows immediately by direct verification. ∎

**Proof of Theorem 11** From the property $K[\mathbf{x}] = G[\mathbf{x}] \otimes R$ and the definitions $M_B = I_{u+l} \otimes M_m \otimes I_{\mathcal{Y}}$, $M_W = L \otimes I_{\mathcal{Y}}$, we have

$$\gamma_I M K[\mathbf{x}] + \gamma_A I_{\mathcal{W}^{u+l}} = (\gamma_B M_B + \gamma_W M_W) K[\mathbf{x}] + \gamma_A I_{u+l} \otimes I_{\mathcal{W}}$$

$$= (\gamma_B I_{u+l} \otimes M_m \otimes I_{\mathcal{Y}} + \gamma_W L \otimes I_{\mathcal{Y}})(G[\mathbf{x}] \otimes R) + \gamma_A I_{u+l} \otimes I_m \otimes I_{\mathcal{Y}}$$

$$= (\gamma_B I_{u+l} \otimes M_m + \gamma_W L) G[\mathbf{x}] \otimes R + \gamma_A I_{m(u+l)} \otimes I_{\mathcal{Y}}.$$

With the spectral decomposition of $R$,

$$R = \sum_{i=1}^{\dim(\mathcal{Y})} \lambda_{i,R} \mathbf{r}_i \mathbf{r}_i^T,$$

we have

$$(\gamma_B I_{u+l} \otimes M_m + \gamma_W L) G[\mathbf{x}] \otimes R = \sum_{i=1}^{\dim(\mathcal{Y})} \lambda_{i,R} (\gamma_B I_{u+l} \otimes M_m + \gamma_W L) G[\mathbf{x}] \otimes \mathbf{r}_i \mathbf{r}_i^T.$$

It follows from Lemma 30 that

$$(\gamma_I M K[\mathbf{x}] + \gamma_A I_{\mathcal{W}^{u+l}})^{-1} = \sum_{i=1}^{\dim(\mathcal{Y})} [\lambda_{i,R}(\gamma_B I_{u+l} \otimes M_m + \gamma_W L) G[\mathbf{x}] + \gamma_A I_{m(u+l)}]^{-1} \otimes \mathbf{r}_i \mathbf{r}_i^T$$

$$= \sum_{i=1}^{\dim(\mathcal{Y})} M_{\text{reg}}^i \otimes \mathbf{r}_i \mathbf{r}_i^T, \quad \text{where} \quad M_{\text{reg}}^i = [\lambda_{i,R}(\gamma_B I_{u+l} \otimes M_m + \gamma_W L) G[\mathbf{x}] + \gamma_A I_{m(u+l)}]^{-1}.$$

For $C = \mathbf{c}^T \otimes I_{\mathcal{Y}} \in \mathbb{R}^{\dim(\mathcal{Y}) \times m \dim(\mathcal{Y})}$, we have

$$C^* S = (\mathbf{c} \otimes I_{\mathcal{Y}}) S = \mathbf{c} \otimes S,$$

$$S^* C = S^*(\mathbf{c}^T \otimes I_{\mathcal{Y}}) = \mathbf{c}^T \otimes S^*,$$

$$I_{(u+l) \times l}^T \otimes S^* C = I_{(u+l) \times l}^T \otimes \mathbf{c}^T \otimes S^*,$$

$$I_{(u+l) \times l} \otimes C^* S = I_{(u+l) \times l} \otimes \mathbf{c} \otimes S.$$

It follows that

$$(\gamma_I M K[\mathbf{x}] + \gamma_A I_{\mathcal{W}^{u+l}})^{-1}(I_{(u+l) \times l} \otimes C^* S)$$

$$= (\sum_{i=1}^{\dim(\mathcal{Y})} M_{\mathrm{reg}}^i \otimes \mathbf{r}_i \mathbf{r}_i^T)(I_{(u+l)\times l} \otimes \mathbf{c} \otimes S)$$

$$= \sum_{i=1}^{\dim(\mathcal{Y})} M_{\mathrm{reg}}^i(I_{(u+l)\times l} \otimes \mathbf{c}) \otimes \mathbf{r}_i \mathbf{r}_i^T S,$$

from which we obtain the expression for $\mathbf{a}$. Next,

$$K[\mathbf{x}](\gamma_I M K[\mathbf{x}] + \gamma_A I)^{-1} = (G[\mathbf{x}] \otimes R)(\sum_{i=1}^{\dim(\mathcal{Y})} M_{\mathrm{reg}}^i \otimes \mathbf{r}_i \mathbf{r}_i^T)$$

$$= (\sum_{i=1}^{\dim(\mathcal{Y})} G[\mathbf{x}] \otimes \lambda_{i,R} \mathbf{r}_i \mathbf{r}_i^T)(\sum_{i=1}^{\dim(\mathcal{Y})} M_{\mathrm{reg}}^i \otimes \mathbf{r}_i \mathbf{r}_i^T) = \sum_{i=1}^{\dim(\mathcal{Y})} G[\mathbf{x}] M_{\mathrm{reg}}^i \otimes \lambda_{i,R} \mathbf{r}_i \mathbf{r}_i^T.$$

Thus for $Q[\mathbf{x}, C]$, we have

$$Q[\mathbf{x}, C] = (I_{(u+l)\times l}^T \otimes S^* C) K[\mathbf{x}](\gamma_I M K[\mathbf{x}] + \gamma_A I)^{-1}(I_{(u+l)\times l} \otimes C^* S)$$

$$= (I_{(u+l)\times l}^T \otimes \mathbf{c}^T \otimes S^*)(\sum_{i=1}^{\dim(\mathcal{Y})} G[\mathbf{x}] M_{\mathrm{reg}}^i \otimes \lambda_{i,R} \mathbf{r}_i \mathbf{r}_i^T)(I_{(u+l)\times l} \otimes \mathbf{c} \otimes S)$$

$$= (\sum_{i=1}^{\dim(\mathcal{Y})} (I_{(u+l)\times l}^T \otimes \mathbf{c}^T) G[\mathbf{x}] M_{\mathrm{reg}}^i \otimes \lambda_{i,R} S^* \mathbf{r}_i \mathbf{r}_i^T)(I_{(u+l)\times l} \otimes \mathbf{c} \otimes S)$$

$$= \sum_{i=1}^{\dim(\mathcal{Y})} (I_{(u+l)\times l}^T \otimes \mathbf{c}^T) G[\mathbf{x}] M_{\mathrm{reg}}^i(I_{(u+l)\times l} \otimes \mathbf{c}) \otimes \lambda_{i,R} S^* \mathbf{r}_i \mathbf{r}_i^T S.$$

This completes the proof of the theorem. ∎

**Proof of Theorem 13** For $R = I_{\mathcal{Y}}$ we have $\lambda_{i,R} = 1$, $1 \leq i \leq \dim(\mathcal{Y})$, so that in Theorem 11,

$$M_{\mathrm{reg}}^i = M_{\mathrm{reg}} = [(\gamma_B I_{u+l} \otimes M_m + \gamma_W L) G[\mathbf{x}] + \gamma_A I_{m(u+l)}]^{-1}.$$

Since $\sum_{i=1}^{\dim(\mathcal{Y})} \mathbf{r}_i \mathbf{r}_i^T = I_{\mathcal{Y}}$, by substituting $M_{\mathrm{reg}}^i = M_{\mathrm{reg}}$ into the formulas for $\mathbf{a}$ and $Q[\mathbf{x}, C]$ in Theorem 11, we obtain the corresponding expressions (77) and (78). ∎

**Proof of Propositions 12 and 14** By Theorems 6 and 11, we have

$$f_{\mathbf{z},\gamma}(v) = \sum_{j=1}^{u+l} K(v, x_j) a_j = K[v, \mathbf{x}]\mathbf{a} = (G[v, \mathbf{x}] \otimes R)\mathbf{a}$$

$$= -\frac{1}{2}(G[v, \mathbf{x}] \otimes \sum_{i=1}^{\dim(\mathcal{Y})} \lambda_{i,R} \mathbf{r}_i \mathbf{r}_i^T)[\sum_{i=1}^{\dim(\mathcal{Y})} M_{\mathrm{reg}}^i(I_{(u+l)\times l} \otimes \mathbf{c}) \otimes \mathbf{r}_i \mathbf{r}_i^T S]\mathrm{vec}(\alpha^{\mathrm{opt}})$$

$$= -\frac{1}{2}[\sum_{i=1}^{\dim(\mathcal{Y})} G[v, \mathbf{x}] M_{\mathrm{reg}}^i(I_{(u+l)\times l} \otimes \mathbf{c}) \otimes \lambda_{i,R} \mathbf{r}_i \mathbf{r}_i^T S]\mathrm{vec}(\alpha^{\mathrm{opt}})$$

$$= -\frac{1}{2}\mathrm{vec}(\sum_{i=1}^{\dim(\mathcal{Y})} \lambda_{i,R} \mathbf{r}_i \mathbf{r}_i^T S \alpha^{\mathrm{opt}}(I_{(u+l)\times l}^T \otimes \mathbf{c}^T)(M_{\mathrm{reg}}^i)^T G[v, \mathbf{x}]^T).$$

The combined function, using the combination operator $C$, is

$$g_{\mathbf{z},\gamma}(v) = Cf_{\mathbf{z},\gamma}(v) = (\mathbf{c}^T \otimes I_{\mathcal{Y}})(G[v,\mathbf{x}] \otimes R)\mathbf{a} = (\mathbf{c}^T G[v,\mathbf{x}] \otimes R)\mathbf{a}$$

$$= -\frac{1}{2}\big[\sum_{i=1}^{\dim(\mathcal{Y})} \mathbf{c}^T G[v,\mathbf{x}] M_{\text{reg}}^i (I_{(u+l)\times l} \otimes \mathbf{c}) \otimes \lambda_{i,R}\mathbf{r}_i\mathbf{r}_i^T S]\text{vec}(\alpha^{\text{opt}})$$

$$= -\frac{1}{2}\text{vec}\big(\sum_{i=1}^{\dim(\mathcal{Y})} \lambda_{i,R}\mathbf{r}_i\mathbf{r}_i^T S\alpha^{\text{opt}}(I_{(u+l)\times l}^T \otimes \mathbf{c}^T)(M_{\text{reg}}^i)^T G[v,\mathbf{x}]^T\mathbf{c}\big)$$

$$= -\frac{1}{2}\sum_{i=1}^{\dim(\mathcal{Y})} \lambda_{i,R}\mathbf{r}_i\mathbf{r}_i^T S\alpha^{\text{opt}}(I_{(u+l)\times l}^T \otimes \mathbf{c}^T)(M_{\text{reg}}^i)^T G[v,\mathbf{x}]^T\mathbf{c} \in \mathbb{R}^{\dim(\mathcal{Y})}.$$

It follows that on a set $\mathbf{v} = \{v_i\}_{i=1}^t \subset \mathcal{X}$,

$$g_{\mathbf{z},\gamma}(\mathbf{v}) = -\frac{1}{2}\sum_{i=1}^{\dim(\mathcal{Y})} \lambda_{i,R}\mathbf{r}_i\mathbf{r}_i^T S\alpha^{\text{opt}}(I_{(u+l)\times l}^T \otimes \mathbf{c}^T)(M_{\text{reg}}^i)^T G[\mathbf{v},\mathbf{x}]^T(I_t \otimes \mathbf{c}) \in \mathbb{R}^{\dim(\mathcal{Y})\times t}.$$

The final SVM decision function is then given by

$$h_{\mathbf{z},\gamma}(\mathbf{v}) = S^T g_{\mathbf{z},\gamma}(\mathbf{v}) = -\frac{1}{2}\sum_{i=1}^{\dim(\mathcal{Y})} \lambda_{i,R}S^T\mathbf{r}_i\mathbf{r}_i^T S\alpha^{\text{opt}}(I_{(u+l)\times l}^T \otimes \mathbf{c}^T)(M_{\text{reg}}^i)^T G[\mathbf{v},\mathbf{x}]^T(I_t \otimes \mathbf{c}) \in \mathbb{R}^{P\times t}.$$

This completes the proof for Proposition 12. Proposition 14 then follows by noting that in Theorem 13, with $R = I_{\mathcal{Y}}$, we have $M_{\text{reg}}^i = M_{\text{reg}}$, $\lambda_{i,R} = 1$, $1 \leq i \leq \dim(\mathcal{Y})$, and $\sum_{i=1}^{\dim(\mathcal{Y})} \mathbf{r}_i\mathbf{r}_i^T = I_{\mathcal{Y}}$. $\blacksquare$

**Proof of Corollary 20** Clearly, for $\gamma_I = 0$ and $u = 0$, we have

$$Q[\mathbf{x},\mathbf{y},C] = \frac{1}{\gamma_A}\text{diag}(S_{\hat{\mathbf{y}}}^*)(I_l \otimes C)K[\mathbf{x}](I_l \otimes C^*)\text{diag}(S_{\hat{\mathbf{y}}}).$$

For $C = \mathbf{c}^T \otimes I_{\mathcal{Y}}$ and $K[\mathbf{x}] = G[\mathbf{x}] \otimes R$, this is

$$Q[\mathbf{x},\mathbf{y},C] = \frac{1}{\gamma_A}\text{diag}(S_{\hat{\mathbf{y}}}^*)[(I_l \otimes \mathbf{c}^T)G[\mathbf{x}](I_l \otimes \mathbf{c}) \otimes R]\text{diag}(S_{\hat{\mathbf{y}}}).$$

For $R = I_{\mathcal{Y}}$, we have

$$Q[\mathbf{x},\mathbf{y},C] = \frac{1}{\gamma_A}\text{diag}(S_{\hat{\mathbf{y}}}^*)[(I_l \otimes \mathbf{c}^T)G[\mathbf{x}](I_l \otimes \mathbf{c}) \otimes I_{\mathcal{Y}}]\text{diag}(S_{\hat{\mathbf{y}}}).$$

With $G[\mathbf{x}] = \sum_{i=1}^m k^i[\mathbf{x}] \otimes \mathbf{e}_i\mathbf{e}_i^T$,

$$(I_l \otimes \mathbf{c}^T)G[\mathbf{x}](I_l \otimes \mathbf{c}) = \sum_{i=1}^m c_i^2 k^i[\mathbf{x}],$$

so that

$$Q[\mathbf{x}, \mathbf{y}, C] = \frac{1}{\gamma_A} \text{diag}(S_{\hat{\mathbf{y}}}^*) \left( \sum_{i=1}^m c_i^2 k^i[\mathbf{x}] \otimes I_{\mathcal{Y}} \right) \text{diag}(S_{\hat{\mathbf{y}}}),$$

which, when $\mathcal{Y} = \mathbb{R}$, reduces to

$$Q[\mathbf{x}, \mathbf{y}, C] = \frac{1}{\gamma_A} \text{diag}(\mathbf{y}) \left( \sum_{i=1}^m c_i^2 k^i[\mathbf{x}] \right) \text{diag}(\mathbf{y}).$$

Similarly, when $\gamma_I = 0$, $u = 0$, we have

$$\mathbf{a} = -\frac{1}{2\gamma_A} (I_l \otimes C^*) \text{diag}(S_{\hat{\mathbf{y}}}) \text{vec}(\beta^{\text{opt}}).$$

For $C = \mathbf{c} \otimes I_{\mathcal{Y}}$, $K(x, t) = G(x, t) \otimes R$, we have for any $v \in \mathcal{X}$,

$$Cf_{\mathbf{z},\gamma}(v) = CK[v, \mathbf{x}]\mathbf{a} = -\frac{1}{2\gamma_A} (\mathbf{c}^T \otimes I_{\mathcal{Y}})(G[v, \mathbf{x}] \otimes R)(I_l \otimes \mathbf{c} \otimes I_{\mathcal{Y}})\text{diag}(S_{\hat{\mathbf{y}}})\text{vec}(\beta^{\text{opt}})$$

$$= -\frac{1}{2\gamma_A}[\mathbf{c}^T G[v, \mathbf{x}](I_l \otimes \mathbf{c}) \otimes R]\text{diag}(S_{\hat{\mathbf{y}}})\text{vec}(\beta^{\text{opt}}),$$

which for $R = I_{\mathcal{Y}}$, simplifies to

$$Cf_{\mathbf{z},\gamma}(v) = -\frac{1}{2\gamma_A}[\mathbf{c}^T G[v, \mathbf{x}](I_l \otimes \mathbf{c}) \otimes I_{\mathcal{Y}}]\text{diag}(S_{\hat{\mathbf{y}}})\text{vec}(\beta^{\text{opt}}),$$

With $G(x, t) = \sum_{i=1}^m k^i(x, t) \otimes \mathbf{e}_i \mathbf{e}_i^T$,

$$\mathbf{c}^T G[v, \mathbf{x}](I_l \otimes \mathbf{c}) = \sum_{i=1}^m c_i^2 k^i[v, \mathbf{x}],$$

so that

$$Cf_{\mathbf{z},\gamma}(v) = -\frac{1}{2\gamma_A} \left( \sum_{i=1}^m c_i^2 k^i[v, \mathbf{x}] \otimes I_{\mathcal{Y}} \right) \text{diag}(S_{\hat{\mathbf{y}}})\text{vec}(\beta^{\text{opt}}).$$

For $\mathcal{Y} = \mathbb{R}$, this simplifies to

$$Cf_{\mathbf{z},\gamma}(v) = -\frac{1}{2\gamma_A} \left( \sum_{i=1}^m c_i^2 k^i[v, \mathbf{x}] \right) \text{diag}(\mathbf{y})\beta^{\text{opt}}.$$

This completes the proof. ∎

### A.4 Sequential Minimal Optimization

This section describes the Sequential Minimal Optimization (SMO) algorithm we use to solve the quadratic optimization problem for MV-SVM in Theorem 6. It is a generalization of the one-step SMO technique described in (Platt, 1999). For simplicity and clarity, we consider the case of the simplex coding, that is the quadratic optimization problem (26). The ideas presented here are readily extendible to the general setting.

Let us first consider the SMO technique for the quadratic optimization problem

$$\text{argmin}_{\alpha \in \mathbb{R}^{Pl}} D(\alpha) = \frac{1}{4} \alpha^T Q \alpha - \frac{1}{P-1} \mathbf{1}_{Pl}^T \alpha, \tag{180}$$

where $Q$ is a symmetric, positive semidefinite matrix of size $Pl \times Pl$, such that $Q_{ii} > 0$, $1 \leq i \leq Pl$, under the constraints

$$0 \leq \alpha_i \leq \frac{1}{l}, \quad 1 \leq i \leq Pl. \tag{181}$$

For $i$ fixed, $1 \leq i \leq Pl$, as a function of $\alpha_i$,

$$D(\alpha) = \frac{1}{4} Q_{ii} \alpha_i^2 + \frac{1}{2} \sum_{j=1, j \neq i}^{Pl} Q_{ij} \alpha_i \alpha_j - \frac{1}{P-1} \alpha_i + Q_{\text{const}}, \tag{182}$$

where $Q_{\text{const}}$ is a quantity constant in $\alpha_i$. Differentiating with respect to $\alpha_i$ gives

$$\frac{\partial D}{\partial \alpha_i} = \frac{1}{2} Q_{ii} \alpha_i + \frac{1}{2} \sum_{j=1, j \neq i}^{Pl} Q_{ij} \alpha_j - \frac{1}{P-1}. \tag{183}$$

Under the condition that $Q_{ii} > 0$, setting this partial derivative to zero gives

$$\alpha_i^* = \frac{1}{Q_{ii}} \left( \frac{2}{P-1} - \sum_{j=1, j \neq i}^{Pl} Q_{ij} \alpha_j \right) = \alpha_i + \frac{1}{Q_{ii}} \left( \frac{2}{P-1} - \sum_{j=1}^{Pl} Q_{ij} \alpha_j \right). \tag{184}$$

Thus the iterative sequence for $\alpha_i$ at step $t$ is

$$\alpha_i^{t+1} = \alpha_i^t + \frac{1}{Q_{ii}} \left( \frac{2}{P-1} - \sum_{j=1}^{Pl} Q_{ij} \alpha_j^t \right), \tag{185}$$

after which we perform a clipping operation, defined by

$$\text{clip}(\alpha_i) = \left\{ \begin{array}{ll} 0 & \text{if } \alpha_i < 0, \\ \alpha_i & \text{if } 0 \leq \alpha_i \leq \frac{1}{l}, \\ \frac{1}{l} & \text{if } \alpha_i > \frac{1}{l}. \end{array} \right. \tag{186}$$

Let us now apply this SMO technique for the quadratic optimization (26) in Theorem 6. Recall that this problem is

$$\alpha^{\text{opt}} = \text{argmin}_{\alpha \in \mathbb{R}^{P \times l}} \left\{ D(\alpha) = \frac{1}{4} \text{vec}(\alpha)^T Q[\mathbf{x}, C] \text{vec}(\alpha) - \frac{1}{P-1} \mathbf{1}_{Pl}^T \text{vec}(\alpha) \right\},$$

with $\mathbf{1}_{Pl} = (1, \ldots, 1)^T \in \mathbb{R}^{Pl}$, subject to the constraints

$$0 \leq \alpha_{ki} \leq \frac{1}{l}(1 - \delta_{k,y_i}), \quad 1 \leq i \leq l, 1 \leq k \leq P.$$

The choice of which $\alpha_{ki}$ to update at each step is made via the Karush-Kuhn-Tucker (KKT) conditions. In the present context, the KKT conditions are:

$$\alpha_{ki}\left(\xi_{ki} - \left[\frac{1}{P-1} + \langle s_k, Cf(x_i)\rangle_{\mathcal{Y}}\right]\right) = 0, \quad 1 \leq i \leq l, k \neq y_i, \tag{187}$$

$$\left(\frac{1}{l} - \alpha_{ki}\right)\xi_{ki} = 0, \quad 1 \leq i \leq l, k \neq y_i. \tag{188}$$

At an optimal point $\alpha^{\mathrm{opt}}$,

$$\xi_{ki}^{\mathrm{opt}} = \max\left(0, \frac{1}{P-1} + \langle s_k, Cf(x_i)\rangle_{\mathcal{Y}}\right), \quad 1 \leq i \leq l, k \neq y_i. \tag{189}$$

We have the following result.

**Lemma 31** *For $1 \leq i \leq l$, $k \neq y_i$,*

$$\alpha_{ki}^{\mathrm{opt}} = 0 \implies \langle s_k, Cf_{\mathbf{z},\gamma}(x_i)\rangle_{\mathcal{Y}} \leq -\frac{1}{P-1}, \tag{190}$$

$$0 < \alpha_{ki}^{\mathrm{opt}} < \frac{1}{l} \implies \langle s_k, Cf_{\mathbf{z},\gamma}(x_i)\rangle_{\mathcal{Y}} = -\frac{1}{P-1}, \tag{191}$$

$$\alpha_{ki}^{\mathrm{opt}} = \frac{1}{l} \implies \langle s_k, Cf_{\mathbf{z},\gamma}(x_i)\rangle_{\mathcal{Y}} \geq -\frac{1}{P-1}. \tag{192}$$

*Conversely,*

$$\langle s_k, Cf_{\mathbf{z},\gamma}(x_i)\rangle_{\mathcal{Y}} < -\frac{1}{P-1} \implies \alpha_{ki}^{\mathrm{opt}} = 0, \tag{193}$$

$$\langle s_k, Cf_{\mathbf{z},\gamma}(x_i)\rangle_{\mathcal{Y}} > -\frac{1}{P-1} \implies \alpha_{ki}^{\mathrm{opt}} = \frac{1}{l}. \tag{194}$$

**Remark 32** *Note that the inequalities in (190) and (192) are not strict. Thus from $\langle s_k, Cf_{\mathbf{z},\gamma}(x_i)\rangle_{\mathcal{Y}} = -\frac{1}{P-1}$ we cannot draw any conclusion about $\alpha_{ki}^{\mathrm{opt}}$.*

**Proof** To prove (190), note that if $\alpha_{ki}^{\mathrm{opt}} = 0$, then from (188), we have $\xi_{ki}^{\mathrm{opt}} = 0$. From (189), we have

$$\frac{1}{P-1} + \langle s_k, Cf(x_i)\rangle_{\mathcal{Y}} \leq 0 \implies \langle s_k, Cf(x_i)\rangle_{\mathcal{Y}} \leq -\frac{1}{P-1}.$$

To prove (191), note that if $0 < \alpha_{ki}^{\mathrm{opt}} < \frac{1}{l}$, then from (188), we have $\xi_{ki}^{\mathrm{opt}} = 0$. On the other hand, from (187), we have

$$\xi_{ki}^{\mathrm{opt}} = \frac{1}{P-1} + \langle s_k, Cf(x_i)\rangle_{\mathcal{Y}}.$$

It follows that

$$\frac{1}{P-1} + \langle s_k, Cf(x_i)\rangle_{\mathcal{Y}} = 0 \iff \langle s_k, Cf(x_i)\rangle_{\mathcal{Y}} = -\frac{1}{P-1}.$$

For (192), note that if $\alpha_{ki}^{\text{opt}} = \frac{1}{l}$, then from (187), we have

$$\xi_{ki}^{\text{opt}} = \frac{1}{P-1} + \langle s_k, Cf(x_i)\rangle_{\mathcal{Y}} \geq 0 \implies \langle s_k, Cf(x_i)\rangle_{\mathcal{Y}} \geq -\frac{1}{P-1}.$$

Conversely, if $\langle s_k, Cf(x_i)\rangle_{\mathcal{Y}} < -\frac{1}{P-1}$, then from (189), we have $\xi_{ki}^{\text{opt}} = 0$. It then follows from (187) that $\alpha_{ki}^{\text{opt}} = 0$. If $\langle s_k, Cf(x_i)\rangle_{\mathcal{Y}} > -\frac{1}{P-1}$, then from (189) we have $\xi_{ki}^{\text{opt}} = \frac{1}{P-1} + \langle s_k, Cf(x_i)\rangle_{\mathcal{Y}}$. Then from (188) it follows that $\alpha_{ki}^{\text{opt}} = \frac{1}{l}$. ∎

**Binary case** ($P = 2$): The binary simplex code is $S = [-1, 1]$. Thus $k \neq y_i$ means that $s_k = -y_i$. Therefore for $1 \leq i \leq l$, $k \neq y_i$, the KKT conditions are:

$$\alpha_{ki}^{\text{opt}} = 0 \implies y_i\langle \mathbf{c}, f_{\mathbf{z},\gamma}(x_i)\rangle_{\mathcal{W}} \geq 1, \tag{195}$$

$$0 < \alpha_{ki}^{\text{opt}} < \frac{1}{l} \implies y_i\langle \mathbf{c}, f_{\mathbf{z},\gamma}(x_i)\rangle_{\mathcal{W}} = 1, \tag{196}$$

$$\alpha_{ki}^{\text{opt}} = \frac{1}{l} \implies y_i\langle \mathbf{c}, f_{\mathbf{z},\gamma}(x_i)\rangle_{\mathcal{W}} \leq 1. \tag{197}$$

Conversely,

$$y_i\langle \mathbf{c}, f_{\mathbf{z},\gamma}(x_i)\rangle_{\mathcal{W}} > 1 \implies \alpha_{ki}^{\text{opt}} = 0, \tag{198}$$

$$y_i\langle \mathbf{c}, f_{\mathbf{z},\gamma}(x_i)\rangle_{\mathcal{W}} < 1 \implies \alpha_{ki}^{\text{opt}} = \frac{1}{l}. \tag{199}$$

Algorithm 3 summarizes the SMO procedure described in this section.

### A.4.1 Numerical implementation of SMO

Let us elaborate on the steps of Algorithm 3 under the hypotheses of Theorem 13, that is the simplex coding with $K[\mathbf{x}] = G[\mathbf{x}] \otimes R$ for $R = I_{P-1}$, which we implement numerically.

**Verifying the Karush-Kuhn-Tucker conditions on the labeled training data**: To verify Lemma 31 on the set of labeled training data $\mathbf{x}_{1:l} = \{x_i\}_{i=1}^l \subset \mathbf{x}$, according to Proposition 14, we compute

$$h_{\mathbf{z},\gamma}(\mathbf{x}_{1:l}) = -\frac{1}{2}S^T S\alpha^{\text{opt}}(I_{(u+l)\times l}^T \otimes \mathbf{c}^T)M_{\text{reg}}^T G[\mathbf{x}_{1:l}, \mathbf{x}]^T(I_l \otimes \mathbf{c}) \in \mathbb{R}^{P\times l}, \tag{200}$$

as a matrix of size $P \times l$, with the $i$th column being $h_{\mathbf{z},\gamma}(x_i) = (\langle s_k, Cf_{\mathbf{z},\gamma}(x_i)\rangle_{\mathcal{Y}})_{k=1}^P$, which is then compared with the margin value $-\frac{1}{P-1}$.

**Efficient evaluation of the update step (201)**: The most important factor underlying the efficiency of Algorithm 3 is that we never compute the whole matrix $Q$ of size $Pl \times Pl$, which can be prohibitively large. At each update step, i.e. (201), we only use the

---

**Algorithm 3** Sequential Minimal Optimization for Multi-class Multi-view SVM

---

Note: We use $\alpha \in \mathbb{R}^{P \times l}$ as a matrix and $\alpha_{\text{vec}} = \text{vec}(\alpha) \in \mathbb{R}^{Pl}$ as a column vector interchangeably.

**Initialization:** Set $\alpha^0 = 0$.

**Stopping criterion**: $\frac{|D(\alpha^{t+1}) - D(\alpha^t)|}{|D(\alpha^{t+1})|} < \epsilon$, for some $\epsilon > 0$.

**Repeat**: - Verify KKT conditions according to Lemma 31.

- Randomly pick an $i \in \mathbb{N}$ such that $\alpha^t_{\text{vec},i}$ is a KKT violator.

- Perform update:

$$\alpha^{t+1}_{\text{vec},i} = \text{clip}\left(\alpha^t_{\text{vec},i} + \frac{1}{Q_{ii}}\left(\frac{2}{P-1} - \sum_{j=1}^{Pl} Q_{ij}\alpha^t_{\text{vec},j}\right)\right), \tag{201}$$

where $Q = Q[\mathbf{x}, C]$.

**Until**: There are no KKT violators or the stopping criterion is met.

---

$i$th row of $Q$, which we denote $Q(i,:)$, which need not be computed explicitly. Recall that we have

$$Q = Q[\mathbf{x}, C] = (I^T_{(u+l) \times l} \otimes \mathbf{c}^T)G[\mathbf{x}]M_{\text{reg}}(I_{(u+l) \times l} \otimes \mathbf{c}) \otimes S^*S = Q_G \otimes Q_S,$$

where

$$Q_G = (I^T_{(u+l) \times l} \otimes \mathbf{c}^T)G[\mathbf{x}]M_{\text{reg}}(I_{(u+l) \times l} \otimes \mathbf{c}), \tag{202}$$

and

$$Q_S = S^*S. \tag{203}$$

Thus for each $i$, the $i$th row of $Q$ is

$$Q(i,:) = Q_G(i_G,:) \otimes Q_S(i_S,:), \tag{204}$$

for a unique pair of indices $i_G$ and $i_S$. It then follows that

$$Q(i,:)\text{vec}(\alpha) = (Q_G(i_G,:) \otimes Q_S(i_S,:))\text{vec}(\alpha) = \text{vec}(Q_S(i_S,:)\alpha Q_G(i_G,:)^T)$$
$$= Q_S(i_S,:)\alpha Q_G(i_G,:)^T = Q_S(i_S,:)\alpha Q_G(:,i_G) \tag{205}$$

since $Q_G$ is symmetric. Also

$$Q_{ii} = Q_G(i_G, i_G)Q_S(i_S, i_S). \tag{206}$$

When proceeding in this way, each update step (201) only uses *one* row from the $l \times l$ matrix $Q_G$ and *one* row from the $P \times P$ matrix $Q_S$. This is the key to the computational efficiency of Algorithm 3.

**Remark 33** *In the more general setting of Theorem 11, with $K[\mathbf{x}] = G[\mathbf{x}] \otimes R$, where $R$ is a positive semi-definite matrix, the evaluation of the matrix $Q = Q[\mathbf{x}, C]$ is done in the same way, except that we need to sum over all non-zero eigenvalues of $R$ (Equation 71).*

### A.5 Proofs for the Optimization of the Combination Operator

In this section, we prove Theorems 15 and 16 stated in Section 6.1. Consider the optimization problem (87), namely

$$\min_{\mathbf{x} \in \mathbb{R}^m} ||A\mathbf{x} - \mathbf{b}||_{\mathbb{R}^n} \text{ subject to } ||\mathbf{x}||_{\mathbb{R}^m} = \alpha.$$

The Lagrangian, with Lagrange multiplier $\gamma$, is given by

$$L(\mathbf{x}, \gamma) = ||A\mathbf{x} - \mathbf{b}||^2 + \gamma(||\mathbf{x}||^2 - \alpha^2).$$

Setting $\frac{\partial L}{\partial \mathbf{x}} = 0$ and $\frac{\partial L}{\partial \gamma} = 0$, we obtain the normal equations

$$(A^T A + \gamma I_m)\mathbf{x} = A^T \mathbf{b}, \tag{207}$$

$$||\mathbf{x}||^2 = \alpha^2. \tag{208}$$

The solutions of the normal equations (207) and (208), if they exist, satisfy the following properties (Gander, 1981).

**Lemma 34** *If $(\mathbf{x}_1, \gamma_1)$ and $(\mathbf{x}_2, \gamma_2)$ are solutions of the normal equations (207) and (208), then*

$$||A\mathbf{x}_2 - \mathbf{b}||^2 - ||A\mathbf{x}_1 - \mathbf{b}||^2 = \frac{\gamma_1 - \gamma_2}{2}||\mathbf{x}_1 - \mathbf{x}_2||^2. \tag{209}$$

**Lemma 35** *The righ hand side of Equation (209) is equal to zero only if $\gamma_1 = \gamma_2 = -\mu$, where $\mu \geq 0$ is an eigenvalue of $A^T A$ and*

$$\mathbf{x}_1 = \mathbf{x}_2 + \mathbf{v}(\mu), \tag{210}$$

*where $\mathbf{v}(\mu)$ is an eigenvector corresponding to $\mu$.*

According to Lemmas 34 and 35, if $(\mathbf{x}_1, \gamma_1)$ and $(\mathbf{x}_2, \gamma_2)$ are solutions of the normal equations (207) and (208), then

$$\gamma_1 > \gamma_2 \implies ||A\mathbf{x}_2 - \mathbf{b}|| > ||A\mathbf{x}_1 - \mathbf{b}||. \tag{211}$$

Consequently, among all possible solutions of the normal equations (207) and (208), we choose the solution $(\mathbf{x}, \gamma)$ with the largest $\gamma$.

**Proof of Theorem 15** Using the assumption $A^T \mathbf{b} = 0$, the first normal equation (207) implies that

$$A^T A\mathbf{x} = -\gamma\mathbf{x}, \tag{212}$$

so that $-\gamma$ is an eigenvalue of $A^T A$ and $\mathbf{x}$ is its corresponding eigenvector, which can be appropriately normalized such that $||\mathbf{x}||_{\mathbb{R}^m} = \alpha$. Since we need the largest value for $\gamma$, we have $\gamma^* = -\mu_m$. The minimum value is then

$$||A\mathbf{x}^* - \mathbf{b}||^2_{\mathbb{R}^n} = \langle A\mathbf{x}^*, A\mathbf{x}^* \rangle_{\mathbb{R}^n} - 2\langle A\mathbf{x}^*, \mathbf{b} \rangle_{\mathbb{R}^n} + ||\mathbf{b}||^2_{\mathbb{R}^n}$$
$$= \langle \mathbf{x}^*, A^T A\mathbf{x}^* \rangle_{\mathbb{R}^m} - 2\langle \mathbf{x}^*, A^T \mathbf{b} \rangle_{\mathbb{R}^m} + ||\mathbf{b}||^2_{\mathbb{R}^n} = -\gamma^*||\mathbf{x}^*||^2_{\mathbb{R}^m} + ||\mathbf{b}||^2_{\mathbb{R}^n}$$
$$= \mu_m \alpha^2 + ||\mathbf{b}||^2_{\mathbb{R}^n}.$$

This solution is clearly unique if and only if $\mu_m$ is a single eigenvalue. Otherwise, there are infinitely many solutions, each being a vector of length $\alpha$ in the eigenspace of $\mu_m$. This completes the proof of the theorem. ∎

**Proof of Theorem 16** We first show that under the assumption $A^T \mathbf{b} \neq 0$ and $\mathbf{c} = U^T \mathbf{b}$, we have $||\mathbf{c}_{1:r}||_{\mathbb{R}^r} \neq 0$, that is $c_i \neq 0$ for at least one index $i$, $1 \leq i \leq r$. To see this, assume that $c_i = 0$ for all $i$, $1 \leq i \leq r$. Then

$$A^T \mathbf{b} = V\Sigma^T U^T \mathbf{b} = V\Sigma^T \mathbf{c} = 0,$$

which is a contradiction. Thus $||\mathbf{c}_{1:r}||_{\mathbb{R}^r} \neq 0$.

There are two cases in this scenario.

(I) If $\gamma \neq -\mu_i$, $1 \leq i \leq m$, then the matrix $(A^T A + \gamma I_m)$ is nonsingular, thus

$$\mathbf{x}(\gamma) = (A^T A + \gamma I_m)^{-1} A^T \mathbf{b} = (VDV^T + \gamma I_m)^{-1} V\Sigma^T U^T \mathbf{b}$$
$$= V(DV^T V + \gamma I_m)^{-1} \Sigma^T U^T \mathbf{b} = V(D + \gamma I_m)^{-1} \Sigma^T U^T \mathbf{b}.$$

Since the matrix $V$ is orthogonal, we have

$$||\mathbf{x}(\gamma)||_{\mathbb{R}^m}^2 = ||(D + \gamma I_m)^{-1} \Sigma^T U^T \mathbf{b}||_{\mathbb{R}^m}^2 = \sum_{i=1}^{r} \frac{\sigma_i^2 c_i^2}{(\sigma_i^2 + \gamma)^2},$$

where $\mathbf{c} = U^T \mathbf{b}$. We now need to find $\gamma$ such that $||\mathbf{x}(\gamma)||_{\mathbb{R}^m} = \alpha$. Consider the function

$$s(\gamma) = \sum_{i=1}^{r} \frac{\sigma_i^2 c_i^2}{(\sigma_i^2 + \gamma)^2} \tag{213}$$

on the interval $(-\sigma_r^2, \infty)$. Under the condition that at least one of the $c_i$'s, $1 \leq i \leq r$, is nonzero, the function $s$ is strictly positive and monotonically decreasing on $(-\sigma_r^2, \infty)$, with

$$\lim_{\gamma \to \infty} s(\gamma) = 0, \quad \lim_{\gamma \to -\sigma_r^2} s(\gamma) = \infty. \tag{214}$$

Thus there must exist a unique $\gamma^* \in (-\sigma_r^2, \infty)$ such that

$$s(\gamma^*) = \sum_{i=1}^{r} \frac{\sigma_i^2 c_i^2}{(\sigma_i^2 + \gamma^*)^2} = \alpha^2. \tag{215}$$

1) If $\text{rank}(A) = m$, then $r = m$ and $\gamma^* > -\sigma_m^2 = -\mu_m \geq -\mu_i$ for all $1 \leq i \leq m$. Thus $\mathbf{x}(\gamma^*)$ is the unique global solution.

2) If $\text{rank}(A) < m$ but $\gamma^* > 0$, then we still have $\gamma^* > -\mu_i$ for all $i$, $1 \leq i \leq m$, and thus $\mathbf{x}(\gamma^*)$ is the unique global solution.

(II) Consider now the case $\text{rank}(A) < m$ and $\gamma^* \leq 0$.

Since $\mu_m = \ldots = \mu_{r+1} = 0$ and $-\mu_r = -\sigma_r^2 < \gamma^* \leq 0$, we need to consider the possible solution of the normal equations with $\gamma = 0$. For $\gamma = 0$, we have

$$A^T A\mathbf{x} = A^T \mathbf{b} \iff VDV^T \mathbf{x} = V\Sigma^T U^T \mathbf{b} \iff DV^T \mathbf{x} = \Sigma^T U^T \mathbf{b}. \tag{216}$$

Let $\mathbf{y} = V^T\mathbf{x} \in \mathbb{R}^m$. By assumption, the vector $D\mathbf{y} \in \mathbb{R}^m$ satisfies $(D\mathbf{y})_i = 0$, $r+1 \leq i \leq m$. The vector $\mathbf{z} = \Sigma^T U^T \mathbf{b} \in \mathbb{R}^m$ also satisfies $z_i = 0$, $r + 1 \leq i \leq m$. Thus the equation

$$D\mathbf{y} = \mathbf{z} \tag{217}$$

has infinitely many solutions, with $y_i$, $r + 1 \leq i \leq m$, taking arbitrary values. Let $\mathbf{y}_{1:r} = (y_i)_{i=1}^r$, $\mathbf{z}_{1:r} = (z_i)_{i=1}^r$, $D_r = \text{diag}(\mu_1, \ldots, \mu_r)$, $\Sigma_r = \Sigma(:, 1 : r)$ consisting of the first $r$ columns of $\Sigma$. Then

$$\mathbf{y}_{1:r} = D_r^{-1}\mathbf{z}_{1:r}, \tag{218}$$

or equivalently,

$$y_i = \frac{c_i}{\sigma_i}, \quad 1 \leq i \leq r.$$

Since $V$ is orthogonal, we have

$$\mathbf{x} = (V^T)^{-1}\mathbf{y} = V\mathbf{y},$$

with

$$||\mathbf{x}||_{\mathbb{R}^m} = ||V\mathbf{y}||_{\mathbb{R}^m} = ||\mathbf{y}||_{\mathbb{R}^m}.$$

The second normal equation, namely

$$||\mathbf{x}||_{\mathbb{R}^m} = \alpha,$$

then is satisfied if and only if

$$||\mathbf{y}_{1:r}||_{\mathbb{R}^r} \leq ||\mathbf{y}||_{\mathbb{R}^m} = ||\mathbf{x}||_{\mathbb{R}^m} = \alpha. \tag{219}$$

This condition is equivalent to

$$\sum_{i=1}^r \frac{c_i^2}{\sigma_i^2} \leq \alpha^2. \tag{220}$$

Assuming that this is satisfied, then

$$A\mathbf{x}(0) = U\Sigma V^T\mathbf{x} = U\Sigma V^T V\mathbf{y} = U\Sigma\mathbf{y} = U\Sigma_r\mathbf{y}_{1:r} = U\Sigma_r D_r^{-1}\mathbf{z}_{1:r}$$
$$= U\Sigma_r D_r^{-1}\Sigma_r^T(U^T\mathbf{b}) = UJ_r^n U^T\mathbf{b}.$$

The minimum value is thus

$$||A\mathbf{x}(0) - \mathbf{b}||_{\mathbb{R}^n} = ||(UJ_r^n U^T - I_n)\mathbf{b}||_{\mathbb{R}^n} = ||(UJ_r^n U^T - UU^T)\mathbf{b}||_{\mathbb{R}^n}$$
$$= ||U(J_r^n - I_n)U^T\mathbf{b}||_{\mathbb{R}^n} = ||(J_r^n - I_n)U^T\mathbf{b}||_{\mathbb{R}^n}.$$

If $r = n$, then $J_r^n = I_n$, and

$$||A\mathbf{x}(0) - \mathbf{b}||_{\mathbb{R}^m} = 0.$$

Since $s(0) = \sum_{i=1}^r \frac{c_i^2}{\sigma_i^2}$ and $s$ is monotonically decreasing on $(-\sigma_r^2, \infty)$, we have

$$\sum_{i=1}^r \frac{c_i^2}{\sigma_i^2} = \alpha^2 \Longleftrightarrow \gamma^* = 0. \tag{221}$$

In this case, because $\sum_{i=1}^{r} y_i^2 = \alpha^2$, we must have $y_{r+1} = \cdots = y_m = 0$ and consequently $\mathbf{x}(0)$ is the unique global minimum. If

$$\sum_{i=1}^{r} \frac{c_i^2}{\sigma_i^2} < \alpha^2, \tag{222}$$

then $\gamma^* \neq 0$. In this case, we can choose arbitrary values $y_{r+1}, \ldots, y_m$ such that $y_{r+1}^2 + \cdots + y_m^2 = \alpha^2 - \sum_{i=1}^{r} \frac{c_i^2}{\sigma_i^2}$. Consequently, there are infinitely many solutions $\mathbf{x} = V\mathbf{y}$ which achieve the global minimum.

If condition (220) is not met, that is $\sum_{i=1}^{r} \frac{c_i^2}{\sigma_i^2} > \alpha^2$, then the second normal equation $||\mathbf{x}||_{\mathbb{R}^m} = \alpha$ cannot be satisfied and thus there is no solution for the case $\gamma = 0$. Thus the global solution is still $\mathbf{x}(\gamma^*)$. This completes the proof of the theorem. ∎

For completeness, we provide the proofs of Lemmas 34 and 35 here. Lemma 34 is a special case of Theorem 1 in (Gander, 1981) and thus the proof given here is considerably simpler. Our proof for Lemma 35 is different from that given in (Gander, 1981), since we do *not* make the assumption that rank $\begin{pmatrix} A \\ I \end{pmatrix} = m$.

**Proof of Lemma 34** By equation (208), we have

$$\frac{\gamma_1 - \gamma_2}{2}||\mathbf{x}_1 - \mathbf{x}_2||^2 = \frac{\gamma_1 - \gamma_2}{2}(||\mathbf{x}_1||^2 + ||\mathbf{x}_2||^2 - 2\langle\mathbf{x}_1, \mathbf{x}_2\rangle) = (\gamma_1 - \gamma_2)\alpha^2 + (\gamma_2 - \gamma_1)\langle\mathbf{x}_1, \mathbf{x}_2\rangle$$

From equation (207),

$$||A\mathbf{x}_2 - \mathbf{b}||^2 - ||A\mathbf{x}_1 - \mathbf{b}||^2 = (\langle\mathbf{x}_2, A^T A\mathbf{x}_2\rangle - 2\langle\mathbf{x}_2, A^T\mathbf{b}\rangle) - (\langle\mathbf{x}_1, A^T A\mathbf{x}_1\rangle - 2\langle\mathbf{x}_1, A^T\mathbf{b}\rangle)$$

$$= (\langle\mathbf{x}_2, A^T\mathbf{b} - \gamma_2\mathbf{x}_2\rangle - 2\langle\mathbf{x}_2, A^T\mathbf{b}\rangle) - (\langle\mathbf{x}_1, A^T\mathbf{b} - \gamma_1\mathbf{x}_1\rangle - 2\langle\mathbf{x}_1, A^T\mathbf{b}\rangle)$$

$$= \gamma_1||\mathbf{x}_1||^2 + \langle\mathbf{x}_1, A^T\mathbf{b}\rangle - \gamma_2||\mathbf{x}_2||^2 - \langle\mathbf{x}_2, A^T\mathbf{b}\rangle = (\gamma_1 - \gamma_2)\alpha^2 + \langle\mathbf{x}_1, A^T\mathbf{b}\rangle - \langle\mathbf{x}_2, A^T\mathbf{b}\rangle.$$

Also from equation (207), we have

$$\langle\mathbf{x}_1, (A^T A + \gamma_2 I_m)\mathbf{x}_2\rangle = \langle\mathbf{x}_1, A^T\mathbf{b}\rangle,$$

$$\langle\mathbf{x}_2, (A^T A + \gamma_1 I_m)\mathbf{x}_1\rangle = \langle\mathbf{x}_2, A^T\mathbf{b}\rangle,$$

Subtracting the second expression from the first, we obtain

$$\langle\mathbf{x}_1, A^T\mathbf{b}\rangle - \langle\mathbf{x}_2, A^T\mathbf{b}\rangle = (\gamma_2 - \gamma_1)\langle\mathbf{x}_1, \mathbf{x}_2\rangle.$$

Thus

$$||A\mathbf{x}_2 - \mathbf{b}||^2 - ||A\mathbf{x}_1 - \mathbf{b}||^2 = (\gamma_1 - \gamma_2)\alpha^2 + (\gamma_2 - \gamma_1)\langle\mathbf{x}_1, \mathbf{x}_2\rangle = \frac{\gamma_1 - \gamma_2}{2}||\mathbf{x}_1 - \mathbf{x}_2||^2.$$

This completes the proof. ∎

**Proof of Lemma 35** There are two possible cases under which the right hand side of (209) is equal to zero.

(I) $\gamma_1 = \gamma_2 = \gamma$ and $\mathbf{x}_1 \neq \mathbf{x}_2$. By equation (207),

$$(A^T A + \gamma I_m)(\mathbf{x}_1 - \mathbf{x}_2) = 0 \iff A^T A(\mathbf{x}_1 - \mathbf{x}_2) = -\gamma(\mathbf{x}_1 - \mathbf{x}_2).$$

This means that $\gamma = -\mu$, where $\mu \geq 0$ is an eigenvalue of $A^T A$ and

$$\mathbf{x}_1 = \mathbf{x}_2 + \mathbf{v}(\mu),$$

where $\mathbf{v}(\mu)$ is an eigenvector corresponding to $\mu$.

(II) $\gamma_1 \neq \gamma_2$ and $\mathbf{x}_1 = \mathbf{x}_2 = \mathbf{x}$. This case is not possible, since by equation (207), we have

$$(A^T A + \gamma_1 I_m)\mathbf{x} = A^T \mathbf{b} = (A^T A + \gamma_2 I_m)\mathbf{x} \implies (\gamma_1 - \gamma_2)\mathbf{x} = 0 \implies \mathbf{x} = 0,$$

contradicting the assumption $\alpha > 0$. ∎

## Appendix B. Learning with General Bounded Linear Operators

The present framework generalizes naturally beyond the point evaluation operator

$$f(x) = K_x^* f.$$

Let $\mathcal{H}$ be a separable Hilbert space of functions on $\mathcal{X}$. We are *not* assuming that the functions in $\mathcal{H}$ are defined pointwise or with values in $\mathcal{W}$, rather we assume that $\forall x \in \mathcal{X}$, there is a bounded linear operator

$$E_x : \mathcal{H} \to \mathcal{W}, \quad ||E_x|| < \infty, \tag{223}$$

with adjoint $E_x^* : \mathcal{W} \to \mathcal{H}$. Consider the minimization

$$f_{\mathbf{z},\gamma} = \operatorname{argmin}_{\mathcal{H}_K} \frac{1}{l} \sum_{i=1}^{l} V(y_i, CE_{x_i} f) + \gamma_A ||f||_{\mathcal{H}}^2$$
$$+ \gamma_I \langle \mathbf{f}, M\mathbf{f} \rangle_{\mathcal{W}^{u+l}}, \quad \text{where} \quad \mathbf{f} = (E_{x_i} f)_{i=1}^{u+l}, \tag{224}$$

and its least square version

$$f_{\mathbf{z},\gamma} = \operatorname{argmin}_{\mathcal{H}_K} \frac{1}{l} \sum_{i=1}^{l} ||y_i - CE_{x_i} f||_{\mathcal{Y}}^2 + \gamma_A ||f||_{\mathcal{H}}^2 + \gamma_I \langle \mathbf{f}, M\mathbf{f} \rangle_{\mathcal{W}^{u+l}}. \tag{225}$$

Following are the corresponding Representer Theorem and Proposition stating the explicit solution for the least square case. When $\mathcal{H} = \mathcal{H}_K$, $E_x = K_x^*$, we recover Theorem 2 and Theorem 3, respectively.

**Theorem 36** *The minimization problem (224) has a unique solution, given by* $f_{\mathbf{z},\gamma} = \sum_{i=1}^{u+l} E_{x_i}^* a_i$ *for some vectors* $a_i \in \mathcal{W}$, $1 \leq i \leq u + l$.

**Proposition 37** *The minimization problem (225) has a unique solution $f_{\mathbf{z},\gamma} = \sum_{i=1}^{u+l} E_{x_i}^* a_i$, where the vectors $a_i \in \mathcal{W}$ are given by*

$$l\gamma_I \sum_{j,k=1}^{u+l} M_{ik} E_{x_k} E_{x_j}^* a_j + C^* C (\sum_{j=1}^{u+l} E_{x_i} E_{x_j}^* a_j) + l\gamma_A a_i = C^* y_i, \tag{226}$$

*for $1 \leq i \leq l$, and*

$$\gamma_I \sum_{j,k=1}^{u+l} M_{ik} E_{x_k} E_{x_j}^* a_j + \gamma_A a_i = 0, \tag{227}$$

*for $l+1 \leq i \leq u+l$.*

The reproducing kernel structures come into play through the following.

**Lemma 38** *Let $E : \mathcal{X} \times \mathcal{X} \to \mathcal{L}(\mathcal{W})$ be defined by*

$$E(x,t) = E_x E_t^*. \tag{228}$$

*Then $E$ is a positive definite operator-valued kernel.*

**Proof of Lemma 38.** For each pair $(x,t) \in \mathcal{X} \times \mathcal{X}$, the operator $E(x,t)$ satisfies

$$E(t,x)^* = (E_t E_x^*)^* = E_x E_t^* = E(x,t).$$

For every set $\{x_i\}_{i=1}^N$ in $\mathcal{X}$ and $\{w_i\}_{i=1}^N$ in $\mathcal{W}$,

$$\sum_{i,j=1}^N \langle w_i, E(x_i,x_j) w_j \rangle_{\mathcal{W}} = \sum_{i,j=1}^N \langle w_i, E_{x_i} E_{x_j}^* w_j \rangle_{\mathcal{W}}$$

$$= \sum_{i,j=1}^N \langle E_{x_i}^* w_i, E_{x_j}^* w_j \rangle_{\mathcal{H}} = \| \sum_{i=1}^N E_{x_i}^* w_i \|_{\mathcal{H}}^2 \geq 0.$$

Thus $E$ is an $\mathcal{L}(\mathcal{W})$-valued positive definite kernel. ∎

**Proof of Theorem 36 and Proposition 37**. These are entirely analogous to those of Theorem 2 and Theorem 3, respectively. Instead of the sampling operator $S_{\mathbf{x}}$, we consider the operator $E_{\mathbf{x}} : \mathcal{H} \to \mathcal{W}^l$, with

$$E_{\mathbf{x}} f = (E_{x_i} f)_{i=1}^l, \tag{229}$$

with the adjoint $E_{\mathbf{x}}^* : \mathcal{W}^l \to \mathcal{H}$ given by

$$E_{\mathbf{x}}^* \mathbf{b} = \sum_{i=1}^l E_{x_i}^* b_i. \tag{230}$$

for all $\mathbf{b} = (b_i)_{i=1}^l \in \mathcal{W}^l$. The operator $E_{C,\mathbf{x}} : \mathcal{H} \to \mathcal{Y}^l$ is now defined by

$$E_{C,\mathbf{x}} f = (C E_{x_1} f, \dots, C E_{x_l} f). \tag{231}$$

The adjoint $E^*_{C,\mathbf{x}} : \mathcal{Y}^l \to \mathcal{H}$ is

$$E^*_{C,\mathbf{x}} \mathbf{b} = \sum_{i=1}^{l} E^*_{x_i} C^* b_i, \tag{232}$$

for all $\mathbf{b} \in \mathcal{Y}^l$, and $E^*_{C,\mathbf{x}} E_{C,\mathbf{x}} : \mathcal{H} \to \mathcal{H}$ is

$$E^*_{C,\mathbf{x}} E_{C,\mathbf{x}} f = \sum_{i=1}^{l} E^*_{x_i} C^* C E_{x_i} f. \tag{233}$$

We then apply all the steps in the proofs of Theorem 2 and Theorem 3 to get the desired results. ∎

**Remark 39** *We stress that in general, the function $f_{\mathbf{z},\gamma}$ is not defined pointwise, which is the case in the following example. Thus one cannot make a statement about $f_{\mathbf{z},\gamma}(x)$ for all $x \in \mathcal{X}$ without additional assumptions.*

**Example 1** *Wahba (1977) $\mathcal{X} = [0,1]$, $\mathcal{H} = L^2(\mathcal{X})$, $\mathcal{W} = \mathbb{R}$. Let $G : \mathcal{X} \times \mathcal{X} \to \mathbb{R}$ be continuous and*

$$E_x f = \int_0^1 G(x,t) f(t) dt. \tag{234}$$

*for $f \in \mathcal{H}$. One has the reproducing kernel*

$$E_x E_t^* = E(x,t) = \int_0^1 G(x,u) G(t,u) du. \tag{235}$$

## Appendix C. The Degenerate Case

This section considers the Gaussian kernel $k(x,t) = \exp\left(-\frac{||x-t||^2}{\sigma^2}\right)$ when $\sigma \to \infty$ and other kernels with similar behavior. We show that for $G(x,t) = \sum_{i=1}^{m} k^i(x,t) \mathbf{e}_i \mathbf{e}_i^T$, $R = I_{\mathcal{Y}}$, the matrix $A$ in Theorem 10 has an analytic expression. This can be used to verify the correctness of an implementation of Algorithm 1.

At $\sigma = \infty$, for $R = I_{\mathcal{Y}}$, for each pair $(x,t)$, we have

$$K(x,t) = I_{\mathcal{Y}^m}, \tag{236}$$

and

$$f_{\mathbf{z},\gamma}(x) = \sum_{i=1}^{u+l} K(x_i,x) a_i = \sum_{i=1}^{u+l} a_i. \tag{237}$$

Thus $f_{\mathbf{z},\gamma}$ is a constant function. Let us examine the form of the coefficients $a_i$'s for the case

$$C = \frac{1}{m} \mathbf{1}_m^T \otimes I_{\mathcal{Y}}.$$

We have

$$G[\mathbf{x}] = \mathbf{1}_{u+l} \mathbf{1}_{u+l}^T \otimes I_m.$$

For $\gamma_I = 0$, we have

$$B = \frac{1}{m^2}(J_l^{u+l} \otimes \mathbf{1}_m \mathbf{1}_m^T)(\mathbf{1}_{u+l}\mathbf{1}_{u+l}^T \otimes I_m),$$

which is

$$B = \frac{1}{m^2}(J_l^{u+l} \mathbf{1}_{u+l}\mathbf{1}_{u+l}^T \otimes \mathbf{1}_m \mathbf{1}_m^T).$$

Equivalently,

$$B = \frac{1}{m^2}(J_{ml}^{(u+l)m} \mathbf{1}_{(u+l)m}\mathbf{1}_{(u+l)m}^T).$$

The inverse of $B + l\gamma_A I_{(u+l)m}$ in this case has a closed form:

$$(B + l\gamma_A I_{(u+l)m})^{-1} = \frac{I_{(u+l)m}}{l\gamma_A} - \frac{J_{ml}^{(u+l)m} \mathbf{1}_{(u+l)m}\mathbf{1}_{(u+l)m}^T}{l^2 m \gamma_A(m\gamma_A + 1)}, \tag{238}$$

where we have used the identity

$$\mathbf{1}_{(u+l)m}\mathbf{1}_{(u+l)m}^T J_{ml}^{(u+l)m} \mathbf{1}_{(u+l)m}\mathbf{1}_{(u+l)m}^T = ml\mathbf{1}_{(u+l)m}\mathbf{1}_{(u+l)m}^T. \tag{239}$$

We have thus

$$A = (B + l\gamma_A I_{(u+l)m})^{-1} Y_C = \left( \frac{I_{(u+l)m}}{l\gamma_A} - \frac{J_{ml}^{(u+l)m} \mathbf{1}_{(u+l)m}\mathbf{1}_{(u+l)m}^T}{l^2 m \gamma_A(m\gamma_A + 1)} \right) Y_C. \tag{240}$$

Thus in this case we have an analytic expression for the coefficient matrix $A$, as we claimed.